%% file: main.tex
\documentclass[final,5p,times,twocolumn]{elsarticle}
\usepackage{hyperref} 

\journal{Accepted at Journal of Computer Vision and Image Understanding (CVIU)}

\usepackage[table,xcdraw]{xcolor}
\usepackage{colortbl}
\usepackage{multirow}
\usepackage{verbatim}
\usepackage{arydshln}
\usepackage{subfig}
\usepackage[export]{adjustbox}
\usepackage{amsmath,amsfonts,amssymb}
\usepackage{url}

\newcommand\revision[1]{\color{black}#1}

\begin{document}

\begin{frontmatter}

\title{Knowledge Distillation for Incremental Learning in Semantic Segmentation}

\author{Umberto Michieli}

\author{Pietro Zanuttigh}
\address{University of Padova, Department of Information Engineering}




\begin{abstract}
Deep learning architectures have shown remarkable results in scene understanding problems, however they exhibit a critical drop of performances when they are required to learn incrementally new tasks without forgetting old ones. This catastrophic forgetting phenomenon impacts on the deployment of artificial intelligence in real world  scenarios where systems need to learn new and different representations over time. Current approaches for incremental learning deal only with image classification and object detection tasks, while in this work we formally introduce incremental learning for semantic segmentation. We  tackle the problem applying various knowledge distillation techniques on the previous model. In this way, we retain the information about learned classes, whilst updating the current model to learn the new ones. We developed four main methodologies of knowledge distillation working on both output layers and internal feature representations. We do not store any image belonging to previous training stages and only the last model is used to preserve high accuracy on previously learned classes. Extensive experimental results on the Pascal VOC2012 and MSRC-v2 datasets show the effectiveness of the proposed approaches in several incremental learning scenarios.
\end{abstract}

\begin{keyword}
Incremental Learning\sep Continual Learning \sep Semantic Segmentation \sep Catastrophic Forgetting \sep Knowledge Distillation
\MSC[2010] 68T45\sep  68U10 \sep 68T05 
\end{keyword}
\end{frontmatter}


\input{sections/introduction.tex}
\input{sections/related.tex}

\input{sections/problem.tex}
\input{sections/methods.tex}
\input{sections/training.tex}
\input{sections/results.tex}

\input{sections/conclusion.tex}

\bibliography{refs}

\pagebreak
\pagebreak

\onecolumn
\begin{center}
  \textbf{\large Knowledge Distillation for Incremental Learning in Semantic Segmentation\\Supplementary Material}\\[.2cm]
  Umberto Michieli and Pietro Zanuttigh\\[.1cm]
  {\itshape University of Padova,\\ Department of Information Engineering, Via G. Gradenigo 6/b, 35131, Padova, Italy}
\end{center}

\setcounter{equation}{0}
\setcounter{figure}{0}
\setcounter{table}{0}
\setcounter{page}{1}
\renewcommand{\theequation}{S\arabic{equation}}
\renewcommand{\thefigure}{S\arabic{figure}}
\renewcommand{\bibnumfmt}[1]{[S#1]}
\renewcommand{\citenumfont}[1]{S#1}
\renewcommand{\thetable}{S\arabic{table}}
\renewcommand{\thepage}{S\arabic{page}} 

In this document we present some additional results related to the paper \textit{Knowledge Distillation for Incremental Learning in Semantic Segmentation}.
{\revision In particular, we analyze the effect of different pre-training strategies and we report the per-class pixel accuracy in various scenarios on the Pascal VOC2012 and MSRC-v2 datasets. Finally, we include an ablation study on multi-layer knowledge distillation.}

\input{sections/supplementary.tex}

\begin{table*}[htbp]
\vspace{-0.1cm}
\caption{Per-class pixel accuracy of the proposed approaches on VOC2012 when the last class, i.e., the tv/monitor class, is added.}
\vspace{-0.2cm}
\label{tab:pascal_0_19_20_pixelaccuracy}
\setlength{\tabcolsep}{1.6pt}
\centering
\footnotesize
\begin{tabular}{|c|cccccccccccccccccccc:c|c|ccc|}
\hline
 $M_1 (20)$ & \scriptsize\rotatebox{90}{backgr.} &  \scriptsize\rotatebox{90}{aero} &  \scriptsize\rotatebox{90}{bike} &  \scriptsize\rotatebox{90}{bird} &\scriptsize\rotatebox{90}{boat} & \scriptsize\rotatebox{90}{bottle} & \scriptsize\rotatebox{90}{bus} 
  &\scriptsize\rotatebox{90}{car} & \scriptsize\rotatebox{90}{cat} & \scriptsize\rotatebox{90}{chair} & \scriptsize\rotatebox{90}{cow} & \scriptsize\rotatebox{90}{din. table}& \scriptsize\rotatebox{90}{dog} & \scriptsize\rotatebox{90}{horse} 
  & \scriptsize\rotatebox{90}{mbike} & \scriptsize\rotatebox{90}{person} & \scriptsize\rotatebox{90}{plant} &  \scriptsize\rotatebox{90}{sheep} & \scriptsize\rotatebox{90}{sofa} & \scriptsize\rotatebox{90}{train} & \scriptsize\rotatebox{90}{\textbf{mCA old}} & \scriptsize\rotatebox{90}{tv} & \scriptsize\rotatebox{90}{\textbf{mIoU}} & \scriptsize\rotatebox{90}{\textbf{mPA}} & \scriptsize\rotatebox{90}{\textbf{mCA}}\\
 \hline

Fine-tuning & 94.7 & 83.9 & 66.9 & 90.2 & 63.9 & 80.2 & 86.6 & 81.6 & 93.6 & 46.1 & 80.9 & 57.4 & 87.8 & 90.4 & 72.4 & 87.8 & 48.5 & 81.1 & 56.2 & 71.8 & 76.1 & \textbf{84.3} & 65.1 & 90.7 & 76.5 \\

$\mathcal{L}_{D}'*$ {\tiny \citep{michieli2019}} & 96.2 & 89.8 & 77.0 & 93.2 & 73.1 & 83.7 & 93.7 & 91.0 & 94.4 & 44.5 & 79.8 & 55.4 & 86.6 & 91.7 & 85.3 & 88.2 & 50.3 & 81.8 & 52.9 & 85.5 & 79.7 & 76.4 & 68.4 & 92.5 & 79.5 \\

$\mathcal{L}_{D}'$ & 96.6 & 92.5 & 56.2 & 93.7 & 76.3 & 84.5 & 95.4 & 90.1 & 95.9 & 41.4 & 86.1 & 61.3 & 90.4 & 91.3 & 83.8 & 88.7 & 60.8 & 83.7 & 66.1 & 83.7 & 80.9 & 64.6 & 70.4 & 93.2 & 80.1 \\

$E_F$ & 96.5 & 91.4 & 56.6 & 94.6 & 79.8 & 83.0 & 97.1 & 87.4 & 95.5 & 33.8 & 90.1 & 66.8 & 92.6 & 94.1 & 89.2 & 86.1 & 64.0 & 87.9 & 62.5 & 86.6 & 81.8 & 74.3 & 70.5 & 93.2 & 81.4 \\

$E_F$, $\mathcal{L}_{D}'$ & 96.7 & 92.3 & 73.8 & 93.5 & 84.6 & 87.3 & 97.5 & 91.8 & 94.9 & 38.7 & 88.4 & 63.2 & 92.0 & 92.5 & 91.6 & 88.5 & 62.7 & 86.5 & 64.0 & 91.0 & 83.6 & 57.6 & \textbf{72.0} & \textbf{93.6} & 82.3 \\

$E_{2LF}$, $\mathcal{L}_{D}'$ & 96.6 & 89.9 & 64.9 & 93.1 & 76.5 & 85.0 & 95.3 & 91.4 & 95.7 & 41.8 & 86.5 & 60.3 & 90.9 & 90.6 & 83.5 & 88.1 & 59.7 & 84.7 & 65.3 & 87.9 & 81.4 & 67.4 & 70.7 & 93.2 & 80.7 \\

$\mathcal{L}_{D}''$ & 96.2 & 93.3 & 69.6 & 95.4 & 84.5 & 87.0 & 97.6 & 91.0 & 96.1 & 41.2 & 91.4 & 71.6 & 93.4 & 91.5 & 90.9 & 85.8 & 63.2 & 85.1 & 69.4 & 88.6 & \textbf{84.1} & 68.0 & 71.6 & 93.4 & \textbf{83.4} \\

$\mathcal{L}_{D}'''$ & 95.6 & 91.0 & 64.7 & 93.3 & 76.7 & 87.3 & 95.8 & 90.2 & 95.1 & 53.1 & 86.0 & 65.0 & 92.5 & 87.9 & 82.1 & 88.6 & 66.2 & 83.6 & 57.9 & 90.8 & 82.2 & 67.5 & 69.5 & 92.6 & 81.5 \\

$E_F$, $\mathcal{L}_{D}'''$ & 95.8 & 94.8 & 58.8 & 95.0 & 86.8 & 88.7 & 96.4 & 89.9 & 95.6 & 41.1 & 90.9 & 68.7 & 94.2 & 91.7 & 89.9 & 86.3 & 70.5 & 88.7 & 61.6 & 88.7 & 83.7 & 60.9 & 70.1 & 93.0 & 82.6 \\
 
 $\mathcal{L}_{D}''''^{*}$ {\tiny \citep{SPKD}} & 96.1 & 92.3 & 68.3 & 93.9 & 75.8 & 83.7 & 91.0 & 88.8 & 95.3 & 47.0 & 91.4 & 64.6 & 91.1 & 92.0 & 86.9 & 86.9 & 63.0 & 87.3 & 66.0 & 85.9 & 82.4 & 75.1 & 70.2 & 92.9 & 82.0
 \\
 
 $\mathcal{L}_{D}''''$ &  96.1 & 91.8 & 63.9 & 93.7 & 79.7 & 85.0 & 92.7 & 89.0 & 95.6 & 48.2 & 90.5 & 66.4 & 91.2 & 92.2 & 86.2 & 87.0 & 63.1 & 86.7 & 66.2 & 87.4 & 82.6 & 75.3 & 70.6 & 93.1 & 82.3
 \\\hline

$M_0(0-19)$ & 96.8 & 94.6 & 82.1 & 92.4 & 85.6 & 88.4 & 97.1 & 92.2 & 94.1 & 48.5 & 89.2 & 64.0 & 90.6 & 90.4 & 90.3 & 88.8 & 65.5 & 86.0 & 56.0 & 92.6 & 84.3 & -  & 73.4 & 93.9 & 84.3 \\

$M_0(0-20)$ & 96.7 & 94.8 & 76.6 & 92.6 & 86.3 & 87.4 & 97.7 & 93.0 & 94.6 & 52.2 & 90.1 & 55.8 & 90.9 & 90.5 & 89.6 & 89.8 & 65.1 & 86.2 & 63.8 & 93.6 & 84.4 & 81.0 & 73.6 & 93.9 & 84.2 \\
\hline
\end{tabular}
\end{table*}

\begin{table*}[htbp]
\vspace{-0.1cm}
\caption{Per-class pixel accuracy of the proposed approaches on VOC2012 when $5$ classes are added at once.}
\vspace{-0.2cm}
\label{tab:pascal_0_15_20_pixelaccuracy}
\setlength{\tabcolsep}{1.6pt}
\centering
\footnotesize
\begin{tabular}{|c|cccccccccccccccc:c|ccccc:c|ccc|}
\hline
 $M_1 (16-20)$ & \scriptsize\rotatebox{90}{backgr.} &  \scriptsize\rotatebox{90}{aero} &  \scriptsize\rotatebox{90}{bike} &  \scriptsize\rotatebox{90}{bird} &\scriptsize\rotatebox{90}{boat} & \scriptsize\rotatebox{90}{bottle} & \scriptsize\rotatebox{90}{bus} 
  &\scriptsize\rotatebox{90}{car} & \scriptsize\rotatebox{90}{cat} & \scriptsize\rotatebox{90}{chair} & \scriptsize\rotatebox{90}{cow} & \scriptsize\rotatebox{90}{din. table}& \scriptsize\rotatebox{90}{dog} & \scriptsize\rotatebox{90}{horse} 
  & \scriptsize\rotatebox{90}{mbike} & \scriptsize\rotatebox{90}{person} & \scriptsize\rotatebox{90}{\textbf{mCA old}} & \scriptsize\rotatebox{90}{plant} &  \scriptsize\rotatebox{90}{sheep} & \scriptsize\rotatebox{90}{sofa} & \scriptsize\rotatebox{90}{train} & \scriptsize\rotatebox{90}{tv} & \scriptsize\rotatebox{90}{\textbf{mCA new}} & \scriptsize\rotatebox{90}{\textbf{mIoU}} & \scriptsize\rotatebox{90}{\textbf{mPA}} & \scriptsize\rotatebox{90}{\textbf{mCA}}\\
 \hline

Fine-tuning & 93.5 & 59.7 & 69.5 & 69.3 & 66.9 & 64.7 & 59.9 & 87.2 & 86.6 & 36.5 & 12.7 & 53.9 & 79.7 & 63.8 & 83.1 & 89.7 & 67.3 & 61.0 & 93.0 & 79.6 & 92.9 & 80.0 & \textbf{81.3} & 55.4 & 88.4 & 70.6 \\

{\revision $E_F$} & 94.8 & 73.3 & 58.7 & 57.7 & 57.0 & 74.3 & 10.5 & 85.6 & 90.8 & 29.8 & 10.6 & 69.7 & 89.9 & 70.4 & 85.9 & 85.0 & 65.2 & 59.9 & 93.7 & 74.7 & 87.6 & 71.7 & 77.5 & 54.2 & 88.5 & 68.2 \\

$\mathcal{L}_{D}'$ & 95.5 & 88.7 & 77.0 & 90.6 & 83.1 & 80.2 & 88.6 & 93.2 & 90.0 & 47.9 & 75.6 & 47.3 & 81.5 & 85.7 & 85.9 & 90.4 & \textbf{81.3} & 39.7 & 88.0 & 55.6 & 86.2 & 68.1 & 67.5 & \textbf{65.8} & \textbf{91.6} & \textbf{78.1} \\

$E_F$, $\mathcal{L}_{D}'$ & 96.0 & 86.6 & 71.3 & 80.3 & 79.1 & 81.2 & 66.7 & 91.3 & 92.4 & 43.9 & 60.0 & 63.2 & 89.9 & 83.3 & 89.6 & 88.2 & 78.9 & 44.7 & 89.4 & 57.0 & 83.0 & 61.0 & 67.0 & 64.3 & 91.5 & 76.1 \\

$E_{2LF}$, $\mathcal{L}_{D}'$ & 94.9 & 81.8 & 75.3 & 86.4 & 78.2 & 74.5 & 83.2 & 92.5 & 90.4 & 45.2 & 61.2 & 49.7 & 81.4 & 84.0 & 85.1 & 89.9 & 78.4 & 47.5 & 90.8 & 67.2 & 88.1 & 71.0 & 72.9 & 64.0 & 91.0 & 77.1 \\

$\mathcal{L}_{D}''$ & 94.4 & 84.3 & 67.1 & 82.6 & 74.8 & 77.0 & 74.0 & 89.8 & 92.0 & 39.7 & 0.4 & 62.7 & 86.5 & 75.1 & 87.0 & 89.1 & 73.5 & 58.7 & 90.9 & 78.9 & 90.9 & 76.7 & 79.2 & 60.5 & 90.0 & 74.9 \\

$\mathcal{L}_{D}'''$ & 94.5 & 91.9 & 62.4 & 85.1 & 78.5 & 83.0 & 85.4 & 92.2 & 93.2 & 42.0 & 35.2 & 62.3 & 91.4 & 86.5 & 87.9 & 90.5 & 78.9 & 51.0 & 89.3 & 69.4 & 87.6 & 66.4 & 72.7 & 63.9 & 91.0 & 77.4 \\
 
 $\mathcal{L}_{D}''''$ &  94.2 & 84.9 & 63.3 & 82.7 & 73.2 & 75.0 & 69.1 & 90.5 & 94.2 & 40.4 & 4.8 & 62.1 & 85.9 & 65.2 & 87.2 & 87.2 & 72.5 & 57.6 & 90.6 & 77.9 & 90.2 & 77.3 & 78.7 & 59.5 & 89.7 & 74.0
 \\ \hline

$M_0(0-15)$ & 96.9 & 94.8 & 77.8 & 93.4 & 87.1 & 86.7 & 97.1 & 92.8 & 94.8 & 53.5 & 91.3 & 56.6 & 90.2 & 89.6 & 90.9 & 89.4 & 86.4 & - & - & - & - & - & - & 75.5 & 94.6 & 86.4 \\

$M_0(0-20)$ & 96.7 & 94.8 & 76.6 & 92.6 & 86.3 & 87.4 & 97.7 & 93.0 & 94.6 & 52.2 & 90.1 & 55.8 & 90.9 & 90.5 & 89.6 & 89.8 & 86.2 & 65.1 & 86.2 & 63.8 & 93.6 & 81.0 & 78.0 & 73.6 & 93.9 & 84.2 \\
\hline
\end{tabular}
\end{table*}

\begin{table*}[htbp]
\vspace{-0.1cm}
\caption{Per-class pixel accuracy of the proposed approaches on VOC2012 when $10$ classes are added sequentially.}
\vspace{-0.2cm}
\label{tab:pascal_0_10_11_12_13_14_15_16_17_18_19_20_pixel_accuracy}
\setlength{\tabcolsep}{1.6pt}
\centering
\footnotesize
\begin{tabular}{|c|ccccccccccc:c|c:c:c:c:c:c:c:c:c:c:c|ccc|}
\hline
 $M_{10} (11\rightarrow20)$ & \scriptsize\rotatebox{90}{backgr.} &  \scriptsize\rotatebox{90}{aero} &  \scriptsize\rotatebox{90}{bike} &  \scriptsize\rotatebox{90}{bird} &\scriptsize\rotatebox{90}{boat} & \scriptsize\rotatebox{90}{bottle} & \scriptsize\rotatebox{90}{bus} 
  &\scriptsize\rotatebox{90}{car} & \scriptsize\rotatebox{90}{cat} & \scriptsize\rotatebox{90}{chair} & \scriptsize\rotatebox{90}{cow} & \scriptsize\rotatebox{90}{\textbf{mCA old}} & \scriptsize\rotatebox{90}{din. table}& \scriptsize\rotatebox{90}{dog} & \scriptsize\rotatebox{90}{horse} 
  & \scriptsize\rotatebox{90}{mbike} & \scriptsize\rotatebox{90}{person} & \scriptsize\rotatebox{90}{plant} &  \scriptsize\rotatebox{90}{sheep} & \scriptsize\rotatebox{90}{sofa} & \scriptsize\rotatebox{90}{train} & \scriptsize\rotatebox{90}{tv} & \scriptsize\rotatebox{90}{\textbf{mCA new}} & \scriptsize\rotatebox{90}{\textbf{mIoU}} & \scriptsize\rotatebox{90}{\textbf{mPA}} & \scriptsize\rotatebox{90}{\textbf{mCA}}\\
 \hline

Fine-tuning & 93.7 & 29.0 & 55.2 & 49.8 & 2.2 & 59.8 & 1.5 & 55.3 & 89.7 & 38.4 & 7.8 & 43.8 & 51.0 & 78.2 & 17.1 & 23.4 & \textbf{89.3} & 30.1 & 62.0 & 66.8 & 43.5 & \textbf{84.7} & 54.6 & 37.1 & 83.6 & 49.0 \\

{\revision $E_F$} & 94.0 & 50.5 & 45.1 & 73.2 & 57.8 & 74.2 & 73.8 & 77.6 & 35.1 & 30.2 & 55.8 & 60.7 & 52.9 & 70.5 & 45.2 & 40.1 & 80.1 & 41.2 & 66.2 & 60.1 & 65.7 & 57.9 & 58.0 & 54.4 & 85.8 & 59.4 \\

$\mathcal{L}_{D}'$ & 93.2 & 28.1 & 48.5 & 47.4 & 0.2 & 50.4 & 1.6 & 60.7 & 79.2 & 25.3 & 28.0 & 42.0 & 45.4 & 66.3 & 22.3 & 31.4 & 88.9 & 32.4 & \textbf{83.5} & \textbf{78.8} & \textbf{81.3} & 79.6 & 61.0 & 37.5 & 83.8 & 51.1 \\

$E_F$, $\mathcal{L}_{D}'$ & 94.8 & 66.5 & 53.9 & 75.0 & 66.0 & 80.5 & 75.4 & 82.4 & 87.8 & 36.8 & 60.5 & \textbf{70.9} & 56.8 & 80.3 & \textbf{47.6} & 42.7 & 81.8 & \textbf{50.3} & 68.8 & 64.7 & 72.1 & 65.1 & 63.0 & {\revision 54.9} & {\revision 88.5} & {\revision\textbf{67.1}} \\


$\mathcal{L}_{D}''$ & 94.7 & 54.9 & 53.1 & 77.0 & 46.5 & 81.2 & 74.5 & 82.9 & 95.6 & 41.6 & 60.1 & 69.3 & \textbf{57.7} & \textbf{84.9} & 40.0 & \textbf{50.5} & 80.8 & 49.2 & 62.9 & 70.8 & 66.2 & 75.0 & \textbf{63.8} & \textbf{55.3} & \textbf{88.5} & 66.7 \\

$\mathcal{L}_{D}'''$ & 94.8 & 65.4 & 53.6 & 63.0 & 30.0 & 59.3 & 31.0 & 69.5 & 91.2 & 33.5 & 36.5 & 57.1 & 50.6 & 83.3 & 43.4 & 48.6 & \textbf{89.3} & 31.7 & 46.2 & 70.4 & 71.9 & 82.4 & 61.8 & 47.4 & 87.1 & 59.3 \\
 
 $\mathcal{L}_{D}''''$ & 94.9 & 65.1 & 56.6 & 70.0 & 36.9 & 75.8 & 78.5 & 84.9 & 93.9 & 47.1 & 17.0 & 65.5 & 56.6 & 77.8 & 26.4 & 34.6 & 88.4 & 37.2 & 47.7 & 67.4 & 59.9 & 81.8 & 57.8 & 50.0 & 87.8 & 61.8
  \\\hline
 

$M_0(0-10)$ & 96.9 & 94.7 & 88.2 & 93.7 & 89.5 & 90.4 & 96.7 & 94.6 & 96.2 & 58.1 & 95.2 & 90.4 & - & - & - & - & - & - & - & - & - & - & - & 78.4 & 96.1 & 90.4
 \\
 
$M_0(0-20)$ & 96.7 & 94.8 & 76.6 & 92.6 & 86.3 & 87.4 & 97.7 & 93.0 & 94.6 & 52.2 & 90.1 & 87.4 & 55.8 & 90.9 & 90.5 & 89.6 & 89.8 & 65.1 & 86.2 & 63.8 & 93.6 & 81.0 & 80.6 & 73.6 & 93.9 & 84.2 \\
\hline
\end{tabular}
\end{table*}

\begin{table*}[htbp]
\vspace{-0.1cm}
\caption{Per-class pixel accuracy of the proposed approaches on MSRC-v2 when the last class, i.e. boat, is added}
\vspace{-0.2cm}
\label{tab:MSRC_0_19_20_pixelaccuracy}
\setlength{\tabcolsep}{1.6pt}
\centering
\footnotesize
\begin{tabular}{|c|cccccccccccccccccccc:c|c|ccc|}
\hline
 $M_1 (20)$ & \scriptsize\rotatebox{90}{grass} &  \scriptsize\rotatebox{90}{building} &  \scriptsize\rotatebox{90}{sky} &  \scriptsize\rotatebox{90}{road} &\scriptsize\rotatebox{90}{tree} & \scriptsize\rotatebox{90}{water} & \scriptsize\rotatebox{90}{book} 
  &\scriptsize\rotatebox{90}{car} & \scriptsize\rotatebox{90}{cow} & \scriptsize\rotatebox{90}{bicycle} & \scriptsize\rotatebox{90}{flower} & \scriptsize\rotatebox{90}{body}& \scriptsize\rotatebox{90}{sheep} & \scriptsize\rotatebox{90}{sign} 
  & \scriptsize\rotatebox{90}{face} & \scriptsize\rotatebox{90}{cat} & \scriptsize\rotatebox{90}{chair} &  \scriptsize\rotatebox{90}{aeroplane} & \scriptsize\rotatebox{90}{dog} & \scriptsize\rotatebox{90}{bird}& \scriptsize\rotatebox{90}{\textbf{mCA old}} & \scriptsize\rotatebox{90}{boat} & \scriptsize\rotatebox{90}{\textbf{mIoU}} & \scriptsize\rotatebox{90}{\textbf{mPA}} & \scriptsize\rotatebox{90}{\textbf{mCA}}\\
 \hline

Fine-tuning & 98.5 & 92.4 & 95.5 & 71.4 & 94.2 & 98.2 & 94.8 & 97.3 & 91.8 & 97.2 & 95.4 & 89.2 & 95.9 & 94.2 & 98.1 & 83.1 & 90.0 & 79.7 & 80.6 & 75.3 & 90.6 & 68.9 & 83.7 & 92.4 & 89.6
 \\
 
{\revision $E_F$} & 98.2 & 90.5 & 94.8 & 59.3 & 95.8 & 98.4 & 98.7 & 98.6 & 92.0 & 97.5 & 97.7 & 89.1 & 97.5 & 96.0 & 97.3 & 98.7 & 95.7 & 89.6 & 93.0 & 79.8 & 92.9 & 67.6 & 86.0 & 92.5 & 91.7 \\

$\mathcal{L}_{D}'$ & 97.7 & 91.0 & 95.6 & 86.8 & 94.6 & 95.8 & 98.7 & 98.6 & 95.4 & 98.4 & 97.9 & 88.9 & 98.0 & 97.2 & 98.7 & 87.4 & 98.0 & 89.3 & 88.0 & 88.5 & 94.2 & 65.9 & 87.6 & 94.5 & 92.9
\\

$E_F$, $\mathcal{L}_{D}'$ &  97.7 & 90.3 & 95.3 & 84.3 & 95.6 & 96.1 & 99.0 & 99.1 & 94.1 & 98.4 & 97.9 & 89.0 & 98.4 & 97.3 & 98.3 & 98.8 & 99.1 & 92.3 & 93.9 & 89.4 & \textbf{95.2} & 64.4 & \textbf{88.6} & \textbf{94.6} & \textbf{93.7}
 \\

$\mathcal{L}_{D}''$ & 98.1 & 91.5 & 94.2 & 79.0 & 95.8 & 98.1 & 98.6 & 98.7 & 91.9 & 98.4 & 97.7 & 90.4 & 98.0 & 95.0 & 97.9 & 98.5 & 98.5 & 89.5 & 92.0 & 84.0 & 94.3 & 61.9 & 87.6 & 94.1 & 92.8
 \\

$\mathcal{L}_{D}'''$ & 98.1 & 90.4 & 96.1 & 82.3 & 94.6 & 97.7 & 97.6 & 99.1 & 91.8 & 99.0 & 97.9 & 91.9 & 98.6 & 98.6 & 98.4 & 78.2 & 96.0 & 89.3 & 84.9 & 90.9 & 93.6 & 42.3 & 85.0 & 93.9 & 91.1
 \\
 
 $\mathcal{L}_{D}''''$ &   97.9 & 91.7 & 95.1 & 64.0 & 92.8 & 98.7 & 98.9 & 98.8 & 97.0 & 98.3 & 97.9 & 93.9 & 97.9 & 93.3 & 97.7 & 77.4 & 94.1 & 84.7 & 82.6 & 80.9 & 91.7 & \textbf{78.7} & 84.8 & 92.4 & 91.1

 \\ \hline

$M_0(0-19)$ & 97.0 & 89.9 & 94.9 & 92.5 & 95.3 & 94.8 & 99.0 & 99.3 & 95.4 & 98.9 & 98.0 & 89.7 & 98.9 & 98.3 & 98.5 & 98.6 & 99.3 & 93.7 & 95.1 & 92.4 & 96.0 & - & 90.8 & 95.5 & 96.0
 \\

$M_0(0-20)$ & 97.3 & 88.9 & 95.8 & 91.4 & 95.5 & 96.3 & 99.2 & 99.3 & 97.4 & 99.1 & 98.0 & 92.8 & 98.8 & 93.9 & 98.7 & 95.7 & 98.5 & 90.4 & 94.2 & 94.5 & 93.4 & 89.6 & 90.5 & 95.4 & 95.5
 \\
\hline
\end{tabular}
\end{table*}

\begin{table*}[htbp]
\vspace{-0.1cm}
\caption{Per-class pixel accuracy of the proposed approaches on MSRC-v2 when $5$ classes are added at once.}
\vspace{-0.2cm}
\label{tab:MSRC_0_15_20_pixelaccuracy}
\setlength{\tabcolsep}{1.6pt}
\centering
\footnotesize
\begin{tabular}{|c|cccccccccccccccc:c|ccccc:c|ccc|}
\hline
 $M_1 (16-20)$ & \scriptsize\rotatebox{90}{grass} &  \scriptsize\rotatebox{90}{building} &  \scriptsize\rotatebox{90}{sky} &  \scriptsize\rotatebox{90}{road} &\scriptsize\rotatebox{90}{tree} & \scriptsize\rotatebox{90}{water} & \scriptsize\rotatebox{90}{book} 
  &\scriptsize\rotatebox{90}{car} & \scriptsize\rotatebox{90}{cow} & \scriptsize\rotatebox{90}{bicycle} & \scriptsize\rotatebox{90}{flower} & \scriptsize\rotatebox{90}{body}& \scriptsize\rotatebox{90}{sheep} & \scriptsize\rotatebox{90}{sign} 
  & \scriptsize\rotatebox{90}{face} & \scriptsize\rotatebox{90}{cat} & \scriptsize\rotatebox{90}{\textbf{mCA old}} & \scriptsize\rotatebox{90}{chair} &  \scriptsize\rotatebox{90}{aeroplane} & \scriptsize\rotatebox{90}{dog} & \scriptsize\rotatebox{90}{bird} & \scriptsize\rotatebox{90}{boat} & \scriptsize\rotatebox{90}{\textbf{mCA new}} & \scriptsize\rotatebox{90}{\textbf{mIoU}} & \scriptsize\rotatebox{90}{\textbf{mPA}} & \scriptsize\rotatebox{90}{\textbf{mCA}}\\
 \hline

Fine-tuning & 98.9 & 88.9 & 96.6 & 92.0 & 91.4 & 96.4 & 96.1 & 95.3 & 49.1 & 96.0 & 93.8 & 86.2 & 59.6 & 88.1 & 97.2 & 37.5 & 85.2 & 94.2 & 79.2 & 97.6 & 82.9 & 72.4 & \textbf{85.3} & 78.0 & 91.1 & 85.2 \\

{\revision $E_F$} & 97.7 & 87.5 & 94.5 & 89.9 & 93.1 & 92.6 & 97.9 & 96.7 & 80.3 & 98.3 & 95.8 & 73.4 & 84.7 & 94.9 & 97.2 & 58.5 & 89.6 & 91.4 & 70.1 & 90.0 & 76.5 & 60.2 & 77.6 & 80.7 & 92.1 & 86.2  \\

$\mathcal{L}_{D}'$ & 98.2 & 90.0 & 94.4 & 88.1 & 93.5 & 92.2 & 98.5 & 97.5 & 88.1 & 98.4 & 96.7 & 85.1 & 93.4 & 98.7 & 98.0 & 71.0 & 92.6 & 91.4 & 79.1 & 87.9 & 82.9 & 84.2 & 85.1 & 84.5 & 93.3 & \textbf{90.8} \\

$E_F$, $\mathcal{L}_{D}'$ & 97.9 & 88.0 & 95.0 & 90.2 & 94.3 & 93.3 & 99.0 & 98.2 & 84.9 & 99.0 & 97.2 & 70.7 & 89.1 & 97.8 & 97.7 & 60.2 & 90.8 & 90.6 & 67.9 & 91.1 & 77.1 & 60.6 & 77.5 & 80.9 & 92.4 & 87.6  \\

$\mathcal{L}_{D}''$ & 99.0 & 89.4 & 95.2 & 91.5 & 96.2 & 97.1 & 98.9 & 95.5 & 72.6 & 97.6 & 95.7 & 83.5 & 83.3 & 93.3 & 97.3 & 57.2 & 90.2 & 88.2 & 75.5 & 96.6 & 79.0 & 75.3 & 82.9 & 82.6 & 93.1 & 88.5 \\

$\mathcal{L}_{D}'''$ & 98.4 & 89.1 & 95.8 & 90.1 & 94.1 & 97.6 & 95.9 & 98.2 & 90.3 & 99.1 & 96.0 & 91.4 & 96.7 & 98.0 & 97.9 & 68.2 & \textbf{93.6} & 88.7 & 84.4 & 86.1 & 81.1 & 69.6 & 82.0 & \textbf{85.1} & \textbf{93.9} & \textbf{90.8}  \\
 
 $\mathcal{L}_{D}''''$ & 98.8 & 91.3 & 96.6 & 91.0 & 90.8 & 97.3 & 95.7 & 95.8 & 74.4 & 95.4 & 93.9 & 88.2 & 74.7 & 89.3 & 97.6 & 45.6 & 88.5 & 89.7 & 70.2 & 95.2 & 83.2 & 66.8 & 81.0 & 80.5 & 92.2 & 86.7  
 \\ \hline

$M_0(0-15)$ & 97.2 & 89.4 & 91.9 & 89.1 & 95.3 & 95.5 & 99.2 & 99.1 & 95.0 & 99.0 & 98.0 & 92.1 & 99.3 & 97.4 & 98.2 & 99.9 & 96.0 & - & - & - & - & - & - & 91.1 & 95.2 & 96.0 \\

$M_0(0-20)$ &  97.3 & 88.9 & 95.8 & 91.4 & 95.5 & 96.3 & 99.2 & 99.3 & 97.4 & 99.1 & 98.0 & 92.8 & 98.8 & 93.9 & 98.7 & 95.7 & 96.1 & 98.5 & 90.4 & 94.2 & 94.5 & 89.6 & 93.4 & 90.5 & 95.4 & 95.5 \\
\hline
\end{tabular}
\end{table*}

\begin{table*}[htbp]
\vspace{-0.1cm}
\caption{Per-class pixel accuracy of the proposed approaches on MSRC-v2 when $5$ classes are added sequentially.}
\vspace{-0.2cm}
\label{tab:MSRC_0_15_16_17_18_19_20_pixelaccuracy}
\setlength{\tabcolsep}{1.6pt}
\centering
\footnotesize
\begin{tabular}{|c|cccccccccccccccc:c|c:c:c:c:c:c|ccc|}
\hline
 $M_1 (16\rightarrow20)$ & \scriptsize\rotatebox{90}{grass} &  \scriptsize\rotatebox{90}{building} &  \scriptsize\rotatebox{90}{sky} &  \scriptsize\rotatebox{90}{road} &\scriptsize\rotatebox{90}{tree} & \scriptsize\rotatebox{90}{water} & \scriptsize\rotatebox{90}{book} 
  &\scriptsize\rotatebox{90}{car} & \scriptsize\rotatebox{90}{cow} & \scriptsize\rotatebox{90}{bicycle} & \scriptsize\rotatebox{90}{flower} & \scriptsize\rotatebox{90}{body}& \scriptsize\rotatebox{90}{sheep} & \scriptsize\rotatebox{90}{sign} 
  & \scriptsize\rotatebox{90}{face} & \scriptsize\rotatebox{90}{cat} & \scriptsize\rotatebox{90}{\textbf{mCA old}} & \scriptsize\rotatebox{90}{chair} &  \scriptsize\rotatebox{90}{aeroplane} & \scriptsize\rotatebox{90}{dog} & \scriptsize\rotatebox{90}{bird} & \scriptsize\rotatebox{90}{boat} & \scriptsize\rotatebox{90}{\textbf{mCA new}} & \scriptsize\rotatebox{90}{\textbf{mIoU}} & \scriptsize\rotatebox{90}{\textbf{mPA}} & \scriptsize\rotatebox{90}{\textbf{mCA}}\\
 \hline

Fine-tuning & 96.8 & 88.8 & 97.1 & 73.9 & 88.1 & 97.9 & 91.3 & 71.9 & 53.6 & 90.5 & 79.6 & 72.3 & 51.6 & 75.0 & 93.8 & 34.5 & 78.5 & 35.2 & 53.0 & 44.6 & 59.5 & \textbf{88.3} & 56.1 & 63.9 & 83.8 & 73.2
 \\

{\revision $E_F$} & 96.9 & 85.9 & 95.5 & 79.8 & 92.7 & 89.8 & 98.9 & 89.8 & 85.4 & 97.0 & 95.8 & 70.1 & 94.5 & 94.0 & 95.7 & 79.2 & 90.1 & 82.7 & 49.5 & 35.4 & 82.3 & 56.6 & 61.3 & 70.5 & 87.2 & 83.2 \\

$\mathcal{L}_{D}'$ & 97.6 & 87.1 & 93.7 & 88.2 & 94.1 & 78.6 & 99.9 & 97.3 & 94.2 & 98.5 & 95.5 & 72.1 & 93.2 & 95.0 & 97.0 & 66.1 & 90.5 & \textbf{88.1} & \textbf{72.6} & \textbf{56.8} & \textbf{88.8} & 70.7 & \textbf{75.4} & \textbf{78.5} & 90.7 & \textbf{86.9}
 \\

$E_F$, $\mathcal{L}_{D}'$ & 97.1 & 86.1 & 94.8 & 87.3 & 95.0 & 91.5 & 99.4 & 97.9 & 86.7 & 98.8 & 96.0 & 71.5 & 96.2 & 94.7 & 96.3 & 81.8 & 91.9 & 85.2 & 51.3 & 33.2 & 86.0 & 54.1 & 62.0 & 76.3 & 90.6 & 84.8
 \\

$\mathcal{L}_{D}''$ & 98.2 & 92.3 & 93.3 & 78.9 & 96.3 & 97.6 & 98.9 & 96.4 & 88.3 & 97.1 & 97.0 & 83.8 & 86.1 & 88.3 & 97.4 & 78.2 & 91.7 & 75.6 & 48.1 & 30.7 & 58.6 & 68.8 & 56.4 & 76.2 & 90.5 & 83.3
\\

$\mathcal{L}_{D}'''$ & 97.1 & 88.2 & 96.3 & 83.0 & 95.1 & 96.8 & 98.6 & 99.4 & 97.7 & 99.4 & 96.0 & 83.3 & 97.0 & 97.4 & 98.3 & 94.5 & \textbf{94.9} & 41.7 & 64.2 & 43.7 & 79.8 & 27.5 & 51.4 & 76.6 & \textbf{91.1} & 84.5
 \\
 
 $\mathcal{L}_{D}''''$ &  97.5 & 90.8 & 97.0 & 65.3 & 91.4 & 98.3 & 96.5 & 91.9 & 22.8 & 94.9 & 94.2 & 91.3 & 48.9 & 85.5 & 96.0 & 30.0 & 80.7 & 55.8 & 51.8 & 45.5 & 65.2 & 78.2 & 59.3 & 66.8 & 85.4 & 75.6

 \\ \hline

$M_0(0-15)$ & 97.2 & 89.4 & 91.9 & 89.1 & 95.3 & 95.5 & 99.2 & 99.1 & 95.0 & 99.0 & 98.0 & 92.1 & 99.3 & 97.4 & 98.2 & 99.9 & 96.0 & - & - & - & - & - & - & 91.1 & 95.2 & 96.0 \\

$M_0(0-20)$ &  97.3 & 88.9 & 95.8 & 91.4 & 95.5 & 96.3 & 99.2 & 99.3 & 97.4 & 99.1 & 98.0 & 92.8 & 98.8 & 93.9 & 98.7 & 95.7 & 96.1 & 98.5 & 90.4 & 94.2 & 94.5 & 89.6 & 93.4 & 90.5 & 95.4 & 95.5 \\
\hline
\end{tabular}
\end{table*}

\end{document}

%% file: sections/introduction.tex
\section{Introduction}
\label{sec:intro}

Deep neural networks are nowadays gaining huge popularity and  are one of the key driving elements for the widespread diffusion of artificial intelligence. Despite their success on many visual recognition tasks, neural networks struggle with the incremental learning problem, i.e., improving the learned model to accomplish new tasks without losing previous knowledge.
Traditional training strategies typically require that all the samples corresponding to old and new tasks are available at training time and are not designed to work with new streams of data relative to new tasks only. A system deployed into the real world environment, instead, should be able to update its knowledge. 
Such behavior is inherently present in the human brain which is able to continuously incorporate new tasks while  preserving existing knowledge.
For this reason, incremental learning is gaining wide relevance among the scientific community and has already been explored in image classification and object detection \citep{shmelkov2017incremental, wu2018incremental, castro2018end, rebuffi2017icarl, li2018learning, zhou2019M2KD, furlanello2016active, shin2017continual, aljundi2018memory}.

On the other hand, incremental learning in  dense labeling tasks, such as semantic segmentation, has never been extensively studied.
Nevertheless, semantic segmentation is a key task that artificial intelligence systems must face frequently in various applications, e.g., autonomous driving or robotics \citep{biasetton2019unsupervised, michieli2019adversarial}. 
Differently from image classification, in semantic segmentation each image contains together pixels belonging to many classes (exemplars of new and old classes could co-exist). For this reason, incremental learning in image classification and in semantic segmentation are conceptually distinct.
Differently from many existing methodologies, we consider the most challenging setting where images from old tasks are not stored and cannot be used to drive the incremental learning process. This is particularly important for the vast majority of real world applications where old images are not available due to storage requirements or privacy concerns.

The aim of this paper is to introduce a novel framework to perform incremental learning in semantic segmentation. To the best of our knowledge this work and our conference paper \citep{michieli2019} are the first investigations on incremental learning for semantic segmentation which do not retain previously seen images and work on standard real world benchmarks. 
Specifically, we re-frame the distillation loss concept used in image classification and we propose four novel approaches where knowledge is distilled from the output layer, from intermediate features and from intermediate layers of the decoding phase.
Experimental results on the Pascal VOC2012 and MSRC-v2 datasets demonstrate that the proposed framework is robust in many different settings and across different datasets, {\revision without any previous sample available and even without labeling previous classes in new samples}. 
The proposed schemes 
allow not only to retain the learned information but also to achieve higher accuracy on new tasks, leading to substantial improvements in all the scenarios with respect to the standard approach without distillation.

This paper moves from our previous work \citep{michieli2019}, which is the first investigation on incremental learning for semantic segmentation. 
Compared to it, 
the main contributions of this journal version are the following:
\begin{itemize}
\item The distillation scheme on the output layer is improved and now considers also the uncertainty of the estimations of previous model.
\item A distillation constraint for the intermediate layers inspired from the Similarity-Preserving Knowledge Distillation \citep{SPKD} is introduced. 
\item A novel distillation scheme is proposed to enforce the similarity of multiple decoding stages simultaneously. 
\item A new strategy consisting in freezing only the first layers of the encoder is introduced to preserve unaltered the most task-agnostic part of the feature extraction. 
{\revision \item We investigated the \textit{incremental labeling} scenario in which old classes in future steps are labeled as background.}
\item Extensive experiments are conducted on many different scenarios. The results are reported on Pascal VOC2012 but also on the MSRC-v2 dataset to validate the generalization properties of the proposed methods. 
\end{itemize}

The remainder of this paper is organized as follows. Section~\ref{sec:related} discusses contemporary incremental learning methodologies applied to different problems. In Section~\ref{sec:problem} a precise formulation of the incremental learning task for semantic segmentation is introduced. Section~\ref{sec:methodology} outlines the proposed methodologies. The employed datasets and the network training strategies are detailed in Section \ref{sec:training}. The  results on the  Pascal VOC2012 and MSRC-v2 datasets are shown in Section~\ref{sec:results}. 
Conclusion and future developments are presented in Section~\ref{sec:conclusion}. 

%% file: sections/related.tex
\section{Related Work}
\label{sec:related}

\textbf{Semantic segmentation} of images is a dense prediction task where a class label is assigned to each single pixel. It is attracting a large research interest since it is one of the most challenging scene understanding tasks and a huge number of approaches have been proposed so far. Current state-of-the-art approaches are mostly based on the Fully Convolutional Network (FCN) model \citep{long2015fully} and some of the most successful are DRN \citep{yu2017dilated}, PSPNet \citep{zhao2017pyramid} and DeepLab \citep{chen2018deeplab}. Recent reviews on this topic are \cite{garcia2017review, guo2018review, liu2019recent}. However, the current literature lacks the investigation of the incremental learning problem in semantic segmentation, which is addressed in this paper.

\textbf{Incremental learning} is strictly related to other research fields such as continual learning, lifelong learning, transfer learning, multi-task learning and never ending learning. All such tasks require to design an algorithm able to learn new tasks over time without forgetting previously learned ones \citep{parisi2019continual, lesort2019continual}.
The inability to preserve previous knowledge is a critical issue for these approaches, typically referred as \textit{catastrophic forgetting} \citep{mccloskey1989catastrophic, french1999catastrophic, goodfellow2013empirical}, and it still represents one of the main limitations of deep neural networks. The human brain, on the other hand, can efficiently learn new tasks and 
this ability is essential for the deployment of artificial intelligence systems in challenging scenarios where new tasks or classes appear over time. 

Catastrophic forgetting has been faced even before the rise of neural networks popularity \citep{thrun1996learning, polikar2001learn, cauwenberghs2001incremental} and more recently has been rediscovered and tackled in different ways. 
Focusing on deep neural networks, some methods \citep{xiao2014error, roy2018tree} exploit  architectures which grow over time as a tree structure in a hierarchical manner as new classes are observed. 
\cite{istrate2018incremental} propose a method that partitions the original network into sub-networks which are then gradually incorporated in the main one during training.
\cite{sarwar2017incremental} grow the network incrementally over time while sharing portions of the base module. \cite{dai2019incremental} propose a grow-and-prune approach. First, the network grows new connections 
to accommodate new data; then, the connections are pruned on the basis of the magnitude of weights. 

Alternatively, a different strategy consists in freezing or slowing down the learning process in some parts of the network.
\cite{kirkpatrick2017overcoming} developed Elastic Weight Consolidation (EWC) to remember old tasks by slowing down the learning process on the important weights for those tasks. \cite{oquab2014learning} preserve the learned knowledge by freezing the earlier and the mid-level layers of the models. Similar ideas are used in recent studies \citep{li2018learning, istrate2018incremental}. \cite{aljundi2018memory} introduce the idea that when learning a new task, changes to important parameters can be penalized, effectively preventing meaningful knowledge related to previous tasks from being overwritten.

\textbf{Knowledge distillation} is another way of retaining high performance on old tasks which has recently gained wide success. This technique was originally proposed in \cite{hinton2015distilling} {\revision and} \cite{bucilua2006model} to preserve the output of a complex ensemble of networks when adopting a simpler network for more efficient deployment. The idea was  adapted to maintain unchanged the responses of the network on the old tasks whilst updating it with new training samples in different ways. Various approaches have been presented in recent studies \citep{shmelkov2017incremental, rebuffi2017icarl, li2018learning, wu2018incremental, castro2018end, zhou2019M2KD, furlanello2016active}.
\cite{shmelkov2017incremental} propose an end-to-end learning framework where the representation and the classifier are learned jointly without storing any of the original training samples.
\cite{li2018learning} distill previous knowledge directly from the last trained model. 
\cite{dhar2018learning} introduce an attention distillation loss as an information preserving penalty for the classifiers' attention maps. 
In \citep{zhou2019M2KD} the current model distills knowledge from all previous model snapshots, of which a pruned version is saved. 
Deep Model Consolidation \citep{zhang2019class} proposes the idea  to train a separate model for the new classes, and then combine the two models (for old and new data, respectively) via double distillation objective. The two models are consolidated via publicly available unlabeled auxiliary data.
{\revision The Similarity-Preserving Knowledge Distillation (SPKD) \citep{SPKD} strategy aims at preserving the similarities between features of samples of the same class. Recently, novel schemes have been proposed, e.g., to model the information flow through the various layers of the teacher model in order to train a student model to mimic this information flow \citep{passalis2020heterogeneous}. Another strategy is to use network distillation to efficiently compute image embeddings with small networks for metric learning \citep{yu2019learning}.}
In this paper, we thoroughly investigated knowledge distillation and we further adapted it to the semantic segmentation task, while previous works focused on object detection or classification problems.

\textbf{Keeping a small portion of data} belonging to previous tasks is another strategy used by some works 
to preserve the accuracy on old tasks when dealing with new problems \citep{rebuffi2017icarl, lopez2017gradient, chaudhry2018riemannian, hou2018lifelong, castro2018end, tasar2018incremental}. In those works the exemplar set to store is chosen according to different criteria.
In \citep{rebuffi2017icarl, lopez2017gradient} the authors use an episodic memory which stores a subset of the observed examples from previous tasks, while incrementally learning new classes.
\cite{chaudhry2018riemannian} keep a fraction of samples from previous classes to alleviate intransigence of a model, i.e., the inability of a model to update its knowledge. 
\cite{hou2018lifelong} try to balance between preservation and adaptation of the model via distillation and retrospection by caching a small subset of randomly picked data for old tasks.
In \cite{castro2018end}, the classifier and the features used to select the samples for the representative memory are learned jointly in an end-to-end fashion and herding selection \citep{welling2009herding} is used to pick them.
A controlled sampling of memories for replay is proposed by \cite{aljundi2019online}, where samples which are most interfered, i.e., whose prediction will be most negatively impacted by the foreseen parameters update, are chosen.

Another example of this family is the first work on incremental learning for image segmentation \citep{tasar2018incremental}, which however  focused on a very specific setting related to satellite images and has several limitations when applied to generic semantic segmentation problems. Indeed, it considers the segmentation task as a multi-task learning problem, where  a binary classification for each class replaces the multi-class labeling. \cite{tasar2018incremental} store some patches chosen according to an importance value determined by a weight assigned to each class and some other patches chosen at random. 
The capabilities on old classes are preserved by storing a subset of old images.
However, for large amount of classes and different applications, the methodology does not scale properly.
Moreover, storing previously seen data could represent a serious limitation for  applications where privacy issues or limited storage budgets are present.

Generative Adversarial Networks (GANs) have also been used by some recent methods \citep{wu2018incremental, shin2017continual} to generate images containing previous classes instead of storing old classes data, thus retaining high accuracy on old tasks.

%% file: sections/problem.tex
\section{Problem Formulation}
\label{sec:problem}

\subsection{Incremental Learning}
In this section we introduce the task of incremental learning in semantic segmentation  and we present different possible settings in which it can be considered. The incremental learning task when referring to semantic segmentation is defined as the ability of a learning system (e.g., a neural network) to learn the segmentation and the labeling of new classes without forgetting or deteriorating too much the performance on previously learned classes. Typically, in semantic segmentation old and new classes coexist in the same image, and the algorithm needs to account for the accuracy on new classes as well as the accuracy on old ones. The first should be as large as possible in order to learn new classes, while the second  should be as close as possible to the accuracy experienced before the addition of the new classes, thus avoiding catastrophic forgetting. The critical issue is how to find the optimal trade-off between preservation of previous knowledge and capability of learning new tasks. \\
The considered problem is even harder when no data from previous tasks can be preserved, which is the scenario of interest in the majority of the applications where privacy concerns or limited storage requirements subsist. 
Here the most general incremental framework is addressed, in which:
\begin{itemize}
\item previously seen images are not stored nor used;
\item new images may contain examples of unseen classes combined together with pixels belonging to old ones; 
\item  the approach must scale well with respect to the number of classes.
\end{itemize}

Let us assume that the provided dataset $\mathcal{D}$ contains $N$ images. As usual, part of the data is exploited for training and part for testing and we will refer to the training split of $\mathcal{D}$ with the notation $\mathcal{D}^{tr}$.
Each pixel in each image of $\mathcal{D}$ is associated to a unique element of the {\revision set $\mathcal{T}= \lbrace c_0,c_1,c_2,...,c_{C-1}  \rbrace$ of $C$ possible classes.}
In case a background class is present we associate it to the first class $c_0$ because it has a special and non-conventional behavior being present in almost all the images and having by far the largest occurrence among all the classes.

Moving to the incremental learning steps, we assume that we have trained our network to recognize a subset $\mathcal{S}_0 \subset \mathcal{T}$ of \textit{seen} classes using a labeled subset $\mathcal{D}_0^{tr} \subset \mathcal{D}^{tr}$, whose images contain only pixels belonging to the classes in $\mathcal{S}_0$. We then perform some incremental steps $k=1,2,...$ in which we want to recognize a new subset $\mathcal{U}_k \subset \mathcal{T}$ of \textit{unseen} classes in a new set of training steps. 
During the $k$-th incremental step the set of all previously learned classes is denoted as $\mathcal{S}_{k-1}$ and after the current step, the new set $\mathcal{S}_{k}$ will contain also the last added classes. Formally, $\mathcal{S}_k=\mathcal{S}_{k-1} \cup \mathcal{U}_k$ and  $\mathcal{S}_{k-1} \cap \mathcal{U}_k = \emptyset$.
Each step of training involves a new set of samples, i.e., $\mathcal{D}_k^{tr} \subset \mathcal{D}^{tr}$, whose images contain only elements belonging to $\mathcal{S}_{k-1} \cup \mathcal{U}_k$. Notice that this set is disjoint from previously used samples, i.e., $\left( \bigcup_{j=0,...,k-1} \mathcal{D}_j^{tr} \right) \cap \mathcal{D}_k^{tr} = \emptyset$. 
It is important to notice that images in $\mathcal{D}_k^{tr}$ could also contain classes belonging to $\mathcal{S}_{k-1}$, however their occurrence will be  limited since $\mathcal{D}_k^{tr}$ is restricted to consider only images which contain pixels from at least one class belonging to $\mathcal{U}_k$.  The specific occurrence of a particular class belonging to $\mathcal{S}_{k-1}$ is highly correlated to the set of classes being added (i.e., $\mathcal{U}_k$). For example, if we assume that the set of old classes is $\mathcal{S}_{k-1} = \left\lbrace \mathit{car}, \mathit{sofa}   \right\rbrace$ and the set of new classes is $\mathcal{U}_k=\lbrace \mathit{bus} \rbrace$, then it is reasonable to expect that $\mathcal{D}_{k}^{tr}$ contains examples of the class $\mathit{car}$, that  appears in road scenes together with the $\mathit{bus}$, while the  $\mathit{sofa}$ is extremely unlikely to occur.
 
Given this setting, there exist many different ways of sampling the set $\mathcal{U}_k \subset \mathcal{T}$ of unseen classes. Previous work \citep{shmelkov2017incremental, zhang2019class, li2019efficient, michieli2019} sort the classes using the order in the exploited dataset (e.g., alphabetical order) and the first set of results in this paper stick to this assumption to replicate the same scenarios. However, we also present and discuss the order based on the pixels' frequencies of each class inside the dataset: in real world applications, indeed, it is more likely to start from common classes and then introduce rarer ones. 

Additionally, there are many ways of selecting the cardinality of the sets $\mathcal{U}_k$, leading to different incremental scenarios. Starting from the choices considered in \citep{shmelkov2017incremental} for object detection on the Pascal VOC2007 dataset, we consider a wide range of settings and we evaluate them on the Pascal VOC2012 \citep{pascalvoc2012} and MSRC-v2 \citep{MSRC} datasets for semantic segmentation. Namely, as in \citep{shmelkov2017incremental, zhang2019class, michieli2019} we deeply analyze the behavior of our algorithms when adding a single class, a batch of classes and  multiple classes sequentially one after the other.

{\revision
\subsection{Incremental Labeling}
We define another interesting scenario that we  denote with \textit{incremental labeling}. The key  difference with respect to the incremental learning scenario is that  labeling information for pixels belonging to old classes in the samples of the upcoming steps is not present (they are labeled as background). This represents a more challenging protocol as the statistics of the background changes and incorporates samples belonging to known classes which may mislead the model.
In other words, each incremental step comes with a dataset of previously unseen images, whose pixels belong either to old or new classes. However, the ground truth annotated segmentation maps comes with labels for pixels of novel classes only, while old ones are labeled as background.\\
We call this incremental protocol as incremental labeling to underline that new samples come only with the new classes of interest annotated.
}

%% file: sections/methods.tex
\section{Knowledge Distillation for Semantic Segmentation} 
\label{sec:methodology}
In this work, starting from our previous publication  \citep{michieli2019},
we propose a set of  methodologies based on different types of knowledge distillation strategies. 

\subsection{Network Architecture}
\label{sec:netarchi}

The methods proposed in this paper can be fitted into any deep network architecture; however, since most recent architectures for semantic segmentation are based on the auto-encoder scheme, we focus on this representation.
In particular, for the experimental evaluation of the results we use the Deeplab v2 network \citep{chen2018deeplab}, which is a widely used approach with state-of-the art performance.
More in detail, we exploit the Deeplab v2 network with ResNet-101 as the backbone, whose weights were pre-trained \citep{nekrasov}
 on MSCOCO \citep{lin2014microsoft}  {\revision or ImageNet \citep{deng2009imagenet}}. 
 The pre-training of the feature extractor, as done also in other incremental learning works as \cite{li2018learning}, is needed since VOC2012 and MSRC-v2 datasets are too small to be used for training a complex network like the Deeplab v2 from scratch. {\revision  However, MSCOCO data are used only for the initialization of the feature extractor since  the labeling information of this dataset could contain information in some way related to the classes to be learned during the incremental steps. 
 To further evaluate the impact of the pre-training, we also tried a different initialization of the backbone using the ImageNet \citep{deng2009imagenet} dataset for image classification (see Section~\ref{sec:ablation_imagenet} and the Supplementary Material). }
As previously introduced, the Deeplab v2 model is based on an auto-encoder structure (i.e., it consists of an encoder part followed by a decoder phase) where the decoder is composed by Atrous Spatial Pyramid Pooling (ASPP) layers in which multiple atrous convolutions with different rates are applied in parallel on the input feature map and then merged together to enhance the accuracy at multiple scales. The original work exploits also a post-processing step based on Conditional Random Fields, but we removed this module to train the network end-to-end and to measure the performance of the incremental approaches without considering the contribution of post-processing steps not related to the training.

\subsection{Incremental Steps}

\begin{figure*}
\centering
\includegraphics[width=0.9\textwidth]{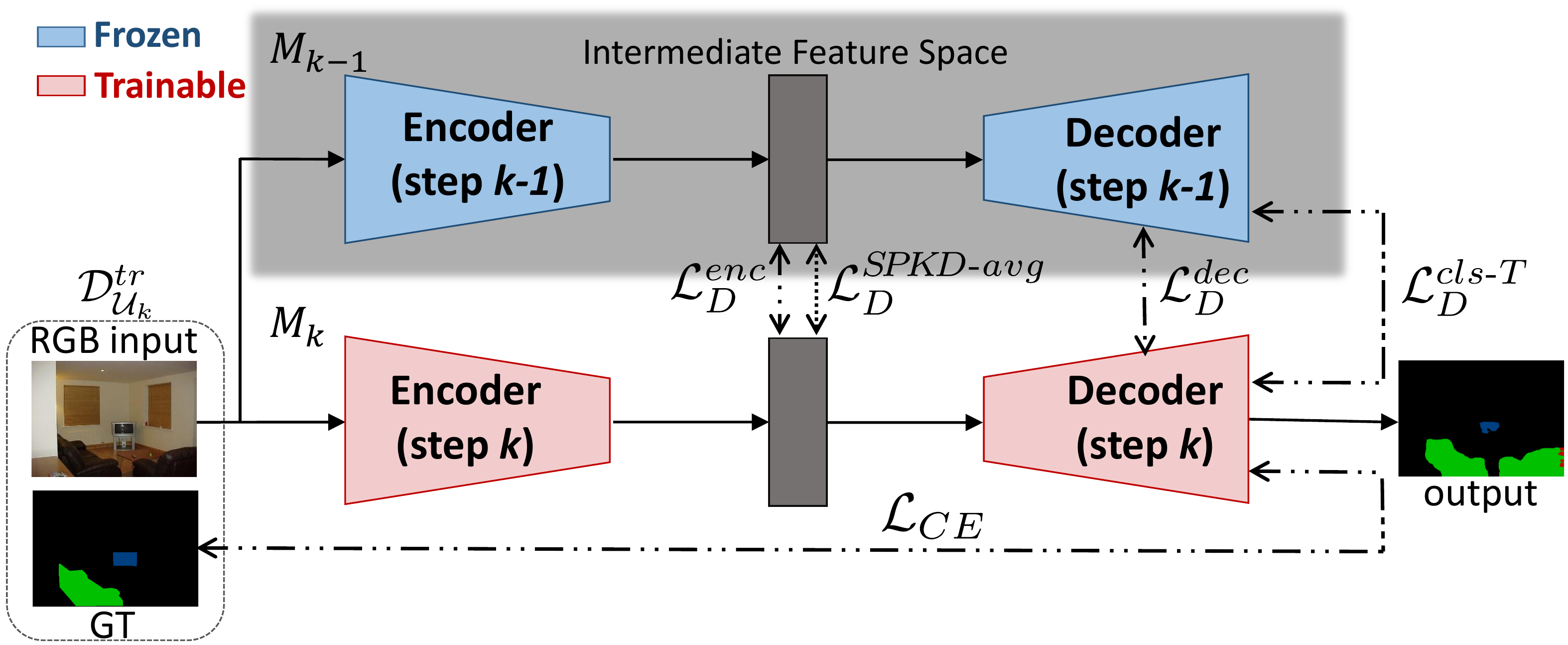}
\vspace{-0.4cm}
\caption{Overview of the $k$-th incremental step of our learning framework for semantic segmentation of RGB images. The scenario in which the current model $M_k$ is completely trainable, i.e. not frozen, is reported. The model $M_{k-1}$, instead, is frozen and is not being updated during the current step.}
\label{fig:architecture}
\end{figure*}

The proposed incremental learning schemes for semantic segmentation are now introduced:  a general overview of the proposed approach is shown in Fig.~\ref{fig:architecture}.
We start by training the chosen network architecture 
to recognize the classes in $\mathcal{S}_0$ with the corresponding training data $\mathcal{D}^{tr}_0$. As detailed in Section~\ref{sec:problem}, $\mathcal{D}^{tr}_0$ contains only images with pixels belonging to classes in $\mathcal{S}_0$. The network is trained in a supervised way with a standard cross-entropy loss. After training, we save the obtained model as $M_0$.

Then, we  perform a set of incremental steps indexed by $k=1,2,...$ to make the model learn every time a new set of classes $\mathcal{U}_k$. 
At the $k$-th incremental step, the current training set $\mathcal{D}_k^{tr}$ is built with images that contain at least one of the new classes  (but they can possibly contain also pixels belonging to previously seen classes
). During step $k$, the model $M_{k-1}$ is loaded and trained exploiting  a linear combination of two losses: a cross-entropy loss $\mathcal{L}_{CE}$, which learns how to identify and label the classes, and a distillation loss $\mathcal{L}_{D}$, which retains knowledge of previously seen classes and will be detailed in the following. 
After the $k$-th incremental step, we save the current model as $M_k$ and we repeat the described procedure every time there is a new set of classes to learn.

The  loss $\mathcal{L}$ used to train the model is defined as:

\begin{equation}
\mathcal{L} = \mathcal{L}_{CE} + \lambda_D \mathcal{L}_D
\end{equation}
where $\mathcal{L}_D \in \left\lbrace\mathcal{L}_D', \mathcal{L}_D'', \mathcal{L}_D''', \mathcal{L}_D''''\right\rbrace$ is one of the various distillation loss models which will be detailed in the following, while $\lambda_D$ is an experimentally tuned parameter balancing the two terms. Setting $\lambda_D=0$ corresponds to the fine-tuning scenario in which no distillation is applied and the cross-entropy loss is applied to both unseen and seen classes. We expect this case to exhibit some sort of catastrophic forgetting, as already pointed out in the literature.

During the $k$-th incremental step, the cross-entropy loss $\mathcal{L}_{CE}$ is applied to all the classes.
It is defined as:

\begin{equation}
\label{eq:CE}
\mathcal{L}_{CE} = 
- \frac{1}{\lvert \mathcal{D}_{k}^{tr} \rvert}  
\sum_{\mathbf{X}_n \in \mathcal{D}_{k}^{tr}} 
\sum_{c \in \mathcal{S}_{k} }
\mathbf{Y}_n [c] \cdot 
\log \left( M_k \left(\mathbf{X}_n \right) [c]   \right)
\end{equation}
where $\mathbf{Y}_n [c]$ and $M_k \left(\mathbf{X}_n \right) [c]$ are respectively the one-hot encoded ground truth and the output of the segmentation network corresponding to the estimated score for class $c$. Note that, since $\mathcal{S}_{k}  = \mathcal{S}_{k-1} \cup \mathcal{U}_k$, the sum is computed on both old and newly added classes, but since new ones are much more likely in $\mathcal{D}_k^{tr}$, there is a clear unbalance toward them leading to catastrophic forgetting \citep{wu2019}.

As regards the distillation loss $\mathcal{L}_{D}$, we focus on losses that only depend only on the previous model $M_{k-1}$ to avoid the need for large storage requirements. 
 
\subsection{Distillation on the Output Layer ($\mathcal{L}_{D}^{cls-T}$)}

The first considered distillation term $\mathcal{L}_D'$ for semantic segmentation is the cross-entropy loss computed on already seen classes between the probabilities produced by the output of the softmax layer 
of the previous model $M_{k-1}$ and the output of the softmax layer of the current model $M_{k}$ (if we assume to be at the $k$-th incremental step). 
Notice that the cross-entropy is computed only on already seen classes, i.e., on classes in $\mathcal{S}_{k-1}$, since we want to guide the learning process to preserve the behavior on such classes. 
The distillation loss is then defined as in \citep{michieli2019}:

\begin{equation}
\label{eq:D1}
\begin{aligned}
\mathcal{L}_{D}^{cls} \! = \!
- \frac{1}{\lvert \mathcal{D}_{k}^{tr} \rvert}  
 \! \sum_{\mathbf{X}_n \in \mathcal{D}_{k}^{tr}}
 \!  \sum_{c \in \mathcal{S}_{k-1}}
\! \! \! M_{k-1}  \left(\mathbf{X}_n \right) [c] \  \!\!\! \cdot  \! 
\log \left( M_k \left(\mathbf{X}_n \right) [c]   \right)
\end{aligned}
\end{equation}
Furthermore, we improve the model 
 by rescaling the logits using a softmax function with temperature $T$ , i.e.: 

\begin{equation}
\label{eq:sigma}
\sigma(z_c) = \frac{\exp( z_c/T )}{\sum_j \exp( z_j/T ) }
\end{equation}

where {\revision $z_c$ is the logit value corresponding to class $c$}.  Hence, denoting with $M_k^T$  the output of the segmentation network for the estimated score of class $c$ after the procedure of Eq.~(\ref{eq:sigma}), we can rewrite Eq.~(\ref{eq:D1}) as: 

\begin{equation}
\label{eq:D1T}
\begin{aligned}
\mathcal{L}_{D}^{cls-T} \! = \!
- \frac{1}{\lvert \mathcal{D}_{k}^{tr} \rvert}  
 \! \sum_{\mathbf{X}_n \in \mathcal{D}_{k}^{tr}}
 \!  \sum_{c \in \mathcal{S}_{k-1}}
\! \! \! M_{k-1}^T  \left(\mathbf{X}_n \right) [c] \  \!\!\! \cdot  \! 
\log \left( M_k^T \left(\mathbf{X}_n \right) [c]   \right)
\end{aligned}
\end{equation}

Intuitively, when $T>1$ the model produces a softer probability distribution over classes thus helping to retain information about the uncertainty of the classification scores \citep{hinton2015distilling, guo2017}. In the experiments we empirically set $T$ ranging from $1$ to $10^3$ depending on the scenario. Temperature scaling was not present in the conference version of the work \citep{michieli2019} and it reveals to be useful especially when one class is added at a time.


\begin{figure*}%
    \centering
    \subfloat[Encoder trainable]{{\includegraphics[trim={0cm 12cm 22.5cm 0cm}, clip, width=0.25\textwidth, valign=b]{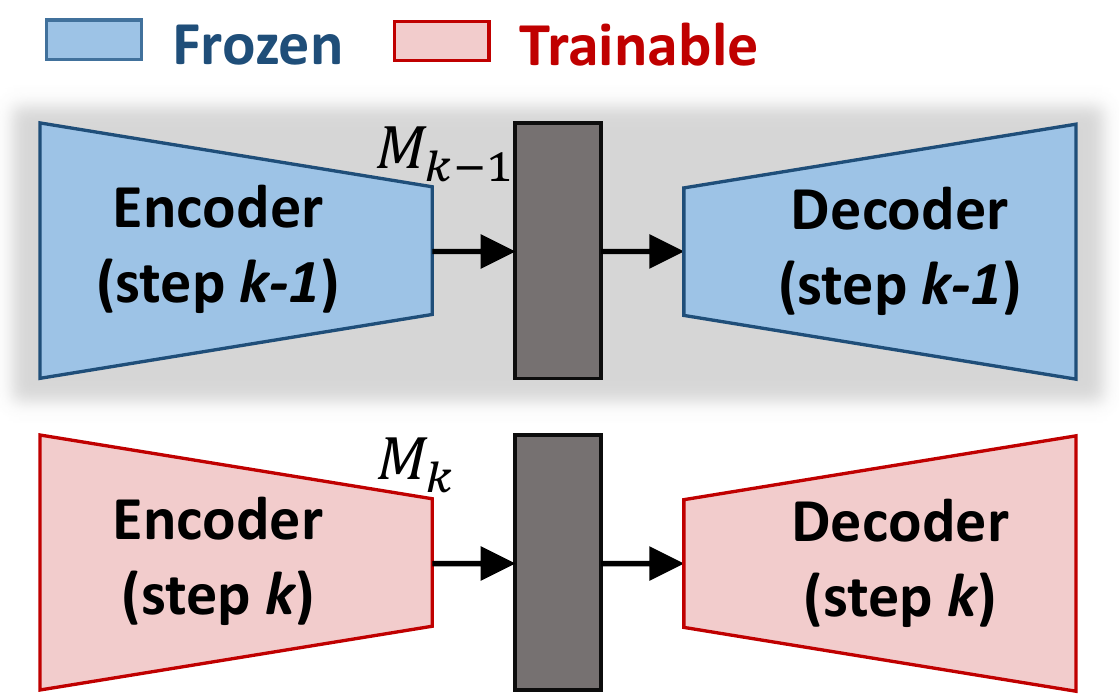} }}%
    \subfloat[Encoder frozen $E_F$]{{\includegraphics[trim={0cm 12cm 22.5cm 0cm}, clip, width=0.25\textwidth, valign=b]{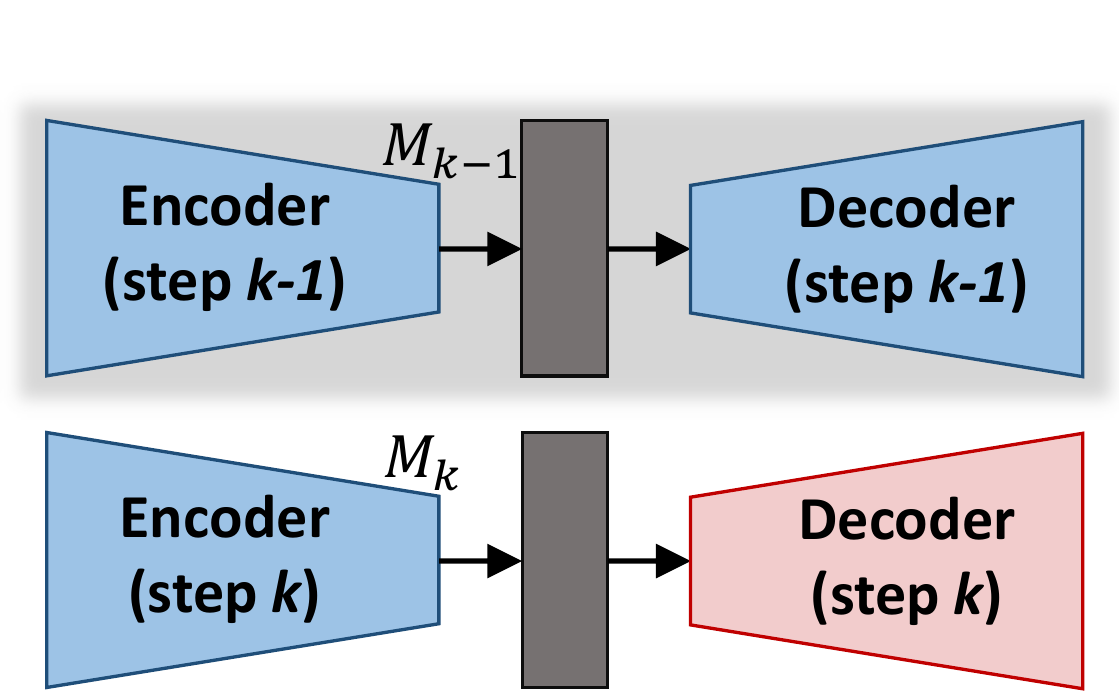} }}%
    \subfloat[Two layers of encoder frozen $E_{2LF}$]{{\includegraphics[trim={0cm 12cm 22.5cm 0cm}, clip, width=0.25\textwidth, valign=b]{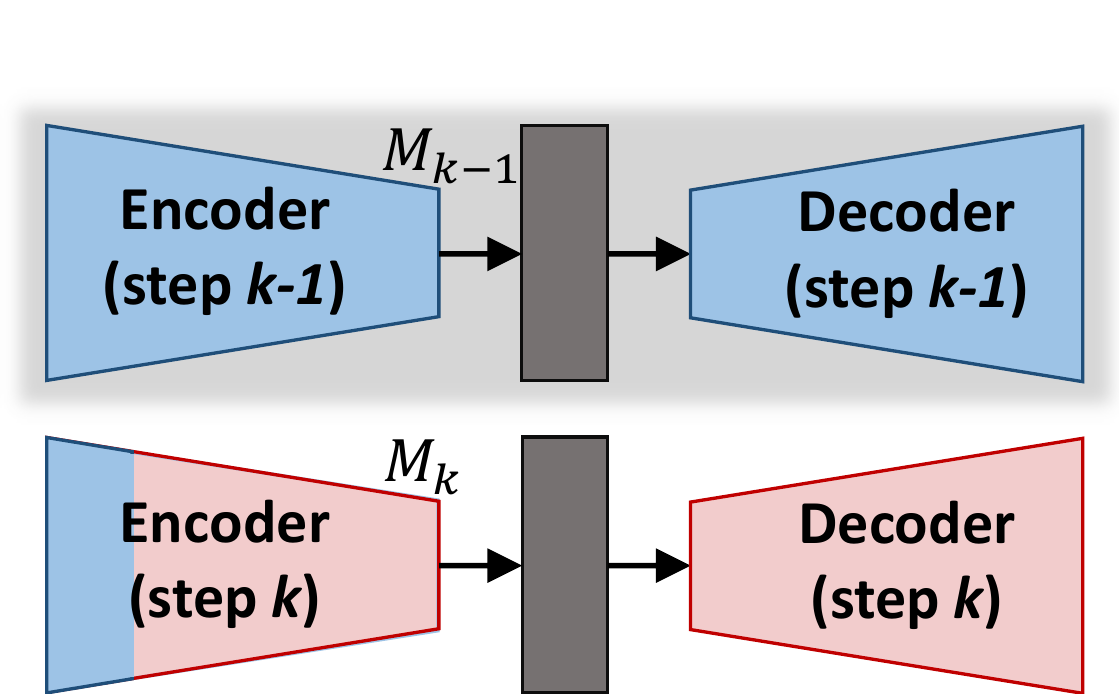} }}%
    \vspace{-0.07cm}
    \caption{Comparison of the different freezing schemes of the encoder at the $k$-th incremental step. The whole model at previous step, i.e. $M_{k-1}$, is always completely frozen and it is employed only for knowledge distillation purposes.}%
    \label{fig:encoders}%
\end{figure*}

When the new task is quite similar with respect to previous ones, the encoder $E$, which aims at extracting some intermediate feature representation from the input information, could be frozen to the status it reached after the initial training phase (we call it $E_F$ in short). In this way, the network is constrained to learn new classes only through the decoder, while preserving the features extraction capabilities unchanged from the training performed on $\mathcal{S}_0$. We evaluated this approach both with and without the application of the distillation loss in Eq.~(\ref{eq:D1}) and Eq.~(\ref{eq:D1T}).
Since the procedure of freezing the whole encoder could appear too restrictive
{\revision (and in fact it does not scale well to the addition of a large number of classes),} we tried also to freeze only the first couple of convolutional layers of the encoder {\revision (i.e., the first two layers of the ResNet 101 network)}. We call this version 
$E_{2LF}$.
Freezing only the first layers allows to preserve the lower level descriptions  
while updating the weights of the task-specific layers of the encoder and of the decoder. A comparison of the different encoder freezing schemes is shown in Fig.~\ref{fig:encoders}.

\subsection{Distillation on the Intermediate Feature Space ($\mathcal{L}_{D}^{enc}$)}

Another approach 
 to preserve the feature extraction capabilities of the encoder  is to apply a distillation loss on the intermediate level corresponding to the output of the encoder $E_k$, i.e., on the features' space before the decoding phase. 
 The distillation function working on the features' space in this case can  no longer be the cross-entropy but rather a geometrical penalty. At that level, indeed, the considered layer is not anymore a classification layer but instead  just an internal stage where the output should be kept close to the previous one in, e.g., Frobenius norm.  
We also considered using L$1$ loss, but we verified empirically that both L$1$  and cross-entropy lead to worse results.
Considering that the network corresponding to model $M_k$ can be decomposed into an encoder $E_k$ and a decoder, the distillation term becomes:

\begin{equation}
\label{eq:D2}
\mathcal{L}_{D}^{enc} =   \frac{1}{\lvert \mathcal{D}_{k}^{tr} \rvert} \sum_{\mathbf{X}_n \in \mathcal{D}_{k}^{tr}}
\Vert  E_{k-1}(\mathbf{X}_n) - E_{k}(\mathbf{X}_n)   \Vert_F^2
\end{equation}
where $E_k(\mathbf{X}_n)$ denotes the features computed by $E_k$ when a generic image $\mathbf{X}_n \in \mathcal{D}_{k}^{tr}$ is fed as input. 

\subsection{Distillation on Dilation Layers ($\mathcal{L}_{D}^{dec}$)}

We also tried to apply geometrical penalties at different points inside the network. In particular, we found that a reliable strategy is to apply the distillation on the four dilation layers contained in the ASPP block of the decoder \citep{chen2018deeplab}. Hence the distillation term becomes:
\begin{equation}
\label{eq:D3}
  \mathcal{L}_{D}^{dec} = \frac{1}{\lvert \mathcal{D}_{k}^{tr} \rvert} \sum_{\mathbf{X}_n \in \mathcal{D}_{k}^{tr}}
\sum_{i=1}^4
\frac{\Vert  d^i_{k-1}(\mathbf{X}_n) - d^i_k(\mathbf{X}_n)   \Vert_F^2}
{4}
\end{equation}
where  $d^i_{k}(\mathbf{X}_n)$ is the output of the dilation layer $d^i_k$ with $i=1,2,3,4$ when $\mathbf{X}_n \in \mathcal{D}_{k}^{tr}$ is fed as input.
 This strategy was not considered in \citep{michieli2019} and proved to be effective in preserving the learned knowledge.
 
{\revision
An ablation study on multi-layer knowledge distillation is presented in the Supplementary Material. We can appreciate how distilling early layers of the decoding stage pushes the results toward distillation of the intermediate features (i.e., close to $\mathcal{L}_{D}^{enc}$), while distilling later layers of the decoding stage pushes the results toward distillation on the output layer (i.e., close to $\mathcal{L}_{D}^{cls-T}$).
}

\subsection{Similarity Preserving Distillation on the Intermediate Feature Space ($\mathcal{L}_{D}^{SPKD}$)}

Finally, we introduce a modified version of the Similarity-Preserving Knowledge Distillation (SPKD) \citep{SPKD} aiming at preserving the similarities between features of samples of the same class.
Let us denote with  $\mathcal{B}$ a  training batch containing  $B$ images, with $W$ and $H$ the (reduced) spatial dimension in the features' space and with $F$ the number of features channels. In the original version of the SPKD approach, the content of the feature layers $E_{k}(\mathcal{B}) \in \mathbb{R}^{B \times H \times W \times F}$ is reshaped as $E'_{k}(\mathcal{B}) \in \mathbb{R}^{B \times HWF}$ and then the matrix $\tilde{A'}_k = E'_{k}(\mathcal{B}) \cdot E'_{k}(\mathcal{B})^{T}  \in \mathbb{R}^{B \times B}$ is computed and row-wise normalized to $A'_k$. 
The SPKD loss  \citep{SPKD} is then computed as:
\begin{equation}
\mathcal{L}_{D}^{SPKD} = \frac{1}{\lvert \mathcal{D}_{k}^{tr} \rvert} \sum_{\mathcal{B} \in \mathcal{D}_{k}^{tr}} \frac{1}{B} ||A'_k-A'_{k-1}||^2_F
\label{eq:SPKD_orig}
\end{equation}
The approach was originally introduced for image classification and is based on the idea that each image contains mostly one object in foreground and is associated to a single label. 
In practice, it does not capture that, in semantic segmentation, multiple classes co-exist in the same image and that an object belonging to a certain class can be a small part of the image.

For this reason, we introduce a variation of this approach. We accumulate the activations over all spatial locations in the feature space, i.e., we compute the matrix $E''_{k}(\mathcal{B} ) \in \mathbb{R}^{B \times F}$ where $E''_{k}[b,f]=\sum_{h=1}^H \sum_{w=1}^W E_{k}[b,h,w,f]$. The loss is then computed as in the previous case but using matrix $E''_{k}$ in place of $E'_{k}$, i.e., $\tilde{A}''_k = E''_{k}(\mathcal{B} ) \cdot E''_{k}(\mathcal{B} )^{T}$ and row-wise normalized to ${A''}_k$. The loss is then computed as:

\begin{equation}
\mathcal{L}_{D}^{SPKD-avg} = \frac{1}{\lvert \mathcal{D}_{k}^{tr} \rvert} \sum_{\mathcal{B} \in \mathcal{D}_{k}^{tr}} \frac{1}{B} ||A''_k-A''_{k-1}||^2_F
\label{eq:SPKD_new}
\end{equation}

This allows to avoid the dependency on the spatial locations of the objects and reduces the computation time due to the smaller matrices size. We verify the validity of this modification  with respect to Eq.~(\ref{eq:SPKD_orig}) in Section~\ref{sec:results}.

A summary of the proposed strategies for the incremental {\revision learning steps}  is shown in Fig.~\ref{fig:architecture}, which points out the four losses.
As a final remark, we also tried a combination of the described distillation losses without achieving significant enhancements. 

%% file: sections/training.tex
\section{Training Procedure}
\label{sec:training}

\subsection{Datasets}

To evaluate the effectiveness and the robustness of the proposed methodologies we choose to employ two publicly available datasets for semantic segmentation: namely, the Pascal VOC2012 \citep{pascalvoc2012} and the MSRC-v2 \citep{MSRC} datasets. 
These benchmarks have been widely used to evaluate semantic segmentation schemes \citep{garcia2017review, csurka2013good}.

The \textbf{Pascal VOC2012} dataset \citep{pascalvoc2012} contains $10582$ variable-sized images in the training split and $1449$ in the validation split. The semantic labeling assigns the pixels to $21$ different classes ($20$ plus the background). Since the test set has not been made available, all the results have been computed on the images belonging to the Pascal VOC2012 validation split (i.e., using the validation split as a test set) as done by most approaches in the literature \citep{shmelkov2017incremental, zhang2019class, michieli2019}.

The \textbf{MSRC-v2} dataset \citep{MSRC} consists of $335$ images in the \textit{trainval} split and $256$ in the test split with variable resolution. It is annotated using $23$ semantic classes, however the \textit{horse} and \textit{mountain} classes have been excluded as suggested  by the dataset creators \citep{MSRC, shotton2008semantic}, because they are underrepresented inside the dataset. The results are computed and reported on the original test set.

For both datasets, we randomly flipped and scaled the images of a random factor between $0.5$ and $1.5$ with bilinear interpolation. For training, random crops of $321\times321$ pixels have been used for memory limitations. The testing phase has been conducted at the original resolution of the images.

\subsection{Implementation Details}

The proposed incremental learning strategies are independent of the backbone architecture and generalize well to different scenarios where new tasks should be learned over time. For the experimental evaluation we select the architecture presented in Section \ref{sec:netarchi}.
We optimize the network weights with Stochastic Gradient Descent (SGD) as done in \cite{chen2018deeplab}. The initial stage of network training  on the set $\mathcal{S}_0$ is performed by setting the starting learning rate to $10^{-4}$ and training for $\lvert \mathcal{S}_0 \rvert \cdot 1000$ steps decreasing the learning rate to $10^{-6}$ with a polynomial decay rule with power $0.9$. 
Notice that the number of training steps is linearly proportional to the number of classes in $\mathcal{S}_0$.
We employ weight decay regularization of $10^{-4}$ and a batch size of $4$ images. 

The incremental training steps $k=1,2,...$ are performed employing a lower learning rate to better preserve previous weights. In this case the learning rate starts from $5 \cdot 10^{-5}$ and decreases to $10^{-6}$ after $\lvert \mathcal{U}_k \rvert \cdot 1000$ steps of polynomial decay. As before, we train the network for a number of steps which is proportional to the number of classes contained in the considered incremental step thus allowing to automatically adapt the training length to the number of new classes being learned.
The considered metrics are the most widely used for semantic segmentation problems: namely, per-class Pixel Accuracy (PA), per-class Intersection over Union (IoU), mean PA (mPA), mean Class Accuracy (mCA) and mean IoU (mIoU) \citep{csurka2013good}.

We use TensorFlow \citep{abadi2016tensorflow} to develop and train the network: the overall training of the considered architecture takes around 5 hours on a NVIDIA 2080 Ti GPU. The code is available online at \url{https://lttm.dei.unipd.it/paper_data/KDSemantic} .

%% file: sections/results.tex
\section{Experimental Results}
\label{sec:results}

Following the experimental scenarios presented in \cite{shmelkov2017incremental} and \cite{michieli2019}, we start by analyzing {\revision the incremental learning task with} the addition of a single class, in alphabetical or frequency-based order, and then move to the addition of 5 and 10 classes, either all together or sequentially. We firstly present the results on the Pascal VOC2012 dataset and then move to the MSRC-v2. {\revision Then, we evaluate the proposed approach in the incremental labeling setting. Finally, we summarize the main achievements of each proposed strategy.}

\subsection{Addition of One Class on VOC2012}
\label{subsec:single}

\begin{table*}[htbp]
\vspace{-0.1cm}
\caption{Per-class IoU of the proposed approaches on VOC2012 when the last class, i.e., the \textit{tv/monitor} class, is added.}
\vspace{-0.2cm}
\label{tab:pascal_0_19_20}
\setlength{\tabcolsep}{1.6pt}
\centering
\footnotesize
\begin{tabular}{|c|cccccccccccccccccccc:c|c|ccc|}
\hline
$M_1 (20)$ & \scriptsize\rotatebox{90}{backgr.} &  \scriptsize\rotatebox{90}{aero} &  \scriptsize\rotatebox{90}{bike} &  \scriptsize\rotatebox{90}{bird} &\scriptsize\rotatebox{90}{boat} & \scriptsize\rotatebox{90}{bottle} & \scriptsize\rotatebox{90}{bus} 
  &\scriptsize\rotatebox{90}{car} & \scriptsize\rotatebox{90}{cat} & \scriptsize\rotatebox{90}{chair} & \scriptsize\rotatebox{90}{cow} & \scriptsize\rotatebox{90}{din. table}& \scriptsize\rotatebox{90}{dog} & \scriptsize\rotatebox{90}{horse} 
  & \scriptsize\rotatebox{90}{mbike} & \scriptsize\rotatebox{90}{person} & \scriptsize\rotatebox{90}{plant} &  \scriptsize\rotatebox{90}{sheep} & \scriptsize\rotatebox{90}{sofa} & \scriptsize\rotatebox{90}{train} & \scriptsize\rotatebox{90}{\textbf{mIoU old}} & \scriptsize\rotatebox{90}{tv} & \scriptsize\rotatebox{90}{\textbf{mIoU}} & \scriptsize\rotatebox{90}{\textbf{mPA}} & \scriptsize\rotatebox{90}{\textbf{mCA}}\\
 \hline

Fine-tuning & 90.2 & 80.8 & 33.3 & 83.1 & 53.7 & 68.2 & 84.6 & 78.0 & 83.2 & 32.1 & 73.4 & 52.6 & 76.6 & 72.7 & 68.8 & 79.8 & 43.8 & 76.5 & 46.5 & 68.4 & 67.3 & 20.1 & 65.1 & 90.7 & 76.5 \\

$E_F$ & 92.7 & 86.2 & 32.6 & 82.9 & 61.7 & 74.6 & 92.9 & 83.1 & 87.7 & 27.4 & 79.4 & 59.0 & 79.4 & 76.9 & 77.2 & 81.2 & 49.6 & 80.8 & 49.3 & 83.4 & 71.9 & 43.3 & 70.5 & 93.2 & 81.4 \\

{\revision $E_{2LF}$} & 92.5 & 85.7 & 32.6 & 81.5 & 60.4 & 74.3 & 93.0 & 83.3 & 87.5 & 26.9 & 79.5 & 59.2 & 78.8 & 76.2 & 77.5 & 81.0 & 49.4 & 80.4 & 49.8 & 83.2 & 71.6 & 45.3 & 70.4 & 93.2 & 81.1 \\

$\mathcal{L}_{D}^{cls}$ {\tiny \citep{michieli2019}} & 92.0 & 83.9 & 37.0 & 84.0 & 58.8 & 70.9 & 90.9 & 82.5 & 86.1 & 32.1 & 72.5 & 51.0 & 79.9 & 72.3 & 77.3 & 80.9 & 45.1 & 78.1 & 45.7 & 79.9 & 70.0 & 35.3 & 68.4 & 92.5 & 79.5 \\

$\mathcal{L}_{D}^{cls-T}$ & 92.6 & 85.7 & 33.4 & 85.3 & 63.1 & 74.0 & 92.6 & 83.0 & 86.4 & 30.4 & 78.1 & 55.0 & 79.1 & 77.8 & 76.4 & 81.7 & 49.7 & 80.2 & 48.5 & 80.4 & 71.7 & 44.4 & 70.4 & 93.2 & 80.1 \\

$E_F$, $\mathcal{L}_{D}^{cls-T}$ & 93.1 & 85.9 & 37.3 & 85.5 & 63.1 & 77.5 & 93.2 & 82.2 & 88.8 & 29.4 & 80.1 & 57.1 & 80.6 & 79.4 & 76.9 & 82.5 & 50.0 & 81.8 & 51.1 & 85.0 & \textbf{73.0} & \textbf{51.9} & \textbf{72.0} & \textbf{93.6} & 82.3 \\

$E_{2LF}$, $\mathcal{L}_{D}^{cls-T}$ & 92.7 & 84.7 & 35.3 & 86.0 & 60.7 & 73.3 & 92.8 & 82.6 & 87.6 & 29.9 & 78.6 & 54.4 & 80.3 & 78.0 & 76.3 & 81.5 & 50.0 & 80.9 & 49.5 & 82.8 & 71.9 & 47.4 & 70.7 & 93.2 & 80.7 \\

$\mathcal{L}_{D}^{enc}$ & 92.9 & 84.8 & 36.4 & 82.6 & 63.5 & 75.0 & 92.2 & 83.6 & 88.3 & 29.5 & 80.3 & 59.6 & 79.7 & 80.2 & 78.9 & 81.2 & 49.7 & 78.9 & 51.0 & 84.1 & 72.6 & 50.6 & 71.6 & 93.4 & \textbf{83.4} \\

$\mathcal{L}_{D}^{dec}$ & 92.2 & 85.4 & 34.3 & 82.4 & 61.6 & 73.4 & 91.7 & 82.7 & 86.4 & 32.4 & 77.2 & 57.4 & 76.3 & 72.6 & 76.1 & 81.1 & 53.7 & 79.2 & 46.1 & 81.5 & 71.2 & 35.6 & 69.5 & 92.6 & 81.5 \\

$E_F$, $\mathcal{L}_{D}^{dec}$ & 92.5 & 84.7 & 33.8 & 80.4 & 60.8 & 76.1 & 91.5 & 82.9 & 87.1 & 29.5 & 78.4 & 58.7 & 76.1 & 73.7 & 78.8 & 81.0 & 51.1 & 78.3 & 48.3 & 84.9 & 71.4 & 42.7 & 70.1 & 93.0 & 82.6 
 \\
 
 $\mathcal{L}_{D}^{SPKD}$ {\tiny \citep{SPKD}} & 92.5 & 83.5 & 35.8 & 84.3 & 60.1 & 71.7 & 88.9 & 83.2 & 87.0 & 32.0 & 79.9 & 57.5 & 78.7 & 78.1 & 77.8 & 81.0 & 50.4 & 80.1 & 49.5 & 78.4 & 71.5 & 43.0 & 70.2 & 92.9 & 82.0
 \\
 
 $\mathcal{L}_{D}^{SPKD-avg}$ &  92.6 & 84.8 & 34.8 & 84.8 & 61.4 & 71.9 & 90.5 & 83.8 & 87.4 & 32.0 & 80.0 & 58.1 & 78.6 & 77.9 & 77.6 & 81.3 & 50.4 & 80.6 & 49.6 & 80.5 & 71.9 & 44.5 & 70.6 & 93.1 & 82.3
 \\\hline


$M_0(0-19)$ & 93.4 & 85.5 & 37.1 & 86.2 & 62.2 & 77.9 & 93.4 & 83.5 & 89.3 & 32.6 & 80.7 & 57.3 & 81.5 & 81.2 & 77.7 & 83.0 & 51.5 & 81.6 & 48.2 & 85.0 & 73.4 & -  & 73.4 & 93.9 & 84.3 \\

$M_0(0-20)$ & 93.4 & 85.4 & 36.7 & 85.7 & 63.3 & 78.7 & 92.7 & 82.4 & 89.7 & 35.4 & 80.9 & 52.9 & 82.4 & 82.0 & 76.8 & 83.6 & 52.3 & 82.4 & 51.1 & 86.4 & 73.7 & 70.5 & 73.6 & 93.9 & 84.2 \\
\hline
\end{tabular}
\end{table*}

 We start from the addition of the last class, in alphabetical order, to our classifier. Specifically, we consider $\mathcal{S}_0 = \lbrace  c_0,c_1,...,c_{19}  \rbrace$ and $\mathcal{U}_1 = \lbrace  c_{20}  \rbrace = \lbrace \mathit{tv \backslash monitor}  \rbrace$. 
  The evaluation {\revision on the VOC2012 validation split is reported in Table~\ref{tab:pascal_0_19_20}}. The table reports the IoU for each single class and the average values of the pixel and class accuracy, while the pixel accuracy for each single class can be found in the \textit{Supplementary Material}.
 The network is firstly optimized on the train split containing samples belonging to any of the classes in $\mathcal{S}_0$, i.e., $\mathcal{D}^{tr}_0$: we indicate with $M_0 (0-19)$ the initial training of the network on $\mathcal{D}^{tr}_0$. 
 The network is then updated exploiting the dataset $\mathcal{D}^{tr}_1$ and the resulting model is referred to as $M_1 (20)$. In this way, we  always specify both the index of the training step and the indexes of the classes added in the considered step. \\
From the first row of Table~\ref{tab:pascal_0_19_20} we can appreciate that adapting the network in the standard way, i.e., without additional provisions, leads to an evident degradation of the performance with a final mIoU of $65.1\%$ {\revision compared to  $73.6\%$ of the reference model $M_0 (0-20)$, where all the $21$ classes are learned at once. This is a  confirmation of the catastrophic forgetting phenomenon in semantic segmentation, even with the addition of just one single class. 
Furthermore, simple strategies like adding a bias or a scaling factor to the logits of the old classes do not allow to solve this issue.
}
The main issue of the na\"ive approach (we call it \textit{fine-tuning} in the tables) is that it tends to predict too frequently the last  class, even when it is not present, 
as proved by the fact that the model has a very high pixel accuracy for the $\mathit{tv/monitor}$ class of $84.3\%$ but a very poor IoU of $20.1\%$ on the same class. This is due to the high number of false positive detections of the considered class which are not taken into account by the pixel accuracy measure. For this reason  semantic segmentation frameworks are commonly ranked {\revision using the mIoU score}. 
On the same class, the proposed methods are all able to outperform the na\"ive approach in terms of IoU by a large margin: the best method achieves a mIoU of $51.9\%$ on the $\mathit{tv/monitor}$ class. 

\begin{table*}[htbp]
\vspace{-0.1cm}
\caption{Per-class IoU of the proposed approaches on VOC2012 when the last class according to the occurrence in the dataset, i.e. the \textit{bottle} class, is added.}
\vspace{-0.2cm}
\label{tab:pascal_0_19_20_occurrences}
\setlength{\tabcolsep}{1.6pt}
\centering
\footnotesize
\begin{tabular}{|c|cccccccccccccccccccc:c|c|ccc|}
\hline
 $M_1 (20)$ & \scriptsize\rotatebox{90}{backgr.} &  \scriptsize\rotatebox{90}{person} &  \scriptsize\rotatebox{90}{cat} &  \scriptsize\rotatebox{90}{dog} &\scriptsize\rotatebox{90}{car} & \scriptsize\rotatebox{90}{train} & \scriptsize\rotatebox{90}{chair} 
  &\scriptsize\rotatebox{90}{bus} & \scriptsize\rotatebox{90}{sofa} & \scriptsize\rotatebox{90}{mbike} & \scriptsize\rotatebox{90}{din. table} & \scriptsize\rotatebox{90}{aero}& \scriptsize\rotatebox{90}{horse} & \scriptsize\rotatebox{90}{bird} 
  & \scriptsize\rotatebox{90}{bike} & \scriptsize\rotatebox{90}{tv} & \scriptsize\rotatebox{90}{boat} &  \scriptsize\rotatebox{90}{plant} & \scriptsize\rotatebox{90}{sheep} & \scriptsize\rotatebox{90}{cow} & \scriptsize\rotatebox{90}{\textbf{mIoU old}} & \scriptsize\rotatebox{90}{bottle} & \scriptsize\rotatebox{90}{\textbf{mIoU}} & \scriptsize\rotatebox{90}{\textbf{mPA}} & \scriptsize\rotatebox{90}{\textbf{mCA}}\\
 \hline
Fine-tuning & 91.9 & 80.7 & 82.2 & 72.3 & 81.7 & 77.9 & 27.2 & 90.2 & 46.9 & 74.5 & 56.1 & 82.4 & 71.8 & 77.9 & 34.9 & 55.8 & 58.7 & 31.0 & 71.9 & 66.9 & 66.6 & 63.8 & 66.5 & 92.4 & 75.8 \\

$E_F$ & 92.6 & 81.8 & 87.8 & 81.5 & 83.5 & 84.1 & 26.4 & 92.3 & 50.6 & 68.5 & 54.6 & 86.1 & 79.3 & 85.9 & 36.6 & 66.6 & 62.3 & 49.6 & 79.2 & 80.0 & 71.5 & 61.9 & 71.0 & 93.3 & 81.4 \\

{\revision $E_{2LF}$} & 92.1 & 82.0 & 85.2 & 77.8 & 82.9 & 80.5 & 23.4 & 90.8 & 48.7 & 75.6 & 54.0 & 81.6 & 75.0 & 77.2 & 34.7 & 60.3 & 57.6 & 32.6 & 75.6 & 70.1 & 67.9 & 64.7 & 67.7 & 92.7 & 76.9 \\

$\mathcal{L}_{D}^{cls-T}$ & 92.9 & 82.7 & 87.9 & 80.0 & 82.3 & 82.5 & 31.7 & 90.5 & 49.3 & 75.7 & 57.0 & 85.2 & 77.9 & 85.5 & 37.3 & 65.2 & 63.7 & 48.3 & 79.3 & 77.4 & 71.6 & 68.2 & 71.5 & 93.4 & 81.2 \\

$E_F$, $\mathcal{L}_{D}^{cls-T}$ & 92.9 & 82.1 & 89.3 & 82.2 & 83.5 & 85.0 & 28.6 & 92.5 & 50.2 & 74.2 & 55.4 & 86.1 & 79.2 & 85.4 & 36.9 & 66.7 & 62.6 & 52.1 & 80.1 & 79.6 & \textbf{72.2} & 64.2 & \textbf{71.8} & \textbf{93.6} & 81.0 \\

$E_{2LF}$, $\mathcal{L}_{D}^{cls-T}$ & 92.9 & 82.6 & 88.2 & 81.3 & 82.4 & 85.3 & 31.4 & 91.5 & 50.1 & 76.0 & 57.0 & 84.8 & 78.0 & 85.7 & 36.9 & 64.9 & 61.8 & 49.3 & 79.9 & 76.8 & 71.8 & \textbf{69.0} & 71.7 & 93.5 & \textbf{81.8} \\

$\mathcal{L}_{D}^{enc}$ & 92.9 & 81.7 & 88.5 & 81.8 & 83.8 & 85.0 & 27.2 & 92.4 & 51.8 & 73.0 & 56.0 & 85.9 & 79.9 & 85.7 & 37.0 & 65.7 & 61.7 & 48.7 & 80.1 & 80.0 & 71.9 & 62.3 & 71.5 & 93.5 & \textbf{81.8} \\

$\mathcal{L}_{D}^{dec}$ & 92.5 & 82.7 & 86.5 & 79.7 & 83.4 & 83.1 & 28.4 & 91.9 & 46.6 & 68.7 & 54.7 & 83.3 & 75.4 & 83.8 & 32.8 & 65.8 & 62.8 & 48.2 & 78.6 & 73.8 & 70.1 & 67.5 & 70.0 & 93.1 & 78.9 \\
 
 $\mathcal{L}_{D}^{SPKD-avg}$ & 92.7 & 82.0 & 86.8 & 79.3 & 83.1 & 82.5 & 30.4 & 91.7 & 48.3 & 74.6 & 55.7 & 84.8 & 77.3 & 84.8 & 36.0 & 66.8 & 62.2 & 49.0 & 78.5 & 76.4 & 71.1 & 67.6 & 71.0 & 93.3 & 81.0
 \\ \hline

$M_0(0-19)$ & 93.5 & 80.9 & 89.7 & 82.8 & 84.4 & 85.5 & 33.1 & 92.5 & 47.6 & 79.3 & 57.0 & 85.9 & 79.9 & 85.9 & 37.2 & 67.8 & 62.5 & 53.4 & 80.5 & 79.7 & 73.0 & - & 73.0 & 94.0 & 83.3 \\

$M_0(0-20)$ & 93.4 & 83.6 & 89.7 & 82.4 & 82.4 & 86.4 & 35.4 & 92.7 & 51.1 & 76.8 & 52.9 & 85.4 & 82.0 & 85.7 & 36.7 & 70.5 & 63.3 & 52.3 & 82.4 & 80.9 & {\revision 73.3} & 78.7 & 73.6 & 93.9 & 84.2 \\
\hline
\end{tabular}
\end{table*}

Knowledge distillation strategies and the procedure of freezing (part of) the encoder provide better results because they act as regularization constraints. Interestingly, those procedures allow to achieve higher accuracy not only on previously learned classes but also on newly added ones, which might be unexpected if we do not consider the regularization behavior of those terms. 
Hence all the proposed strategies allow to alleviate forgetting and all of them overcome the standard approach (without knowledge distillation) in any of the considered metrics, as can be verified in Table~\ref{tab:pascal_0_19_20}. 
We can appreciate that $\mathcal{L}_{D}'^{cls-T}$ alone is able to improve the average mIoU by $5.3\%$ with respect to the standard case. Notice how the improved version of $\mathcal{L}_{D}^{cls-T}$ with temperature scaling introduced in this work achieves a significant improvement of $2\%$ of mIoU with respect to the conference version of the work \citep{michieli2019}, that corresponds to $\mathcal{L}_{D}^{cls}$. Furthermore, it leads to a much better IoU on the new class, greatly reducing the aforementioned false positives issue. If we completely freeze the encoder $E$ without applying knowledge distillation the model improves the mIoU by $5.4\%$. 
If we combine the two mentioned approaches, i.e., we freeze $E$ and we apply $\mathcal{L}_{D}^{cls-T}$, the mIoU reaches $72.0\%$, with an overall improvement of $6.9\%$, higher than each of the two methods alone (also the performance on the new class is higher). If we just freeze the first two layers of the encoder  and we apply knowledge distillation, i.e., $M_1(20) [ E_{2LF},\mathcal{L}_{D}^{cls-T} ]$ (we use the square brackets to collect the list of employed strategies), a slightly lower mIoU of $70.7\%$ is achieved. Instead, if we apply an L$2$ loss at the intermediate features space ($\mathcal{L}_{D}^{enc}$) the model achieves $71.6\%$ of mIoU, which is $6.5\%$ higher than the standard approach. It is noticeable that two completely different approaches  to preserve knowledge, 
 namely $M_1(20) [ E_F,\mathcal{L}_{D}^{cls-T} ]$ (which applies a cross-entropy between the outputs with encoder frozen) and $M_1(20)  [ \mathcal{L}_{D}^{enc}  ]$ (which applies an L$2$-loss between features spaces), achieve similar and high results both on new and old classes. 
 
If we apply an L$2$ loss on the dilation filters of the decoder, i.e., the $\mathcal{L}_{D}^{dec}$ loss, we obtain a mIoU of $69.5\%$ which is higher than the standard approach but lower than the other strategies. Freezing the encoder yields in this setting to a small improvement from $69.5\%$ to $70.1\%$. Finally, the original SPKD loss achieves $70.2\%$ of mIoU, while the improved version presented in this paper ($M_1(20) [\mathcal{L}_D^{SPKD-avg}] $) achieves a mIoU of $70.6\%$.

{ \revision From the class-wise results we can appreciate that changes in performance on previously seen classes are correlated with the class being added. Some classes have even higher results in terms of mIoU  because their prediction has been reinforced through the new training set.
For example, objects of the classes $\mathit{sofa}$ or $\mathit{dining\ table}$ are typically present in scenes containing a $\mathit{tv/monitor}$, hence in the considered scenario they achieve almost always higher accuracy. 
Other classes, instead, get more easily lost because they represent uncorrelated objects not present inside the new set of samples. 
For example, instances of $\mathit{bird}$ or $\mathit{horse}$ are not present in indoor scenes typically associated  with the $tv/monitor$ class being added.}

\newcommand{\sizefiggg}{0.12}
\begin{figure*}[htbp]{}
\setlength\tabcolsep{1.5pt} 
\centering
\subfloat{
\begin{tabular}{cccccccc}
  {$\scriptstyle RGB$} &  {$\scriptstyle GT$} & {$\scriptstyle Fine-tuning$} &
{$ \scriptstyle M_1(20) [ E_F , \mathcal{L}_{D}^{cls-T} ]$}&
   {$\scriptstyle M_1(20) [ \mathcal{L}_{D}^{enc}]$}&
   {$\scriptstyle M_1(20) [ \mathcal{L}_{D}^{dec}]$}&
   {$\scriptstyle M_1(20) [ \mathcal{L}_{D}^{SPKD-avg}]$}&
 {$\scriptstyle M_0(0-20)$}\\
 
   \includegraphics[width=\sizefiggg\linewidth]{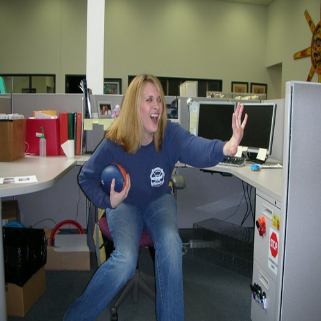} &
  \includegraphics[width=\sizefiggg\linewidth]{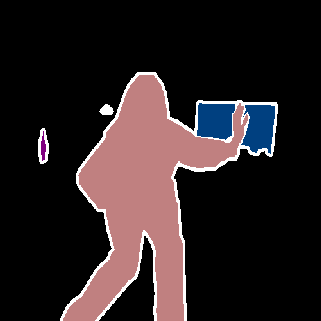} &
  \includegraphics[width=\sizefiggg\linewidth]{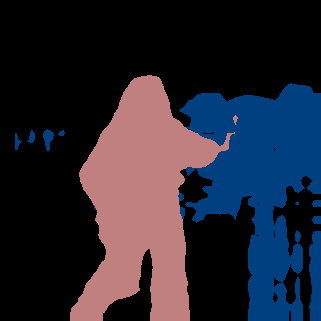} &
  \includegraphics[width=\sizefiggg\linewidth]{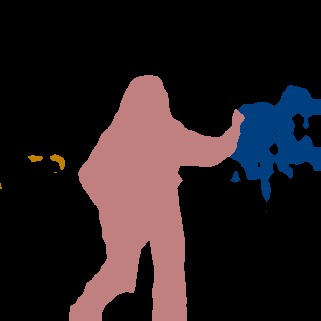} &
  \includegraphics[width=\sizefiggg\linewidth]{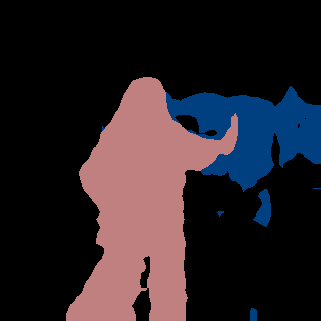} &
  \includegraphics[width=\sizefiggg\linewidth]{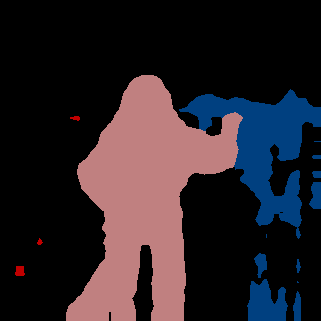} &
  \includegraphics[width=\sizefiggg\linewidth]{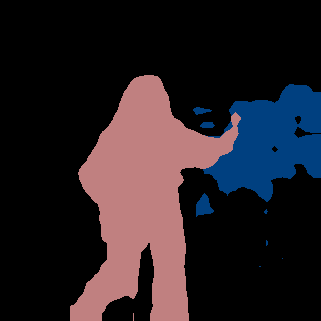} &
  \includegraphics[width=\sizefiggg\linewidth]{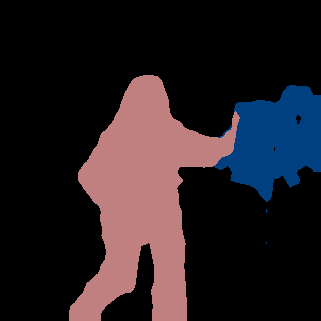} \\
  
    \includegraphics[width=\sizefiggg\linewidth]{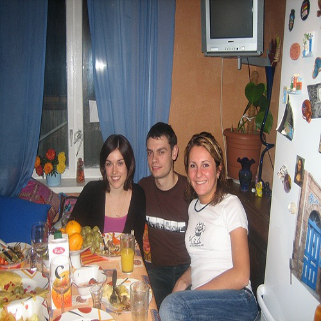} &
  \includegraphics[width=\sizefiggg\linewidth]{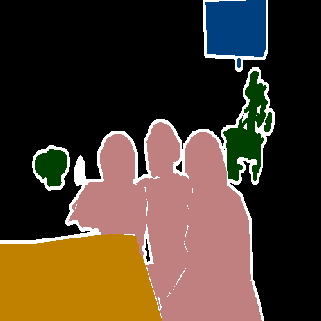} &
  \includegraphics[width=\sizefiggg\linewidth]{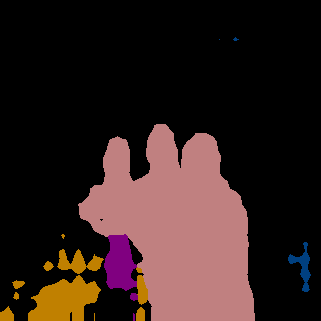} &
  \includegraphics[width=\sizefiggg\linewidth]{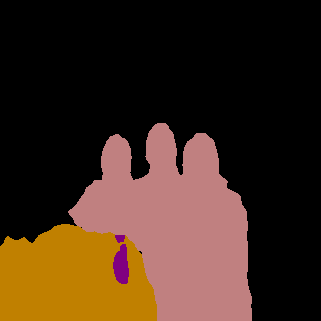} &
  \includegraphics[width=\sizefiggg\linewidth]{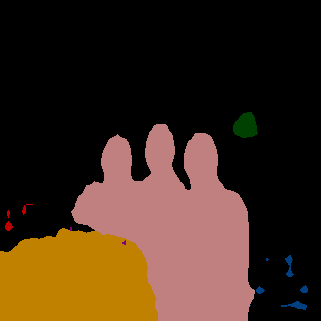} &
  \includegraphics[width=\sizefiggg\linewidth]{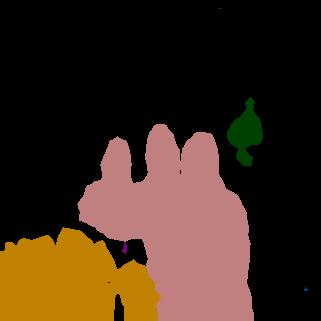} &
  \includegraphics[width=\sizefiggg\linewidth]{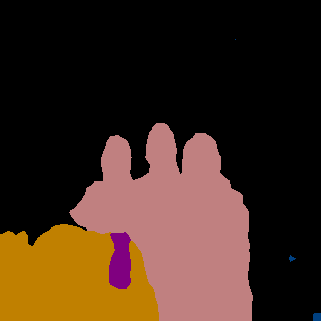}&
  \includegraphics[width=\sizefiggg\linewidth]{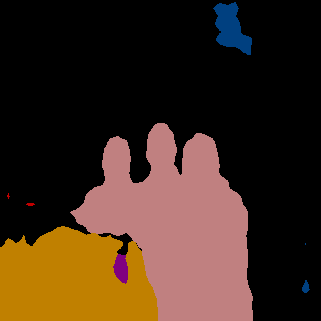} \\

    \includegraphics[width=\sizefiggg\linewidth]{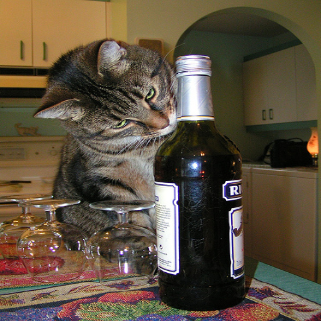} &
  \includegraphics[width=\sizefiggg\linewidth]{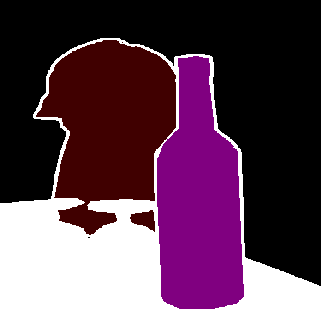} &
  \includegraphics[width=\sizefiggg\linewidth]{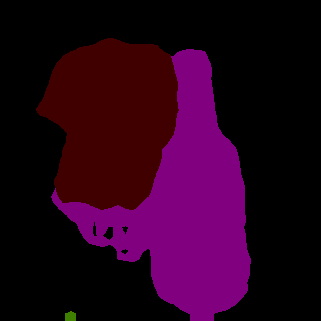} &
  \includegraphics[width=\sizefiggg\linewidth]{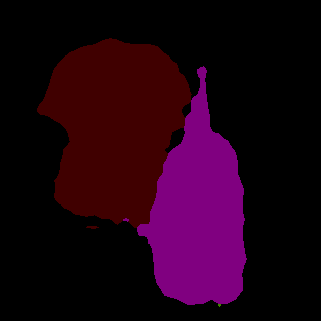} &
  \includegraphics[width=\sizefiggg\linewidth]{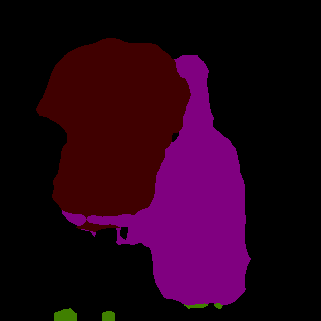} &
  \includegraphics[width=\sizefiggg\linewidth]{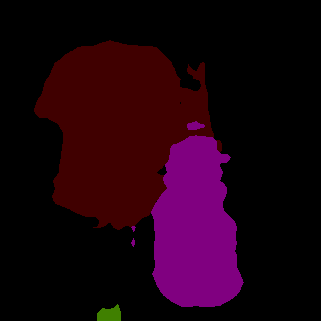}&
  \includegraphics[width=\sizefiggg\linewidth]{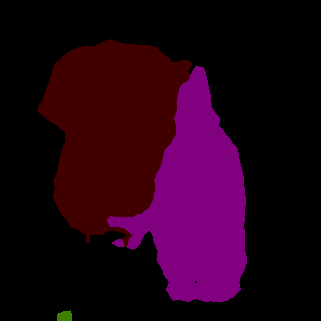}&
  \includegraphics[width=\sizefiggg\linewidth]{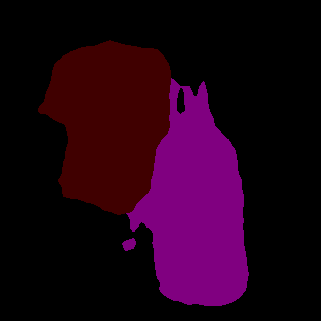} \\

 \end{tabular}
}
\centering
\vspace{-0.65cm}
\subfloat{
\hspace{0.27cm}
\resizebox{8cm}{!}{%
\begin{tabular}{ccccccccc}
& & & & & & & & \\ \cline{9-9}
\cellcolor[HTML]{000000}{\color[HTML]{FFFFFF} \textbf{background}} & 
\cellcolor[HTML]{800080}{\color[HTML]{FFFFFF} \textbf{bottle}} & 
\cellcolor[HTML]{400000}{\color[HTML]{FFFFFF} \textbf{cat}} & 
\cellcolor[HTML]{C00000}{\color[HTML]{FFFFFF} \textbf{chair}} & 
\cellcolor[HTML]{C08000}\textbf{dining table} & 
\cellcolor[HTML]{C08080}\textbf{person} & 
\cellcolor[HTML]{004000}{\color[HTML]{FFFFFF} \textbf{plant}} & 
\multicolumn{1}{c|}{\cellcolor[HTML]{004080}{\color[HTML]{FFFFFF} \textbf{tv}}} &
\multicolumn{1}{c|}{\textbf{unlabeled}}\\ \cline{9-9} 
\end{tabular}%
}
}%
\vspace{-0.1cm}
\caption{Qualitative results on sample scenes for the addition of one class (\textit{best viewed in colors}). In the first two rows the \textit{tv/monitor} class is added, in the last row the \textit{bottle} class is added.}
\label{fig:addition_of_1}
\end{figure*}

{ \revision Some visual examples  are shown in the first two rows of Fig.~\ref{fig:addition_of_1}.} 
We can visually appreciate that knowledge distillation and encoder freezing help in preserving previous classes (e.g., the hand of the $\mathit{person}$ in the first row and the $\mathit{table}$ in the second are better preserved). 
At the same time it does not compromise the learning of the new class (e.g., the $\mathit{tv/monitor}$ in the first row is quite well-localized). In the second row, however, a challenging example is reported where none of the proposed methodologies, and neither the baseline approach, are able to accurately detect the new class.

In Table~\ref{tab:pascal_0_19_20_occurrences} the IoU results in the same scenario are shown, but this time ordering the classes according to the pixels' occurrence of each class, thus the $\mathit{bottle}$ class is added last. 
{ \revision Similarly to the previous case the baseline approach exhibits a large drop in performance while 
 knowledge distillation always helps in every scenario.
 As before, the  best performing strategy is $ M_1(20) [ E_F$, $\mathcal{L}_{D}^{cls-T} ]$.} A visual example is shown in the last row of Fig.~\ref{fig:addition_of_1}: we can verify that knowledge distillation and encoder freezing help not only to retain previously seen classes (e.g., the $\mathit{cat}$ in the example), but also to better detect and localize the new class, i.e. the $\mathit{bottle}$, thus acting as a regularization term.

\subsection{Addition of Multiple Classes on VOC2012}

{ \revision  In this section we consider a more challenging scenario where the initial training is followed by one incremental step to learn multiple classes. }

\begin{figure*}[htbp]{}

\setlength\tabcolsep{1.5pt} 
\subfloat{
\begin{tabular}{cccccccc} 
 {$\scriptstyle RGB$} &
   {$ \scriptstyle GT$} &
   {$\scriptstyle Fine-tuning$} &
  {$\scriptstyle M_1\!(16\!-\!20) \! [ E_F\! , \mathcal{L}_{D}^{cls-T}]\!$}&
  {$\scriptstyle M_1(16-20) [\mathcal{L}_{D}^{cls-T}]$}&
  {$\scriptstyle M_1(16-20) [\mathcal{L}_{D}^{dec}]$}&
  {$\scriptstyle M_1(16-20) [\mathcal{L}_{D}^{SPKD-avg}]$}&
 {$\scriptstyle M_0(0-20)$}\\

   \includegraphics[width=\sizefiggg\linewidth]{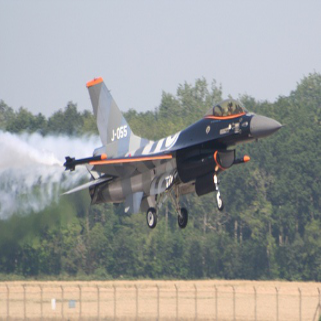} &
  \includegraphics[width=\sizefiggg\linewidth]{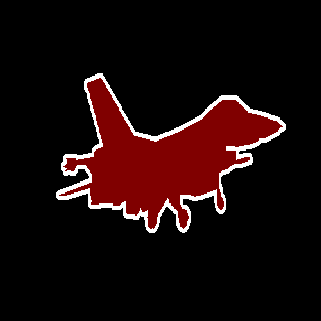} &
  \includegraphics[width=\sizefiggg\linewidth]{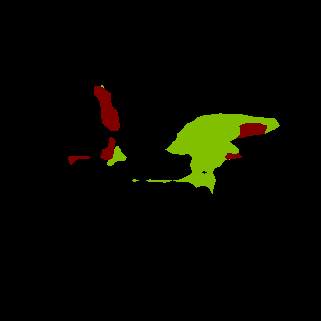} &
    \includegraphics[width=\sizefiggg\linewidth]{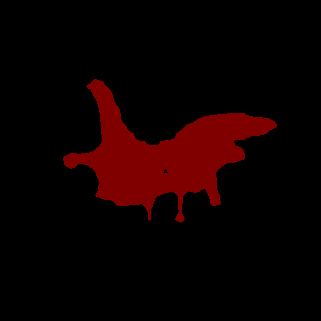} &
  \includegraphics[width=\sizefiggg\linewidth]{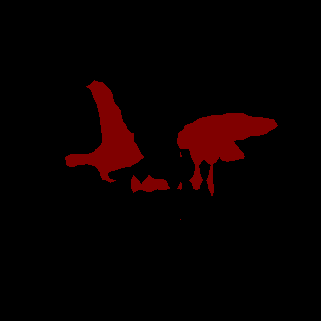} &
  \includegraphics[width=\sizefiggg\linewidth]{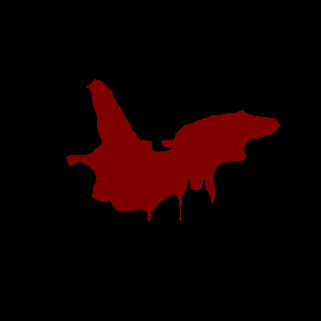} &
  \includegraphics[width=\sizefiggg\linewidth]{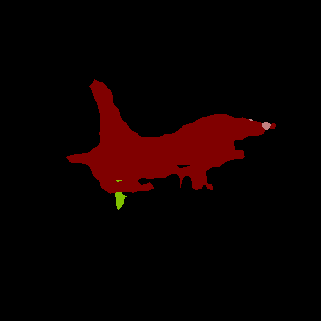} &
    \includegraphics[width=\sizefiggg\linewidth]{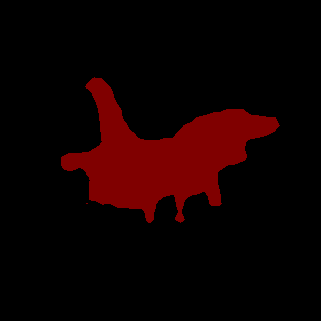} \\

        \includegraphics[width=\sizefiggg\linewidth]{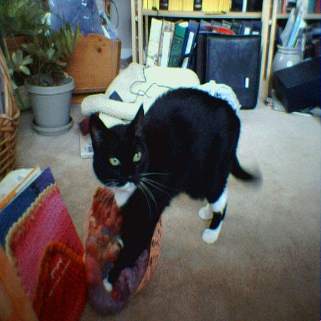}&
    \includegraphics[width=\sizefiggg\linewidth]{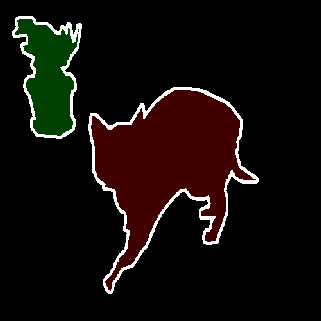} &
    \includegraphics[width=\sizefiggg\linewidth]{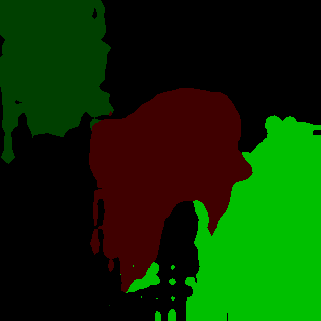}  &
      \includegraphics[width=\sizefiggg\linewidth]{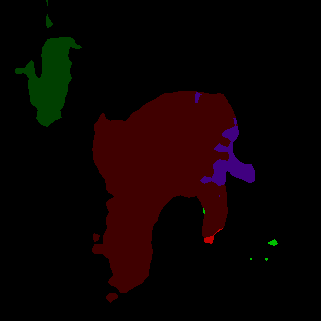} &
  \includegraphics[width=\sizefiggg\linewidth]{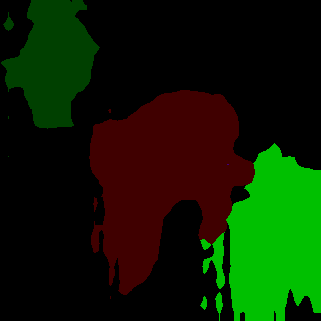} &
    \includegraphics[width=\sizefiggg\linewidth]{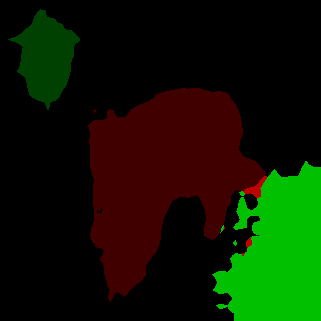} &
  \includegraphics[width=\sizefiggg\linewidth]{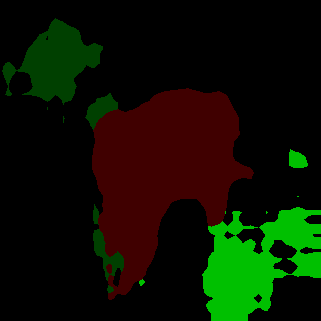} &
\includegraphics[width=\sizefiggg\linewidth]{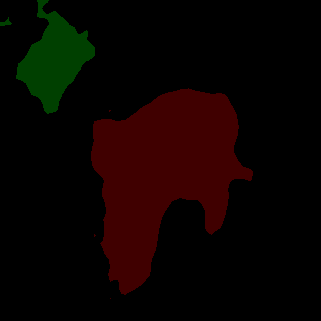} \\

        \includegraphics[width=\sizefiggg\linewidth]{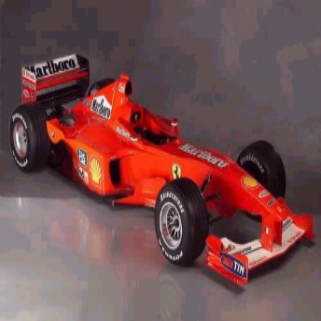} &
    \includegraphics[width=\sizefiggg\linewidth]{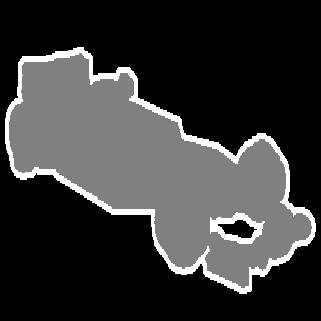}  &
  \includegraphics[width=\sizefiggg\linewidth]{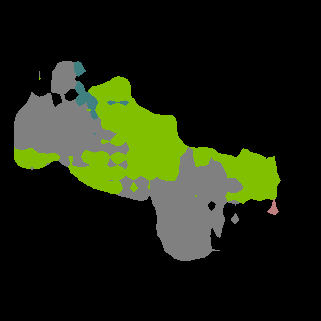} &
  \includegraphics[width=\sizefiggg\linewidth]{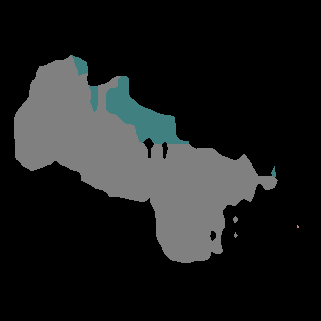}  &
  \includegraphics[width=\sizefiggg\linewidth]{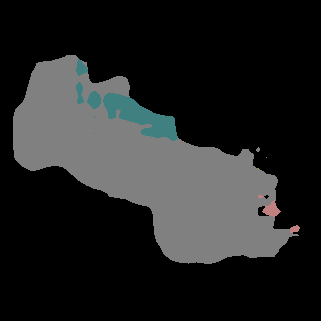} &
    \includegraphics[width=\sizefiggg\linewidth]{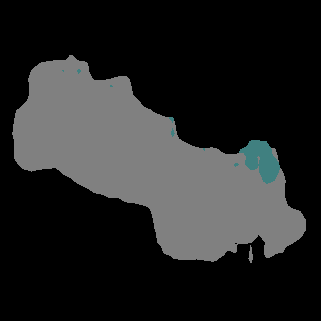} &
  \includegraphics[width=\sizefiggg\linewidth]{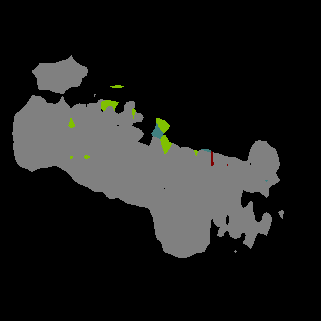} &
\includegraphics[width=\sizefiggg\linewidth]{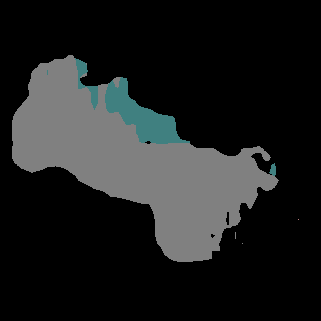} \\

 \end{tabular}
}
\centering
\vspace{-0.58cm}
\subfloat{
\hspace{0.3cm}
\resizebox{8cm}{!}{%
\begin{tabular}{cccccccccccc}
& & & & & & & & \\ \cline{12-12}
\cellcolor[HTML]{000000}{\color[HTML]{FFFFFF} \textbf{background}} & 
\cellcolor[HTML]{800000}{\color[HTML]{FFFFFF} \textbf{aeroplane}} & 
\cellcolor[HTML]{008080}{\color[HTML]{FFFFFF} \textbf{bus}} & 
\cellcolor[HTML]{808080}{\color[HTML]{FFFFFF} \textbf{car}} & 
\cellcolor[HTML]{400000}{\color[HTML]{FFFFFF} \textbf{cat}} & 
\cellcolor[HTML]{C00000}{\color[HTML]{FFFFFF} \textbf{chair}} & 
\cellcolor[HTML]{400080}{\color[HTML]{FFFFFF} \textbf{dog}} & 
\cellcolor[HTML]{C08080}\textbf{person} & 
\cellcolor[HTML]{004000}{\color[HTML]{FFFFFF} \textbf{plant}} & 
\cellcolor[HTML]{00C000}\textbf{sofa} & 
\multicolumn{1}{c|}{\cellcolor[HTML]{80C000}{\color[HTML]{000000} \textbf{train}}} &
\multicolumn{1}{c|}{\textbf{unlabeled}}\\ \cline{12-12} 

\end{tabular}%
}
}%
\vspace{-0.1cm}
\caption{Qualitative results on sample scenes for the addition of 5 classes at once (\textit{best viewed in colors}). The set of new classes is \textit{plant}, \textit{sheep}, \textit{sofa}, \textit{train} and \textit{tv}}
\label{fig:addition_of_5_at_once}
\end{figure*}

First, the addition of the last $5$ classes at once (referred to as $15-20$) is discussed and the results are shown in Table~\ref{tab:pascal_0_15_20}.
{ \revision Results are much lower than in the previous cases 
since there is a larger amount of information to be learned.
The baseline exhibits an even larger drop in accuracy because it tends to overestimate the presence of the new classes 
as shown by the IoU scores of the newly added classes which are often lower 
 by a large margin (see Table~\ref{tab:pascal_0_15_20}), while, on the other side, the pixel accuracy of the new classes is much higher. In this case, $E_F$ and $E_{2LF}$ fail to accommodate the substantial changes in the input distribution.} 
 More details on the per-class pixel accuracy are {\revision shown} in the \textit{Supplementary Material.} 
In this case the distillation on the output layer, i.e., $M_1(16-20) [ \mathcal{L}_{D}^{cls-T}]$, achieves the highest accuracy. Here, the approaches based on $\mathcal{L}_{D}^{cls-T}$ outperform the other ones (even on new classes). Also in this scenario, all the proposals outperform the standard approach on both old and new classes.
Interestingly, some previously seen classes exhibit a clear catastrophic forgetting phenomenon because the updated models mislead them with visually similar classes belonging to the set of new classes. For example, the $\mathit{cow}$ and $\mathit{chair}$ classes are often misled (low IoU and low PA for these classes) with the newly added classes $\mathit{sheep}$ and $\mathit{sofa}$ that have similar shapes (low IoU but high PA for them).\\
Qualitative results are shown in Fig.~\ref{fig:addition_of_5_at_once}:  we can appreciate that the na\"ive approach tends to overestimate the presence of the new classes in spite of previously learned ones or in spite of the background. This can be seen from the figure, where instances of $\mathit{train}$, $\mathit{bus}$ and $\mathit{sofa}$ classes (which are added during the incremental step) are erroneously predicted in place of the new class or in the $\mathit{background}$ region. These classes are correctly removed or strongly reduced applying the proposed strategies even if there is not a clear winner.
On the \textit{aeroplane} in the first row all the proposed approaches work well. Applying $\mathcal{L}_D^{cls-T}$ and freezing the encoder removes the false detection of the $\mathit{sofa}$ in row 2 that is challenging for the other approaches. The \textit{car} in the last row is better detected using the $\mathcal{L}_D^{SPKD-avg}$ loss. 


\begin{table*}[htbp]
\vspace{-0.1cm}
\caption{Per-class IoU of the proposed approaches on VOC2012 when $5$ classes are added at once.}
\vspace{-0.2cm}
\label{tab:pascal_0_15_20}
\setlength{\tabcolsep}{1.6pt}
\centering
\footnotesize
\begin{tabular}{|c|cccccccccccccccc:c|ccccc:c|ccc|}
\hline
 $M_1 (16-20)$ & \scriptsize\rotatebox{90}{backgr.} &  \scriptsize\rotatebox{90}{aero} &  \scriptsize\rotatebox{90}{bike} &  \scriptsize\rotatebox{90}{bird} &\scriptsize\rotatebox{90}{boat} & \scriptsize\rotatebox{90}{bottle} & \scriptsize\rotatebox{90}{bus} 
  &\scriptsize\rotatebox{90}{car} & \scriptsize\rotatebox{90}{cat} & \scriptsize\rotatebox{90}{chair} & \scriptsize\rotatebox{90}{cow} & \scriptsize\rotatebox{90}{din. table}& \scriptsize\rotatebox{90}{dog} & \scriptsize\rotatebox{90}{horse} 
  & \scriptsize\rotatebox{90}{mbike} & \scriptsize\rotatebox{90}{person} & \scriptsize\rotatebox{90}{\textbf{mIoU old}} & \scriptsize\rotatebox{90}{plant} &  \scriptsize\rotatebox{90}{sheep} & \scriptsize\rotatebox{90}{sofa} & \scriptsize\rotatebox{90}{train} & \scriptsize\rotatebox{90}{tv} & \scriptsize\rotatebox{90}{\textbf{mIoU new}} & \scriptsize\rotatebox{90}{\textbf{mIoU}} & \scriptsize\rotatebox{90}{\textbf{mPA}} & \scriptsize\rotatebox{90}{\textbf{mCA}}\\
 \hline

Fine-tuning & 89.7 & 59.5 & 34.6 & 68.2 & 58.1 & 58.8 & 59.2 & 79.2 & 80.2 & 30.0 & 12.7 & 51.0 & 72.5 & 61.7 & 74.4 & 79.4 & 60.6 & 36.4 & 32.4 & 27.2 & 55.2 & 42.4 & 38.7 & 55.4 & 88.4 & 70.6 \\

{\revision $E_F$} & 90.2 & 72.4 & 32.4 & 57.3 & 50.5 & 68.0 & 10.5 & 81.5 & 84.7 & 25.4 & 10.6 & 57.4 & 76.2 & 68.6 & 77.7 & 79.8 & 58.9 & 36.2 & 30.7 & 30.3 & 43.4 & 54.5 & 39.0 & 54.2 & 88.5 & 68.2 \\

{\revision $E_{2LF}$} & 89.6 & 59.2 & 35.2 & 67.8 & 56.2 & 58.0 & 59.3 & 80.9 & 81.6 & 31.3 & 13.4 & 54.9 & 73.3 & 63.3 & 73.0 & 80.0 & 61.1 & 37.5 & 31.9 & 25.8 & 56.2 & 41.4 & 38.5 & 55.7 & 88.4 & 70.4 \\

$\mathcal{L}_{D}^{cls-T}$ & 91.4 & 85.0 & 35.6 & 84.8 & 61.8 & 70.5 & 85.6 & 77.9 & 83.7 & 30.7 & 72.3 & 45.3 & 76.2 & 76.9 & 77.0 & 81.3 & \textbf{71.0} & 33.8 & 55.2 & 30.9 & 73.9 & 51.6 & \textbf{49.1} & \textbf{65.8} & \textbf{91.6} & \textbf{78.1}\\

$E_F$, $\mathcal{L}_{D}^{cls-T}$ & 91.7 & 83.4 & 35.6 & 78.7 & 60.9 & 73.0 & 65.8 & 82.2 & 87.0 & 30.2 & 58.0 & 55.3 & 80.0 & 78.3 & 78.5 & 81.4 & 70.0 & 36.0 & 45.9 & 32.2 & 62.5 & 53.0 & 45.9 & 64.3 & 91.5 & 76.1 \\

$E_{2LF}$, $\mathcal{L}_{D}^{cls-T}$ & 91.0 & 80.3 & 35.8 & 82.9 & 60.9 & 66.4 & 80.9 & 80.1 & 84.3 & 32.8 & 59.4 & 47.7 & 75.9 & 76.0 & 76.4 & 81.6 & 69.5 & 37.7 & 47.2 & 29.9 & 69.8 & 48.0 & 46.5 & 64.0 & 91.0 & 77.1\\

$\mathcal{L}_{D}^{enc}$ & 90.9 & 81.4 & 33.9 & 80.3 & 61.9 & 67.4 & 73.1 & 81.8 & 84.8 & 31.3 & 0.4 & 55.8 & 76.1 & 72.2 & 77.7 & 81.2 & 65.6 & 39.4 & 31.8 & 31.3 & 64.1 & 52.9  & 43.9 & 60.5 & 90.0 & 74.9\\

$\mathcal{L}_{D}^{dec}$ & 91.1 & 85.1 & 31.7 & 80.3 & 62.6 & 72.1 & 82.6 & 79.5 & 84.4 & 31.1 & 34.9 & 56.6 & 77.2 & 75.7 & 77.5 & 81.7 & 69.0 & 40.6 & 43.4 & 30.3 & 70.7 & 52.2 & 47.4 & 63.9 & 91.0 & 77.4 \\
 
 $\mathcal{L}_{D}^{SPKD-avg}$ &  90.6 & 81.7 & 33.0 & 80.4 & 61.9 & 65.4 & 68.4 & 80.9 & 85.6 & 31.1 & 4.8 & 55.1 & 76.3 & 63.8 & 77.3 & 80.0 & 64.8 & 38.5 & 31.4 & 29.9 & 63.2 & 50.8 & 42.7 & 59.5 & 89.7 & 74.0
 \\ \hline

$M_0(0-15)$ & 94.0 & 83.5 & 36.1 & 85.5 & 61.0 & 77.7 & 94.1 & 82.8 & 90.0 & 40.0 & 82.8 & 54.9 & 83.4 & 81.2 & 78.3 & 83.2 & 75.5 & - & - & - & - & - & - & 75.5 & 94.6 & 86.4\\

$M_0(0-20)$ & 93.4 & 85.4 & 36.7 & 85.7 & 63.3 & 78.7 & 92.7 & 82.4 & 89.7 & 35.4 & 80.9 & 52.9 & 82.4 & 82.0 & 76.8 & 83.6 & 75.1 & 52.3 & 82.4 & 51.1 & 86.4 & 70.5 & 68.5 & 73.6 & 93.9 & 84.2 \\
\hline
\end{tabular}
\end{table*}

\begin{table*}[htbp]
\vspace{-0.1cm}
\caption{Per-class IoU of the proposed approaches on VOC2012 when $10$ classes are added at once.}
\vspace{-0.2cm}
\label{tab:pascal_0_10_20}
\setlength{\tabcolsep}{1.6pt}
\centering
\footnotesize
\begin{tabular}{|c|ccccccccccc:c|cccccccccc:c|ccc|}
\hline
 $M_1 (11-20)$ & \scriptsize\rotatebox{90}{backgr.} &  \scriptsize\rotatebox{90}{aero} &  \scriptsize\rotatebox{90}{bike} &  \scriptsize\rotatebox{90}{bird} &\scriptsize\rotatebox{90}{boat} & \scriptsize\rotatebox{90}{bottle} & \scriptsize\rotatebox{90}{bus} 
  &\scriptsize\rotatebox{90}{car} & \scriptsize\rotatebox{90}{cat} & \scriptsize\rotatebox{90}{chair} & \scriptsize\rotatebox{90}{cow} & \scriptsize\rotatebox{90}{\textbf{mIoU old}} & \scriptsize\rotatebox{90}{din. table}& \scriptsize\rotatebox{90}{dog} & \scriptsize\rotatebox{90}{horse} & \scriptsize\rotatebox{90}{mbike} & \scriptsize\rotatebox{90}{person} & \scriptsize\rotatebox{90}{plant} &  \scriptsize\rotatebox{90}{sheep} & \scriptsize\rotatebox{90}{sofa} & \scriptsize\rotatebox{90}{train} & \scriptsize\rotatebox{90}{tv} & \scriptsize\rotatebox{90}{\textbf{mIoU new}} & \scriptsize\rotatebox{90}{\textbf{mIoU}} & \scriptsize\rotatebox{90}{\textbf{mPA}} & \scriptsize\rotatebox{90}{\textbf{mCA}}\\
 \hline

Fine-tuning & 91.9 & 82.4 & 32.0 & 70.8 & 61.7 & 67.7 & 91.1 & 79.8 & 72.7 & 30.5 & 61.6 & 67.5 & 49.1 & 70.6 & 63.4 & 72.9 & 79.4 & 43.5 & 72.0 & 44.8 & 79.1 & 60.7 & \textbf{63.5} & 65.6 & \textbf{91.9} & 78.2 \\

{\revision $E_F$}         & 91.3 & 85.0 & 33.1 & 82.0 & 63.2 & 75.2 & 89.5 & 76.5 & 74.9 & 25.4 & 67.7 & 69.4 & 42.2 & 64.4 & 66.2 & 68.2 & 67.7 & 38.6 & 69.8 & 32.9 & 72.1 & 59.5 & 58.2 & 64.1 & 91.1 & 75.0 \\

{\revision $E_{2LF}$}         & 91.8 & 82.7 & 32.6 & 72.9 & 60.2 & 67.2 & 91.3 & 80.7 & 75.7 & 30.7 & 61.6 & 68.0 & 46.9 & 70.2 & 65.2 & 72.2 & 78.8 & 43.4 & 70.1 & 44.4 & 81.4 & 58.6 & 63.1 & 65.6 & 91.8 & 78.6 \\

$\mathcal{L}_{D}^{cls-T}$ & 91.7 & 83.2 & 33.4 & 80.9 & 62.3 & 72.6 & 89.2 & 76.8 & 77.6 & 28.0 & 64.1 & 69.1 & 48.6 & 73.5 & 65.7 & 72.9 & 76.6 & 41.3 & 74.2 & 39.5 & 79.0 & 62.1 & 63.3 & 66.3 & \textbf{91.9} & 77.3 \\

$E_F$, $\mathcal{L}_{D}^{cls-T}$ & 91.4 & 85.2 & 33.3 & 82.5 & 62.7 & 75.1 & 89.7 & 76.4 & 75.3 & 25.9 & 67.9 & 69.6 & 42.2 & 64.7 & 66.4 & 68.0 & 67.9 & 39.4 & 70.4 & 32.9 & 72.5 & 60.5 & 58.5 & 64.3 & 91.2 & 75.2 \\

$E_{2LF}$, $\mathcal{L}_{D}^{cls-T}$ & 91.8 & 84.0 & 33.6 & 83.2 & 62.7 & 72.4 & 90.9 & 77.0 & 79.9 & 28.2 & 65.4 & \textbf{69.9} & 46.8 & 72.7 & 66.8 & 71.5 & 75.3 & 41.1 & 74.2 & 38.2 & 80.0 & 59.7 & 62.6 & \textbf{66.5} & \textbf{91.9} & 77.7 \\

$\mathcal{L}_{D}^{enc}$ & 92.1 & 83.5 & 34.0 & 79.5 & 61.7 & 69.1 & 90.9 & 78.5 & 72.5 & 29.3 & 61.2 & 68.4 & 46.2 & 66.1 & 65.3 & 74.3 & 79.1 & 43.0 & 70.0 & 47.1 & 78.3 & 63.5 & 63.3 & 66.0 & 91.8 & \textbf{79.4} \\ 

$\mathcal{L}_{D}^{dec}$ & 92.0 & 84.5 & 33.5 & 74.7 & 61.2 & 71.5 & 89.7 & 77.9 & 73.5 & 28.6 & 61.8 & 68.1 & 51.9 & 67.1 & 64.9 & 70.8 & 77.3 & 42.8 & 70.5 & 45.3 & 78.9 & 61.6 & 63.1 & 65.7 & \textbf{91.9} & 77.3 \\
 
 $\mathcal{L}_{D}^{SPKD-avg}$ &  91.9 & 82.1 & 32.3 & 76.4 & 61.4 & 66.1 & 91.3 & 78.5 & 72.9 & 30.0 & 60.7 & 67.6 & 47.1 & 68.6 & 65.2 & 74.8 & 79.7 & 42.1 & 68.9 & 47.4 & 78.4 & 62.2 & 63.4 & 65.6 & 91.8 & 79.0
 \\ \hline
 
$M_0(0-10)$ & 95.3 & 86.4 & 34.4 & 85.6 & 69.7 & 79.3 & 94.6 & 87.6 & 93.1 & 44.2 & 91.9 & 78.4 & - & - & - & - & - & - & - & - & - & - & - & 78.4 & 96.1 & 90.4
 \\
 
$M_0(0-20)$ & 93.4 & 85.4 & 36.7 & 85.7 & 63.3 & 78.7 & 92.7 & 82.4 & 89.7 & 35.4 & 80.9 & 74.9 & 52.9 & 82.4 & 82.0 & 76.8 & 83.6 & 52.3 & 82.4 & 51.1 & 86.4 & 70.5 & 72.1 & 73.6 & 93.9 & 84.2 \\
\hline
\end{tabular}
\end{table*}

The next experiment regards the addition of $10$ classes at once and the results are shown in Table~\ref{tab:pascal_0_10_20}. Here knowledge distillation is less effective. Indeed, even though it enhances the results, the improvement is  smaller with respect to the one in other scenarios. In particular, the idea of freezing only the first two layers of the encoder, introduced in this version of the work, together with knowldege distillation leads to the best results in this setting.
The gap is reduced  because the fine-tuning approach already achieves quite high results preventing other methods to largely overcome it. We argue that the critical aspect is that the cardinality of the set of classes being added is comparable to the one of the set of previously learned classes.

\subsection{Sequential Addition of Multiple Classes on VOC2012}

The last set of experiments on VOC2012 regards the addition of one or more classes more than once (i.e., new classes are progressively added instead of all in one shot).

\begin{table*}[htbp]
\vspace{-0.1cm}
\caption{Per-class IoU of the proposed approaches on VOC2012 when $5$ classes are added two times.}
\vspace{-0.2cm}
\label{tab:pascal_0_10_15_20}
\setlength{\tabcolsep}{1.6pt}
\centering
\footnotesize
\begin{tabular}{|c|ccccccccccc:c|ccccc:ccccc:c|ccc|}
\hline
 $M_2 (11\!-\!15,16\!-\!20)$ & \scriptsize\rotatebox{90}{backgr.} &  \scriptsize\rotatebox{90}{aero} &  \scriptsize\rotatebox{90}{bike} &  \scriptsize\rotatebox{90}{bird} &\scriptsize\rotatebox{90}{boat} & \scriptsize\rotatebox{90}{bottle} & \scriptsize\rotatebox{90}{bus} 
  &\scriptsize\rotatebox{90}{car} & \scriptsize\rotatebox{90}{cat} & \scriptsize\rotatebox{90}{chair} & \scriptsize\rotatebox{90}{cow} & \scriptsize\rotatebox{90}{\textbf{mIoU old}} & \scriptsize\rotatebox{90}{din. table}& \scriptsize\rotatebox{90}{dog} & \scriptsize\rotatebox{90}{horse} 
  & \scriptsize\rotatebox{90}{mbike} & \scriptsize\rotatebox{90}{person} & \scriptsize\rotatebox{90}{plant} &  \scriptsize\rotatebox{90}{sheep} & \scriptsize\rotatebox{90}{sofa} & \scriptsize\rotatebox{90}{train} & \scriptsize\rotatebox{90}{tv} & \scriptsize\rotatebox{90}{\textbf{mIoU new}} & \scriptsize\rotatebox{90}{\textbf{mIoU}} & \scriptsize\rotatebox{90}{\textbf{mPA}} & \scriptsize\rotatebox{90}{\textbf{mCA}}\\
 \hline
Fine-tuning & 89.3 & 56.8 & 32.0 & 60.6 & 56.0 & 55.8 & 36.8 & 75.7 & 76.4 & 28.2 & 4.1 & 52.0 & 47.3 & 67.8 & 50.8 & 69.2 & 78.3 & 34.8 & 27.6 & 25.7 & 44.8 & 37.0 & 48.3 & 50.2 & 86.9 & 66.2 \\

{\revision $E_F$} & 88.5 & 71.5 & 29.7 & 49.9 & 41.2 & 68.4 & 4.4 & 79.0 & 80.3 & 24.8 & 6.2 & 49.5 & 44.7 & 66.2 & 38.5 & 68.1 & 73.7 & 36.4 & 25.3 & 27.1 & 46.3 & 52.7 & 47.9 & 48.7 & 86.5 & 62.7 \\

$\mathcal{L}_{D}^{cls-T}$ & 90.3 & 80.7 & 33.5 & 74.9 & 62.6 & 62.3 & 74.2 & 77.2 & 78.5 & 27.4 & 23.6 & 62.3 & 44.3 & 70.6 & 61.2 & 72.5 & 78.9 & 38.1 & 37.5 & 29.7 & 62.4 & 46.9 & 54.2 & 58.4 & 89.5 & 73.0 \\

$E_F$, $\mathcal{L}_{D}^{cls-T}$ & 90.3 & 85.0 & 32.6 & 73.2 & 61.3 & 72.5 & 79.2 & 79.7 & 81.0 & 27.1 & 31.3 & \textbf{64.8} & 42.5 & 68.5 & 58.5 & 68.5 & 73.1 & 36.5 & 33.9 & 30.5 & 68.0 & 54.8 & 53.5 & \textbf{59.4} & \textbf{89.8} & 71.4 \\



$\mathcal{L}_{D}^{enc}$ & 89.8 & 64.7 & 33.3 & 73.7 & 58.3 & 63.8 & 48.7 & 77.9 & 79.8 & 28.4 & 11.4 & 57.2 & 50.1 & 68.2 & 53.0 & 70.8 & 79.2 & 39.0 & 28.9 & 26.8 & 49.4 & 44.2 & 51.0 & 54.3 & 88.0 & 69.7 \\ 

$\mathcal{L}_{D}^{dec}$ & 90.6 & 81.8 & 32.9 & 77.7 & 62.5 & 66.7 & 78.8 & 78.7 & 79.2 & 27.7 & 25.1 & 63.8 & 49.7 & 69.1 & 56.6 & 72.1 & 79.5 & 40.1 & 34.2 & 28.5 & 65.5 & 50.7 & \textbf{54.6} & \textbf{59.4} & 89.6 & \textbf{73.9} \\
 
 $\mathcal{L}_{D}^{SPKD-avg}$ & 89.8 & 78.1 & 30.3 & 58.6 & 52.6 & 65.6 & 53.8 & 78.0 & 74.3 & 29.9 & 4.7 & 56.0 & 46.9 & 62.6 & 49.3 & 68.6 & 78.4 & 32.7 & 23.0 & 30.0 & 61.7 & 49.0 & 50.2 & 53.2 & 88.0 & 67.7 
 \\ \hline

$M_0(0-10)$ & 95.3 & 86.4 & 34.4 & 85.6 & 69.7 & 79.3 & 94.6 & 87.6 & 93.1 & 44.2 & 91.9 & 78.4 & - & - & - & - & - & - & - & - & - & - & - & 78.4 & 96.1 & 90.4
 \\
 
$M_0(0-20)$ & 93.4 & 85.4 & 36.7 & 85.7 & 63.3 & 78.7 & 92.7 & 82.4 & 89.7 & 35.4 & 80.9 & 74.9 & 52.9 & 82.4 & 82.0 & 76.8 & 83.6 & 52.3 & 82.4 & 51.1 & 86.4 & 70.5 & 72.1 & 73.6 & 93.9 & 84.2 \\
\hline
\end{tabular}
\end{table*}

Let us start from the case in which two sets of $5$ classes are added in two incremental steps after an initial training stage with $10$ classes, leading to the final model  $M_2 (11\!-\!15,16\!-\!20)$. 
The mIoU results are reported in Table~\ref{tab:pascal_0_10_15_20} where we can appreciate a more severe drop in performance if compared with the introduction of all the $10$ classes in a single shot.
In particular, the standard approach without distillation leads to a very poor mIoU of $50.2\%$.
Catastrophic forgetting is largely mitigated by knowledge distillation, that in this case proved to be very 
effective.
In the best settings, that in this case are the distillation $\mathcal{L}_D^{cls-T}$ with $E_F$ and the newly introduced distillation applied to the dilation layers ($\mathcal{L}_D^{dec}$), the mIoU improves of  $9.2\%$ with respect to the standard approach. 
The method using $\mathcal{L}_D^{dec}$ is also the one obtaining the best mIoU on the new classes.

\begin{table*}[htbp]
\vspace{-0.1cm}
\caption{Per-class IoU of the proposed approaches on VOC2012 when $5$ classes are added two times with classes ordered based on the occurrence in the dataset.}
\vspace{-0.2cm}
\label{tab:pascal_0_10_15_20_occurrences}
\setlength{\tabcolsep}{1.6pt}
\centering
\footnotesize
\begin{tabular}{|c|ccccccccccc:c|ccccc:ccccc:c|ccc|}
\hline
 $M_2 (11\!-\!15,16\!-\!20)$ & \scriptsize\rotatebox{90}{backgr.} &  \scriptsize\rotatebox{90}{person} &  \scriptsize\rotatebox{90}{cat} &  \scriptsize\rotatebox{90}{dog} &\scriptsize\rotatebox{90}{car} & \scriptsize\rotatebox{90}{train} & \scriptsize\rotatebox{90}{chair} 
  &\scriptsize\rotatebox{90}{bus} & \scriptsize\rotatebox{90}{sofa} & \scriptsize\rotatebox{90}{mbike} & \scriptsize\rotatebox{90}{din. table} & \scriptsize\rotatebox{90}{\textbf{mIoU old}} & \scriptsize\rotatebox{90}{aero}& \scriptsize\rotatebox{90}{horse} & \scriptsize\rotatebox{90}{bird} 
  & \scriptsize\rotatebox{90}{bike} & \scriptsize\rotatebox{90}{tv} & \scriptsize\rotatebox{90}{boat} &  \scriptsize\rotatebox{90}{plant} & \scriptsize\rotatebox{90}{sheep} & \scriptsize\rotatebox{90}{cow} & \scriptsize\rotatebox{90}{bottle} & \scriptsize\rotatebox{90}{\textbf{mIoU new}} & \scriptsize\rotatebox{90}{\textbf{mIoU}} & \scriptsize\rotatebox{90}{\textbf{mPA}} & \scriptsize\rotatebox{90}{\textbf{mCA}}\\
 \hline

Fine-tuning & 91.3 & 80.5 & 75.7 & 67.8 & 80.2 & 73.4 & 29.7 & 84.4 & 42.1 & 70.4 & 55.6 & 68.3 & 19.7 & 41.1 & 5.7 & 29.8 & 63.9 & 41.2 & 38.4 & 45.7 & 55.1 & 63.3 & 40.4 & 55.0 & 90.3 & 69.1 \\

{\revision $E_F$} & 92.4 & 82 & 81.5 & 75.5 & 82.8 & 82.9 & 29.1 & 92.9 & 46 & 71.4 & 54.9 & 71.9 & 70.2 & 24.5 & 58 & 31.7 & 59.9 & 46.5 & 39 & 49.3 & 53.1 & 60.5 & 49.3 & 61.1 & 91.7 & 72.3 \\

$\mathcal{L}_{D}^{cls-T}$ & 92.5 & 81.0 & 78.6 & 69.9 & 80.5 & 80.3 & 31.0 & 88.8 & 45.1 & 73.3 & 50.2 & 70.1 & 79.7 & 53.4 & 71.5 & 32.7 & 61.6 & 53.1 & 40.1 & 58.8 & 59.9 & 71.0 & \textbf{58.2} & \textbf{64.4} & \textbf{92.2} & \textbf{76.0} \\

$E_F$, $\mathcal{L}_{D}^{cls-T}$ & 92.5 & 81.7 & 82.1 & 76.1 & 83.5 & 83.1 & 29.0 & 92.7 & 46.7 & 71.2 & 55.4 & \textbf{72.2} & 70.9 & 29.5 & 59.2 & 32.0 & 59.6 & 46.3 & 38.5 & 49.0 & 52.4 & 61.6 & 49.9 & 61.6 & 91.9 & 72.6 \\


$\mathcal{L}_{D}^{enc}$ & 92.5 & 82.2 & 82.7 & 74.2 & 81.8 & 78.7 & 31.8 & 88.0 & 46.2 & 73.8 & 58.3 & 71.8 & 66.0 & 39.8 & 56.9 & 31.0 & 63.5 & 42.6 & 45.3 & 54.5 & 60.2 & 69.3 & 52.9 & 62.8 & 92.0 & 74.7 \\

$\mathcal{L}_{D}^{dec}$ & 92.0 & 81.2 & 82.6 & 68.2 & 78.3 & 81.4 & 29.1 & 91.3 & 45.1 & 71.6 & 56.9 & 70.7 & 0.3 & 23.4 & 0.1 & 23.6 & 61.6 & 46.8 & 44.1 & 49.6 & 59.4 & 70.0 & 37.9 & 55.1 & 91.0 & 66.8 \\
 
 $\mathcal{L}_{D}^{SPKD-avg}$ & 92.1 & 81.2 & 82.9 & 74.5 & 82.0 & 79.5 & 30.3 & 88.5 & 42.3 & 74.7 & 55.5 & 71.2 & 62.7 & 25.1 & 42.9 & 31.2 & 61.3 & 45.4 & 43.3 & 47.0 & 55.9 & 70.9 & 48.6 & 60.4 & 91.4 & 73.5 \\ \hline

$M_0(0-10)$ & 92.5 & 80.6 & 89.2 & 85.5 & 86.3 & 86.8 & 30.7 & 93.3 & 46.2 & 80.7 & 59.6 & 75.6 & - & - & - & - & - & - & - & - & - & - & - & 75.6 & 93.5 & 82.8
 \\
 
 $M_0(0-20)$ & 93.4 & 83.6 & 89.7 & 82.4 & 82.4 & 86.4 & 35.4 & 92.7 & 51.1 & 76.8 & 52.9 & 75.2 & 85.4 & 82.0 & 85.7 & 36.7 & 70.5 & 63.3 & 52.3 & 82.4 & 80.9 & 78.7 & 71.8 & 73.6 & 93.9 & 84.2 \\
 
\hline
\end{tabular}
\end{table*}

In Table~\ref{tab:pascal_0_10_15_20_occurrences} the same scenario is evaluated when classes are sorted on the basis of their occurrence inside the dataset. 
Also in this case a large {\revision gain} can be obtained with knowledge distillation which leads to $9.4\%$ of improvement in the best case with respect to fine tuning. We can notice that the old classes are better preserved in this case being also the most frequent inside the dataset. Additionally, some methods struggle in learning new classes needing more samples to detect them.

\begin{table*}[htbp]
\vspace{-0.1cm}
\caption{Per-class IoU of the proposed approaches on VOC2012 when $5$ classes are added sequentially.}
\vspace{-0.2cm}
\label{tab:pascal_0_15_16_17_18_19_20}
\setlength{\tabcolsep}{1.6pt}
\centering
\footnotesize
\begin{tabular}{|c|cccccccccccccccc:c|c:c:c:c:c:c|ccc|}
\hline
 $M_5 (16\rightarrow20)$ & \scriptsize\rotatebox{90}{backgr.} &  \scriptsize\rotatebox{90}{aero} &  \scriptsize\rotatebox{90}{bike} &  \scriptsize\rotatebox{90}{bird} &\scriptsize\rotatebox{90}{boat} & \scriptsize\rotatebox{90}{bottle} & \scriptsize\rotatebox{90}{bus} 
  &\scriptsize\rotatebox{90}{car} & \scriptsize\rotatebox{90}{cat} & \scriptsize\rotatebox{90}{chair} & \scriptsize\rotatebox{90}{cow} & \scriptsize\rotatebox{90}{din. table}& \scriptsize\rotatebox{90}{dog} & \scriptsize\rotatebox{90}{horse} 
  & \scriptsize\rotatebox{90}{mbike} & \scriptsize\rotatebox{90}{person} & \scriptsize\rotatebox{90}{\textbf{mIoU old}} & \scriptsize\rotatebox{90}{plant} &  \scriptsize\rotatebox{90}{sheep} & \scriptsize\rotatebox{90}{sofa} & \scriptsize\rotatebox{90}{train} & \scriptsize\rotatebox{90}{tv} & \scriptsize\rotatebox{90}{\textbf{mIoU new}} & \scriptsize\rotatebox{90}{\textbf{mIoU}} & \scriptsize\rotatebox{90}{\textbf{mPA}} & \scriptsize\rotatebox{90}{\textbf{mCA}}\\
 \hline

Fine-tuning & 87.9 & 25.6 & 29.0 & 51.2 & 1.7 & 57.8 & 10.5 & 64.8 & 80.5 & 30.8 & 22.9 & 52.7 & 66.8 & 52.1 & 51.9 & 78.1 & 47.8 & \textbf{36.5} & 44.7 & 31.8 & 35.1 & 17.1 & 33.0 & 44.2 & 86.1 & 55.7 \\

{\revision $E_F$} & 90.4 & 62.6 & 30.9 & 81.7 & 53.9 & 70.8 & 57.8 & 80.7 & 86.2 & 27.0 & 71.5 & 56.8 & 75.5 & 77.2 & 73.2 & 78.2 & 67.1 & 36.3 & 65.1 & 30.1 & 52.9 & 34.6 & 43.8 & 61.6 & 90.4 & 72.2 \\

$\mathcal{L}_{D}^{cls-T}$ & 89.8 & 51.2 & 29.9 & 77.2 & 15.6 & 62.0 & 29.2 & 78.5 & 75.7 & 24.4 & 55.6 & 44.8 & 76.2 & 62.5 & 65.6 & 80.1 & 57.4 & 27.0 & 35.2 & 30.6 & 42.3 & 39.7 & 35.0 & 52.3 & 88.6 & 63.2 \\

$E_F$, $\mathcal{L}_{D}^{cls-T}$ & 91.1 & 73.9 & 31.9 & 81.4 & 59.5 & 71.9 & 73.1 & 82.1 & 87.1 & 27.2 & 77.4 & 56.4 & 79.1 & 79.9 & 76.1 & 80.7 & \textbf{70.5} & 31.8 & 55.8 & 30.1 & \textbf{62.3} & \textbf{41.4} & 44.3 & \textbf{64.6} & \textbf{91.3} & \textbf{75.2} \\


$\mathcal{L}_{D}^{enc}$ & 90.3 & 54.2 & 28.2 & 78.4 & 52.5 & 69.8 & 59.5 & 78.5 & 86.3 & 28.8 & 72.3 & 57.4 & 76.3 & 77.1 & 65.8 & 79.3 & 65.9 & 36.3 & \textbf{65.5} & 31.6 & 54.7 & 38.9 & \textbf{45.4} & 61.0 & 90.4 & 71.0 \\

$\mathcal{L}_{D}^{dec}$ & 90.2 & 69.1 & 31.0 & 78.4 & 32.1 & 61.8 & 41.9 & 73.7 & 83.7 & 30.0 & 54.8 & 52.5 & 69.5 & 62.8 & 61.2 & 81.0 & 60.8 & 30.0 & 46.5 & \textbf{32.5} & 43.5 & 30.0 & 36.5 & 55.1 & 89.2 & 66.5 \\
 
 $\mathcal{L}_{D}^{SPKD-avg}$ & 89.9 & 70.9 & 31.3 & 73.5 & 43.1 & 68.3 & 67.9 & 77.0 & 82.2 & 31.3 & 22.7 & 55.2 & 74.9 & 57.4 & 62.3 & 79.2 & 61.7 & 34.2 & 36.5 & 31.5 & 62.0 & 33.0 & 39.5 & 56.4 & 89.3 & 69.3
 \\ \hline

$M_0(0-15)$ & 94.0 & 83.5 & 36.1 & 85.5 & 61.0 & 77.7 & 94.1 & 82.8 & 90.0 & 40.0 & 82.8 & 54.9 & 83.4 & 81.2 & 78.3 & 83.2 & 75.5 & - & - & - & - & - & - & 75.5 & 94.6 & 86.4 \\

$M_0(0-20)$ & 93.4 & 85.4 & 36.7 & 85.7 & 63.3 & 78.7 & 92.7 & 82.4 & 89.7 & 35.4 & 80.9 & 52.9 & 82.4 & 82.0 & 76.8 & 83.6 & 75.1 & 52.3 & 82.4 & 51.1 & 86.4 & 70.5 & 68.5 & 73.6 & 93.9 & 84.2 \\
\hline
\end{tabular}
\end{table*}

Then we move to consider the sequential addition of the last $5$ classes one by one, i.e., model $M_5 (16 \rightarrow 20)$.
 The results are reported in Table~\ref{tab:pascal_0_15_16_17_18_19_20} where we can appreciate a huge gain of $20.4\%$ of mIoU between the best proposed method (i.e. $M_5 (16 \rightarrow 20) [ E_F, \mathcal{L}_{D}^{cls-T}]$) and the baseline approach. In this case freezing the encoder and distilling the knowledge is found to be very reliable because the addition of one single class should not alter too much the responses of the whole network. Distilling the knowledge from the previous model when the encoder is fixed guides the decoder to modify only the responses for the new class: in this way the best result is obtained.
\begin{table*}[]
\vspace{-0.1cm}
\caption{mIoU, mPA and mCA of the proposed approaches on VOC2012 when $5$ classes are added sequentially.}
\vspace{-0.2cm}
\label{tab:pascal_0_15_16_17_18_19_20_methods_vs_steps}
\setlength{\tabcolsep}{1.6pt}
\centering
\footnotesize
\begin{tabular}{|l|c|c|c||c|c|c||c|c|c||c|c|c||c|c|c||c|c|c||c|c|c|}
\hline
\multirow{2}{*}{} & \multicolumn{3}{c||}{Fine-tuning} & \multicolumn{3}{c||}{\revision $E_F$} & \multicolumn{3}{c||}{$\mathcal{L}_D^{cls-T}$} & \multicolumn{3}{c||}{$E_F$, $\mathcal{L}_D^{cls-T}$} &
\multicolumn{3}{c||}{$\mathcal{L}_D^{dec}$} & \multicolumn{3}{c||}{$\mathcal{L}_D^{enc}$} & \multicolumn{3}{c|}{$\mathcal{L}_D^{SPKD-avg}$} \\ \cline{2-22}
  & mIoU    & mPA   & mCA & mIoU    & mPA   & mCA   & mIoU    & mPA   & mCA   & mIoU   & mPA   & mCA  & 
  mIoU   & mPA   & mCA & mIoU   & mPA   & mCA & mIoU   & mPA   & mCA  \\
  \hline
$M_1(16)$ & 71.2 & 93.7 & 82.5 & 71.8 & 93.7 & 84.4 & 72.4 & \textbf{94.2} & 83.0 & 72.5 & 94.1 & 83.5 &
\textbf{72.9} & \textbf{94.2} & \textbf{84.5} & 72.2 & 93.9 & 84.3 & 71.3 & 93.7 & 82.6   \\
$M_2(17)$  &  53.8 & 90.0 & 61.8  & 59.1 & 91.2 & 69.0 & 68.1 & \textbf{93.4} & 78.5 & \textbf{68.4} & 93.3 & \textbf{79.5} & 
 68.0 & \textbf{93.4} & 78.6 & 60.0 & 91.6 & 69.4 & 56.5 & 90.7 & 65.6 \\
$M_3(18)$ & 57.7 & 87.7 & 68.7  & 65.7 & 90.4 & 78.0 & 63.3 & 90.8 & 74.5 & \textbf{66.5} & \textbf{91.5} & \textbf{79.4} & 
 64.6 & 90.2 & 76.9 & 65.5 & 90.7 & 76.8 & 58.3 & 88.3 & 70.1 \\
$M_4(19)$ &  39.3 & 85.9 & 47.4 & 52.6 & 89.0 & 61.4 & 54.1 & 89.2 & 64.3 & \textbf{61.3} & \textbf{90.6} & \textbf{72.5} & 
57.9 & 89.7 & 69.0 & 52.1 & 89.0 & 60.6 & 54.3 & 89.5 & 64.9\\
$M_5(20)$&  44.2 & 86.1 & 55.7 & 61.6 & 90.4 & 72.2 & 52.3 & 88.6 & 63.2 & \textbf{64.6} & \textbf{91.3} & \textbf{75.2} & 
55.1 & 89.2 & 66.5 & 61.0 & 90.4 & 71.0 & 56.4 & 89.3 & 69.3 \\
\hline
\end{tabular}
\end{table*}
The evolution of the models' mean performance during the various steps is reported in Table~\ref{tab:pascal_0_15_16_17_18_19_20_methods_vs_steps} where we can appreciate how the drop of performance is distributed during the different steps. In particular, we can notice how the accuracy drop is affected by the specific class being added. As expected, the larger drop is experienced when the classes $\mathit{sheep}$ or $\mathit{train}$ are added because such classes typically appear alone or with the \textit{person} class, i.e., they are only sparsely correlated with a few other classes. The opposite is true when the classes being added show high assortativity coefficient with other classes, for example the presence of the classes $\mathit{potted \ plant}$ and $\mathit{tv/monitor}$ is highly correlated with the presence of classes like $\mathit{dining \ table}$, $\mathit{person}$ or $\mathit{chair}$.
\begin{figure*}[h!]{}
\setlength\tabcolsep{1.5pt} 
\centering
\begin{tabular}{cccccccc} 
  {$\scriptstyle  RGB$} &
   {$\scriptstyle GT$} &
  {$\scriptstyle Fine-tuning$} &
 {$ \scriptstyle \! \! M_1\!(16\!-\!20)\! [E_F\!,\mathcal{L}_{D}^{cls-T}]\!$}&
   {$\scriptstyle M_1(16-20) [\mathcal{L}_{D}^{enc}]$}&
   {$\scriptstyle M_1(16-20) [\mathcal{L}_{D}^{dec}]$} &
   {$\scriptstyle M_1(16-20) [\mathcal{L}_{D}^{SPKD-avg}]$} &
  {$\scriptstyle M_0(0-20)$} \\
  
    \includegraphics[width=\sizefiggg\linewidth]{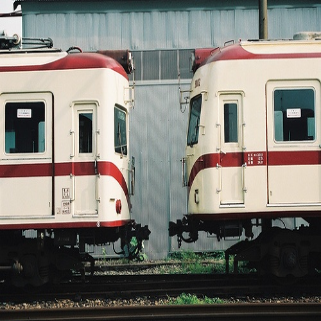} &
  \includegraphics[width=\sizefiggg\linewidth]{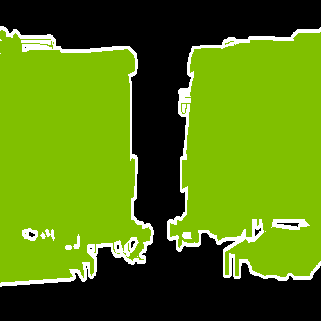} &
  \includegraphics[width=\sizefiggg\linewidth]{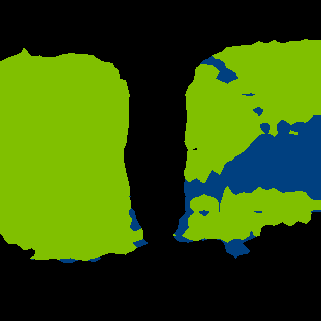} &
    \includegraphics[width=\sizefiggg\linewidth]{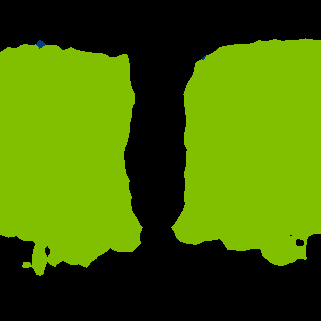} &
  \includegraphics[width=\sizefiggg\linewidth]{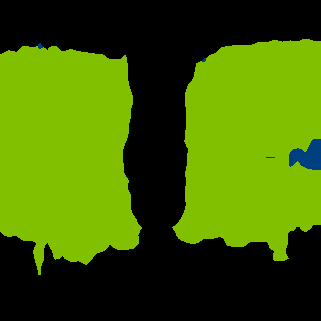} &
  \includegraphics[width=\sizefiggg\linewidth]{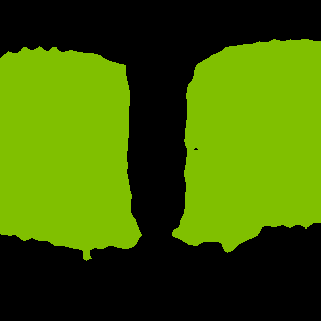} &
  \includegraphics[width=\sizefiggg\linewidth]{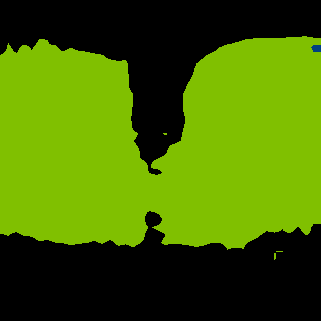} &
    \includegraphics[width=\sizefiggg\linewidth]{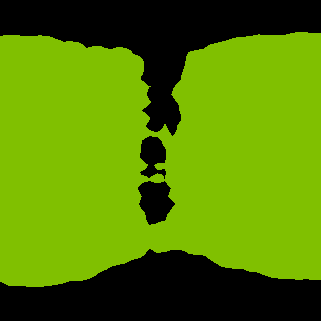} \\

        \includegraphics[width=\sizefiggg\linewidth]{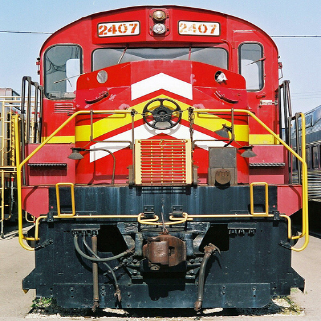}&
    \includegraphics[width=\sizefiggg\linewidth]{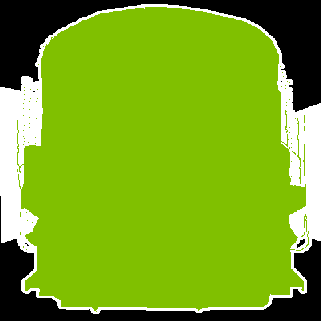} &
    \includegraphics[width=\sizefiggg\linewidth]{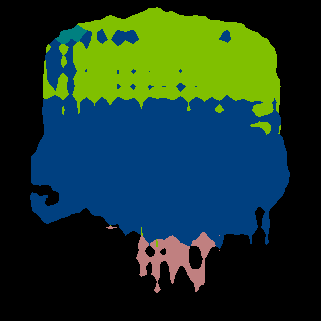}  &
      \includegraphics[width=\sizefiggg\linewidth]{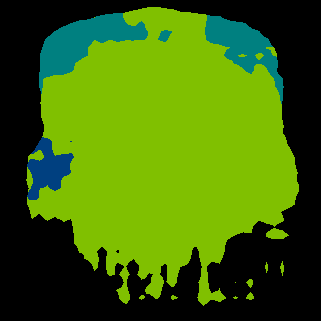} &
  \includegraphics[width=\sizefiggg\linewidth]{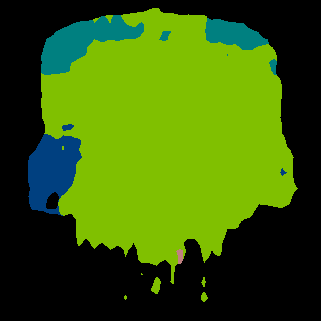} &
  \includegraphics[width=\sizefiggg\linewidth]{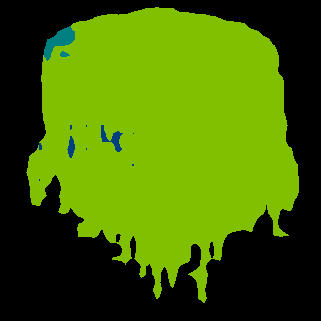} &
  \includegraphics[width=\sizefiggg\linewidth]{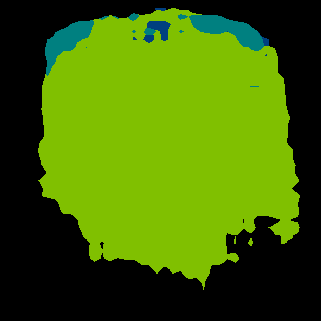} &
  \includegraphics[width=\sizefiggg\linewidth]{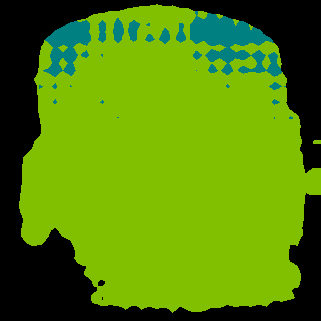} \\

  \includegraphics[width=\sizefiggg\linewidth]{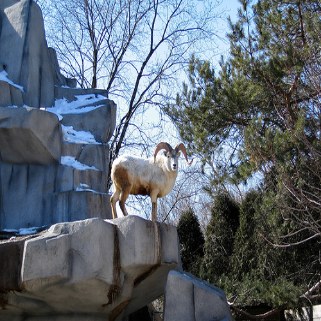} &
    \includegraphics[width=\sizefiggg\linewidth]{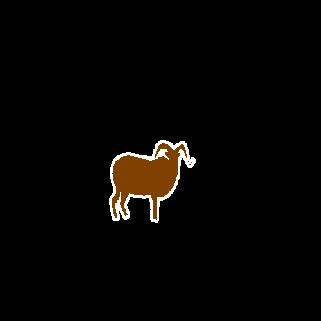}  &
  \includegraphics[width=\sizefiggg\linewidth]{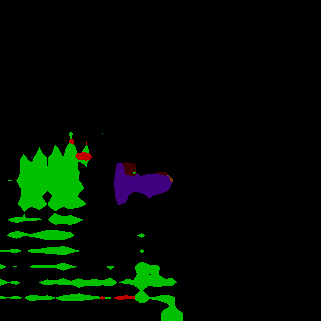} &
  \includegraphics[width=\sizefiggg\linewidth]{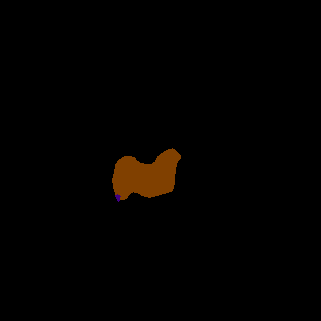}  &
  \includegraphics[width=\sizefiggg\linewidth]{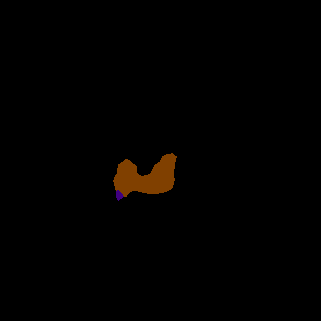}  &
  \includegraphics[width=\sizefiggg\linewidth]{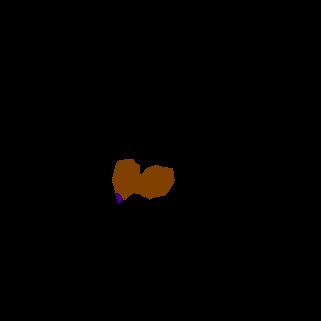}  &
  \includegraphics[width=\sizefiggg\linewidth]{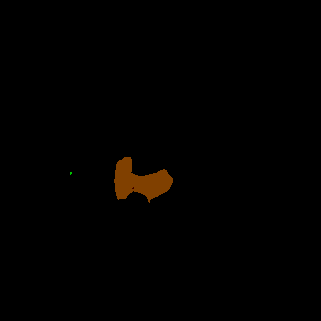}  &
  \includegraphics[width=\sizefiggg\linewidth]{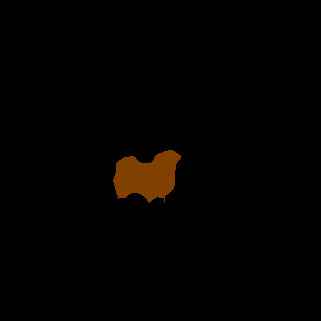}\\

\end{tabular}
\centering
\subfloat{
\hspace{0.22cm}
\resizebox{8cm}{!}{%
\begin{tabular}{ccccccccccc}
\vspace{-0.71cm}
& & & & & & & & & &\\ \cline{11-11}
\cellcolor[HTML]{000000}{\color[HTML]{FFFFFF} \textbf{\enspace background \enspace }} & 
\cellcolor[HTML]{008080}{\color[HTML]{FFFFFF} \textbf{\enspace bus \enspace }} & 
\cellcolor[HTML]{400000}{\color[HTML]{FFFFFF} \textbf{cat}} & 
\cellcolor[HTML]{C00000}{\color[HTML]{FFFFFF} \textbf{chair}} &
\cellcolor[HTML]{400080}{\color[HTML]{FFFFFF} \textbf{dog}} & 
\cellcolor[HTML]{C08080}\textbf{\enspace person\enspace } &
\cellcolor[HTML]{7E3E10}{\color[HTML]{FFFFFF} \textbf{\enspace sheep \enspace }} &  
\cellcolor[HTML]{00C000}\textbf{sofa} &  
\cellcolor[HTML]{80C000}\textbf{\enspace train \enspace } & 
\multicolumn{1}{c|}{\cellcolor[HTML]{004080}{\color[HTML]{FFFFFF} \textbf{\quad tv\quad }}} &
\multicolumn{1}{c|}{\textbf{\enspace unlabeled \enspace }}\\ \cline{11-11} 
\end{tabular}%
}
}%
\vspace{-0.2cm}
\caption{Sample qualitative results  for the addition of 5 classes sequentially (\textit{best viewed in colors}). The  added classes are \textit{boat}, \textit{plant}, \textit{sheep}, \textit{cow} and \textit{bottle}.}
\label{fig:addition_of_5_sequentially}
\end{figure*}
Some visual results for this scenario are reported in Fig.~\ref{fig:addition_of_5_sequentially} where a large gap in performance between the na\"ive approach and some of the best performing proposals can be appreciated. In particular, the standard approach without knowledge distillation tends to overestimate the presence of the last seen class, i.e., $\mathit{tv/monitor}$, in spite of the $\mathit{background}$ or of other previously learned classes.

\begin{figure}[h!]{}
\setlength\tabcolsep{1.5pt} 
\subfloat{
\begin{tabular}{ccccc} 
  \scriptsize \scriptsize\rotatebox{90}{\quad \quad \quad  RGB} &
  \includegraphics[width=0.23\linewidth]{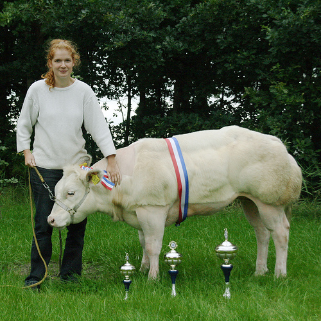} &
    \includegraphics[width=0.23\linewidth]{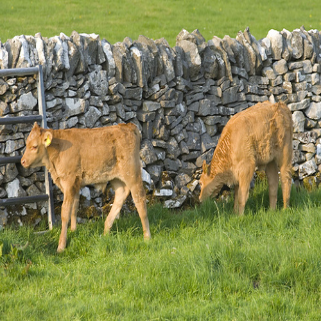}&
    \includegraphics[width=0.23\linewidth]{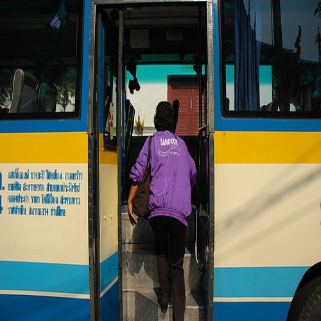} &
    \includegraphics[width=0.23\linewidth]{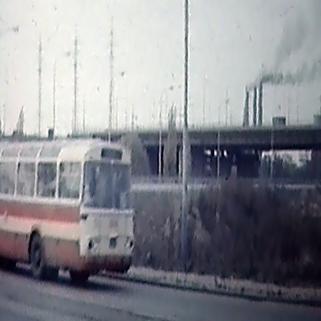} \\
    
      \scriptsize \scriptsize\rotatebox{90}{\quad \quad \ \ \,  GT} &
  \includegraphics[width=0.23\linewidth]{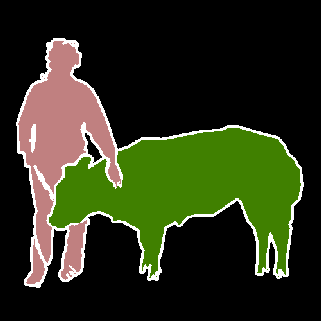} &
    \includegraphics[width=0.23\linewidth]{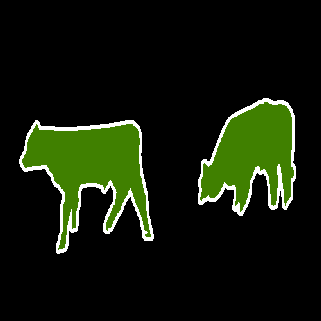}&
    \includegraphics[width=0.23\linewidth]{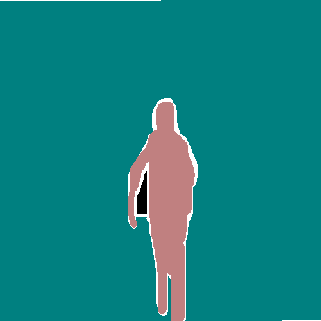} &
    \includegraphics[width=0.23\linewidth]{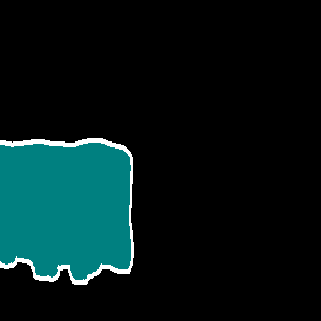} \\
    
    \scriptsize \scriptsize\rotatebox{90}{\quad \quad \ \ before} &
  \includegraphics[width=0.23\linewidth]{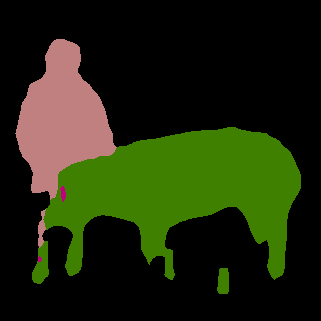} &
    \includegraphics[width=0.23\linewidth]{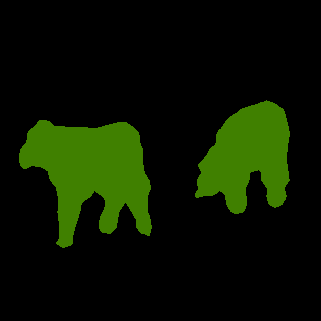}&
    \includegraphics[width=0.23\linewidth]{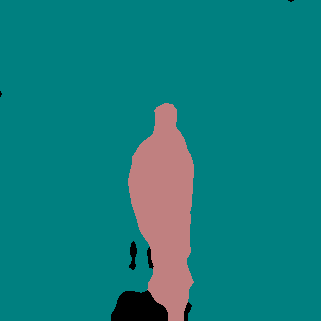} &
    \includegraphics[width=0.23\linewidth]{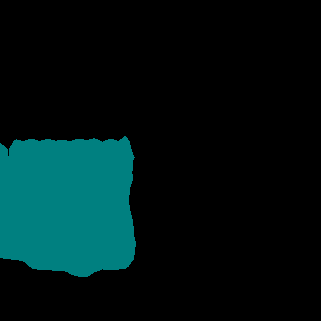} \\
    
    \scriptsize \scriptsize\rotatebox{90}{\quad \quad \quad after} &
  \includegraphics[width=0.23\linewidth]{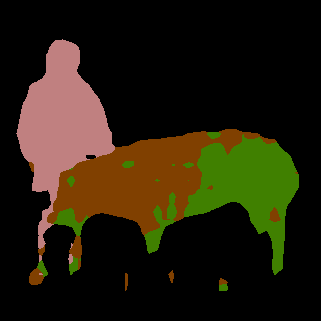} &
    \includegraphics[width=0.23\linewidth]{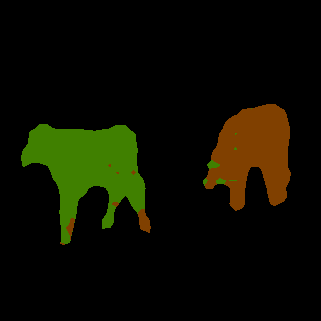}&
    \includegraphics[width=0.23\linewidth]{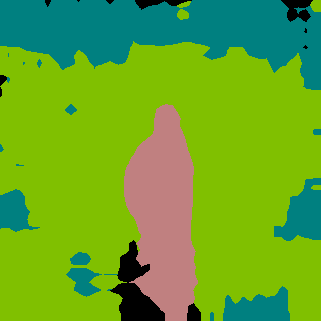} &
    \includegraphics[width=0.23\linewidth]{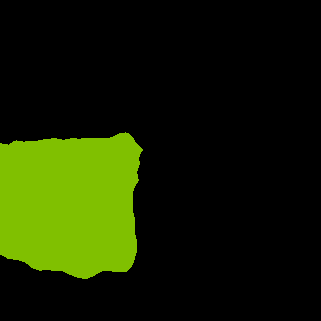} \\
  
\end{tabular}
}

\centering
\vspace{-0.74cm}
\subfloat{
\hspace{0.2cm}
\resizebox{8.45cm}{!}{%
\begin{tabular}{cccccccc}
& & & & & & & \\ \cline{8-8}
\cellcolor[HTML]{000000}{\color[HTML]{FFFFFF} \textbf{\enspace background \enspace }} & 
\cellcolor[HTML]{008080}{\color[HTML]{FFFFFF} \textbf{\enspace bus \enspace }} & 
\cellcolor[HTML]{408000}{\color[HTML]{FFFFFF} \textbf{\enspace cow \enspace }} & 
\cellcolor[HTML]{C00080}{\color[HTML]{FFFFFF} \textbf{\enspace horse \enspace }} & 
\cellcolor[HTML]{C08080}\textbf{\enspace person \enspace } & 
\cellcolor[HTML]{7E3E10}{\color[HTML]{FFFFFF} \textbf{\enspace sheep \enspace }} & 
\multicolumn{1}{c|}{\cellcolor[HTML]{80C000}{\color[HTML]{FFFFFF} \textbf{\enspace train \enspace }}} &
\multicolumn{1}{c|}{\textbf{\enspace unlabeled \enspace }}\\ \cline{8-8} 
\end{tabular}%
}
}%
\vspace{-0.1cm}
\caption{Qualitative comparison on sample scenes of the best model of Table~\ref{tab:pascal_0_15_16_17_18_19_20_best} before and after the addition of a highly correlated class. The first two columns show the performance results after the addition of the $\mathit{sheep}$ class while the last two deals with the addition of the $\mathit{train}$ class (\textit{best viewed in colors}).}
\label{fig:addition_of_5_sequentially_before_and_after}
\end{figure}

\begin{table*}[htbp]
\vspace{-0.1cm}
\caption{Per-class IoU on VOC2012 when $5$ classes are added sequentially. Only the best method of Table~\ref{tab:pascal_0_15_16_17_18_19_20} (``$E_F$ and $\mathcal{L}_D^{cls-T}$") is reported.}
\vspace{-0.2cm}
\label{tab:pascal_0_15_16_17_18_19_20_best}
\setlength{\tabcolsep}{1.6pt}
\centering
\footnotesize
\begin{tabular}{|c|cccccccccccccccc|ccccc|ccc|}
\hline
 & \scriptsize\rotatebox{90}{backgr.} &  \scriptsize\rotatebox{90}{aero} &  \scriptsize\rotatebox{90}{bike} &  \scriptsize\rotatebox{90}{bird} &\scriptsize\rotatebox{90}{boat} & \scriptsize\rotatebox{90}{bottle} & \scriptsize\rotatebox{90}{bus} 
  &\scriptsize\rotatebox{90}{car} & \scriptsize\rotatebox{90}{cat} & \scriptsize\rotatebox{90}{chair} & \scriptsize\rotatebox{90}{cow} & \scriptsize\rotatebox{90}{din. table}& \scriptsize\rotatebox{90}{dog} & \scriptsize\rotatebox{90}{horse} 
  & \scriptsize\rotatebox{90}{mbike} & \scriptsize\rotatebox{90}{person} & \scriptsize\rotatebox{90}{plant} &  \scriptsize\rotatebox{90}{sheep} & \scriptsize\rotatebox{90}{sofa} & \scriptsize\rotatebox{90}{train} & \scriptsize\rotatebox{90}{tv} & \scriptsize\rotatebox{90}{\textbf{mIoU}} & \scriptsize\rotatebox{90}{\textbf{mPA}} & \scriptsize\rotatebox{90}{\textbf{mCA}}\\
 \hline

$M_0(0-15)$ & 94.0 & 83.5 & 36.1 & 85.5 & 61.0 & 77.7 & 94.1 & 82.8 & 90.0 & 40.0 & 82.8 & 54.9 & 83.4 & 81.2 & 78.3 & 83.2 & - & - & - & - & - & 75.5 & 94.6 & 86.4 \\

$M_1(16)$ & 93.5 & 84.0 & 36.1 & 84.8 & 60.5 & 72.5 & 93.4 & 84.2 & 89.7 & 40.0 & 83.0 & 55.7 & 81.9 & 81.6 & 79.4 & 83.2 & 29.0 & - & - & - & - & 72.5 & 94.1 & 83.5  \\

$M_2(17)$ & 93.5 & 84.9 & 35.6 & 72.5 & 61.2 & 73.7 & 93.7 & 83.7 & 79.6 & 39.9 & 73.2 & 57.1 & 78.4 & 74.7 & 79.1 & 83.2 & 29.4 & 37.3 & - & - & - & 68.4 & 93.3 & 79.5 \\

$M_3(18)$ & 91.3 & 83.5 & 34.4 & 76.2 & 61.7 & 72.6 & 93.8 & 83.9 & 85.6 & 26.2 & 77.3 & 57.4 & 78.0 & 77.8 & 78.8 & 81.8 & 30.0 & 46.7 & 26.7 & - & - & 66.5 & 91.5 & 79.4 \\

$M_4(19)$ & 91.2 & 67.8 & 31.7 & 63.9 & 60.5 & 73.1 & 43.2 & 83.5 & 86.4 & 25.1 & 77.7 & 56.7 & 79.1 & 77.9 & 74.3 & 81.7 & 27.0 & 49.2 & 28.0 & 48.7 & - & 61.3 & 90.6 & 72.5 \\

$M_5(20)$ & 91.1 & 73.9 & 31.9 & 81.4 & 59.5 & 71.9 & 73.1 & 82.1 & 87.1 & 27.2 & 77.4 & 56.4 & 79.1 & 79.9 & 76.1 & 80.7 & 31.8 & 55.8 & 30.1 & 62.3 & 41.4 & 64.6 & 91.3 & 75.2 \\ 
\hline

$M_0(0-20)$ & 93.4 & 85.4 & 36.7 & 85.7 & 63.3 & 78.7 & 92.7 & 82.4 & 89.7 & 35.4 & 80.9 & 52.9 & 82.4 & 82.0 & 76.8 & 83.6 & 52.3 & 82.4 & 51.1 & 86.4 & 70.5 & 73.6 & 93.9 & 84.2 \\
\hline
\end{tabular}
\end{table*}

Finally, in Table~\ref{tab:pascal_0_15_16_17_18_19_20_best} we can appreciate the per-class IoU of the best method of Table~\ref{tab:pascal_0_15_16_17_18_19_20}  (i.e., the method combining $\mathcal{L}_D^{cls-T}$ and $E_F$) at each step. An interesting aspect regards the addition of the $\mathit{sofa}$ class which causes a tremendous drop of about one third in terms of IoU for the $\mathit{chair}$ class given their large visual similarity (from $39.9\%$ to $26.2\%$). {\revision An analogue drop for the same reason is experienced also by the $\mathit{bus}$ and $\mathit{aeroplane}$ classes when the $\mathit{train}$ class is added. 
Again, when the $\mathit{sheep}$ class is added visually similar classes of other animals lose accuracy. } 
 These scenarios are depicted in Fig.~\ref{fig:addition_of_5_sequentially_before_and_after} where the first two columns show the results of the best model reported in Table~\ref{tab:pascal_0_15_16_17_18_19_20_best} before and after the addition {\revision of the class $\mathit{sheep}$ 
 and the last two columns show the results before and after the addition of the $\mathit{train}$ class. 
 We can appreciate that regions of the $\mathit{cow}$ class which were correctly identified before the addition of $\mathit{sheep}$ are then confused with the new class and the same happens with the $\mathit{bus}$ and $\mathit{train}$ classes.} 

\begin{table*}[htbp]
\vspace{-0.1cm}
\caption{Per-class IoU of the proposed approaches on VOC2012 when $5$ classes are added sequentially with classes ordered based on their occurrence.}
\vspace{-0.2cm}
\label{tab:pascal_0_15_16_17_18_19_20_occurrences}
\setlength{\tabcolsep}{1.6pt}
\centering
\footnotesize
\begin{tabular}{|c|cccccccccccccccc:c|c:c:c:c:c:c|ccc|}
\hline
$M_5 (16\rightarrow20)$ & \scriptsize\rotatebox{90}{backgr.} &  \scriptsize\rotatebox{90}{person} &  \scriptsize\rotatebox{90}{cat} &  \scriptsize\rotatebox{90}{dog} &\scriptsize\rotatebox{90}{car} & \scriptsize\rotatebox{90}{train} & \scriptsize\rotatebox{90}{chair} 
  &\scriptsize\rotatebox{90}{bus} & \scriptsize\rotatebox{90}{sofa} & \scriptsize\rotatebox{90}{mbike} & \scriptsize\rotatebox{90}{din. table} & \scriptsize\rotatebox{90}{aero}& \scriptsize\rotatebox{90}{horse} & \scriptsize\rotatebox{90}{bird} 
  & \scriptsize\rotatebox{90}{bike} & \scriptsize\rotatebox{90}{tv} & \scriptsize\rotatebox{90}{\textbf{mIoU old}} & \scriptsize\rotatebox{90}{boat} &  \scriptsize\rotatebox{90}{plant} & \scriptsize\rotatebox{90}{sheep} & \scriptsize\rotatebox{90}{cow} & \scriptsize\rotatebox{90}{bottle}  & \scriptsize\rotatebox{90}{\textbf{mIoU new}} & \scriptsize\rotatebox{90}{\textbf{mIoU}} & \scriptsize\rotatebox{90}{\textbf{mPA}} & \scriptsize\rotatebox{90}{\textbf{mCA}}\\
 \hline

Fine-tuning & 91.2 & 78.8 & 81.6 & 69.5 & 80.6 & 69.6 & 28.6 & 85.6 & 41.4 & 74.9 & 57.2 & 77.4 & 38.5 & 60.0 & 34.6 & 64.3 & 64.6 & 27.4 & 18.3 & 1.6 & 46.9 & 49.7 & 28.8 & 56.1 & 90.7 & 66.0 \\

{\revision $E_F$} & 90.2 & 76.5 & 84.4 & 72.9 & 79.3 & 70.4 & 27.4 & 87.3 & 44.2 & 71.5 & 45.2 & 75.8 & 51.6 & 65.5 & 32.0 & 62.4 & 64.8 & 51.0 & 23.9 & 2.5 & 56.3 & 62.1 & 39.1 & 58.7 & 90.6 & 67.9 \\

$\mathcal{L}_{D}^{cls-T}$ & 92.1 & 81.4 & 69.6 & 66.6 & 81.5 & 81.8 & 26.9 & 87.5 & 37.0 & 71.9 & 47.0 & 79.1 & 46.2 & 69.0 & 34.8 & 65.6 & 64.9 & 40.1 & \textbf{35.3} & 11.0 & 37.0 & \textbf{66.1} & 37.9 & 58.5 & 90.9 & 56.8 \\

$E_F$, $\mathcal{L}_{D}^{cls-T}$ & 91.1 & 77.1 & 84.2 & 72.9 & 79.2 & 75.9 & 28.0 & 88.2 & 43.1 & 73.4 & 46.7 & 78.9 & 56.6 & 67.4 & 32.2 & 62.9 & 66.1 & \textbf{49.8} & 29.6 & \textbf{24.4} & 51.3 & 63.1 & \textbf{43.6} & 60.8 & 91.1 & \textbf{70.3} \\



$\mathcal{L}_{D}^{enc}$ & 91.6 & 82.0 & 88.8 & 80.2 & 83.8 & 76.8 & 28.8 & 91.9 & 50.0 & 74.0 & 54.9 & 80.6 & 66.0 & 72.6 & 36.1 & 69.9 & \textbf{70.5} & 34.1 & 18.4 & 4.8 & \textbf{53.0} & 56.9 & 33.4 & \textbf{61.7} & \textbf{92.0} & 70.2 \\

$\mathcal{L}_{D}^{dec}$ & 92.0 & 81.7 & 84.2 & 72.5 & 82.0 & 70.1 & 32.9 & 87.6 & 45.9 & 72.7 & 54.0 & 74.4 & 61.2 & 76.2 & 34.1 & 69.8 & 68.2 & 34.0 & 24.5 & 6.5 & 45.6 & 60.7 & 34.2 & 60.1 & 91.6 & 69.3 \\

$\mathcal{L}_{D}^{SPKD-avg}$ & 91.7 & 81.9 & 84.5 & 76.0 & 81.5 & 70.5 & 31.0 & 89.1 & 44.6 & 76.6 & 56.3 & 78.2 & 50.5 & 69.0 & 35.9 & 69.4 & 67.9 & 40.1 & 30.7 & 11.4 & 48.8 & 61.2 & 38.4 & 60.9 & 91.6 & \textbf{70.3} \\ \hline

$M_0(0-15)$ & 93.5 & 81.1 & 89.3 & 84.3 & 84.6 & 85.4 & 30.0 & 92.9 & 47.5 & 79.0 & 57.8 & 86.0 & 85.5 & 84.7 & 36.4 & 71.3 & 74.3 & - & - & - & - & - & - & 74.3 & 94.1 & 84.2 \\

 $M_0(0-20)$ & 93.4 & 83.6 & 89.7 & 82.4 & 82.4 & 86.4 & 35.4 & 92.7 & 51.1 & 76.8 & 52.9 & 85.4 & 82.0 & 85.7 & 36.7 & 70.5 & 75.1 & 63.3 & 52.3 & 82.4 & 80.9 & 78.7 & 68.5 & 73.6 & 93.9 & 84.2 \\
\hline
\end{tabular}
\end{table*}

The same scenario is then analyzed for classes ordered on the basis of their occurrence inside the dataset in Table~\ref{tab:pascal_0_15_16_17_18_19_20_occurrences}. Similar considerations as before hold, however the accuracy on new classes is slightly lower since less training samples are available for such classes. {\revision In this case the gain of mIoU is smaller: 
no single method is able to get a large improvement, however $\mathcal{L}_D^{enc}$ and $\mathcal{L}_D^{SPKD-avg}$ are the best performing approaches. A critical example is the behavior of the $\mathit{sheep}$ class, whose accuracy is highly reduced after the addition of the  correlated $\mathit{cow}$ class. 
 }

\begin{table*}[htbp]
\vspace{-0.1cm}
\caption{Per-class IoU of the proposed approaches on VOC2012 when $10$ classes are added sequentially.}
\vspace{-0.2cm}
\label{tab:pascal_0_10_11_12_13_14_15_16_17_18_19_20}
\setlength{\tabcolsep}{1.6pt}
\centering
\footnotesize
\begin{tabular}{|c|ccccccccccc:c|c:c:c:c:c:c:c:c:c:c:c|ccc|}
\hline
 $M_{10} (11\rightarrow20)$ & \scriptsize\rotatebox{90}{backgr.} &  \scriptsize\rotatebox{90}{aero} &  \scriptsize\rotatebox{90}{bike} &  \scriptsize\rotatebox{90}{bird} &\scriptsize\rotatebox{90}{boat} & \scriptsize\rotatebox{90}{bottle} & \scriptsize\rotatebox{90}{bus} 
  &\scriptsize\rotatebox{90}{car} & \scriptsize\rotatebox{90}{cat} & \scriptsize\rotatebox{90}{chair} & \scriptsize\rotatebox{90}{cow} & \scriptsize\rotatebox{90}{\textbf{mIoU old}} & \scriptsize\rotatebox{90}{din. table}& \scriptsize\rotatebox{90}{dog} & \scriptsize\rotatebox{90}{horse} 
  & \scriptsize\rotatebox{90}{mbike} & \scriptsize\rotatebox{90}{person} & \scriptsize\rotatebox{90}{plant} &  \scriptsize\rotatebox{90}{sheep} & \scriptsize\rotatebox{90}{sofa} & \scriptsize\rotatebox{90}{train} & \scriptsize\rotatebox{90}{tv} & \scriptsize\rotatebox{90}{\textbf{mIoU new}} & \scriptsize\rotatebox{90}{\textbf{mIoU}} & \scriptsize\rotatebox{90}{\textbf{mPA}} & \scriptsize\rotatebox{90}{\textbf{mCA}}\\
 \hline

Fine-tuning & 86.7 & 29.0 & 28.0 & 49.7 & 2.2 & 54.4 & 1.5 & 54.7 & 75.3 & 29.4 & 7.8 & 38.0 & 46.5 & 60.7 & 17.0 & 23.3 & 75.1 & 29.1 & 38.0 & 31.5 & 27.6 & 11.6 & 36.0 & 37.1 & 83.6 & 49.0 \\

{\revision $E_F$} & 87.9 & 65.2 & 28.6 & 73.1 & 56.8 & 70.0 & 73.4 & 77.5 & 80.1 & 26.9 & 54.5 & 63.1 & 46.8 & 56.4 & 44.1 & 40.2 & 74.5 & 36.8 & 35.6 & 27.4 & 58.7 & 28.4 & 44.9 & 54.4 & 85.8 & 59.4 \\

$\mathcal{L}_{D}^{cls-T}$ & 87.0 & 28.1 & 27.4 & 47.4 & 0.2 & 48.6 & 1.6 & 59.3 & 73.1 & 22.9 & 27.9 & 38.5 & 42.2 & 56.9 & 21.8 & 30.9 & 77.1 & 28.4 & 27.7 & 24.5 & 33.1 & 21.6 & 36.4 & 37.5 & 83.8 & 51.1 \\

$E_F$, $\mathcal{L}_{D}^{cls-T}$ & 89.5 & 66.1 & 28.3 & 72.7 & 58.3 & 70.7 & 74.0 & 78.2 & 80.3 & 27.7 & 55.1 & \textbf{63.7} & 45.7 & 56.2 & \textbf{45.6} & 41.5 & 74.8 & 37.2 & 36.9 & 26.7 & 59.2 & \textbf{28.6} & 45.2 & {\revision 54.9} & {\revision 88.5} & {\revision \textbf{67.1}} \\


$\mathcal{L}_{D}^{enc}$ & 89.1 & 54.7 & 28.7 & 75.3 & 44.5 & 69.4 & 73.2 & 79.4 & 83.4 & 30.2 & 54.9 & 62.1 & \textbf{48.5} & \textbf{62.4} & 38.7 & \textbf{48.5} & 75.0 & \textbf{40.8} & \textbf{52.9} & 28.5 & \textbf{59.8} & 24.1 & \textbf{47.9} & \textbf{55.3} & \textbf{88.5} & 66.7 \\

$\mathcal{L}_{D}^{dec}$ & 89.1 & 64.7 & 28.5 & 62.4 & 29.7 & 54.3 & 30.9 & 67.7 & 79.7 & 27.8 & 35.9 & 51.9 & 46.2 & 52.6 & 40.0 & 47.0 & \textbf{77.3} & 29.9 & 35.6 & \textbf{33.3} & 40.0 & 23.0 & 42.5 & 47.4 & 87.1 & 59.3 \\
 
 $\mathcal{L}_{D}^{SPKD-avg}$ & 88.9 & 64.6 & 29.6 & 68.6 & 35.8 & 64.9 & 76.8 & 76.8 & 74.5 & 31.0 & 16.8 & 57.1 & 46.2 & 55.1 & 26.1 & 34.0 & 77.0 & 34.5 & 33.9 & 33.1 & 56.4 & 25.2 & 42.1 & 50.0 & 87.8 & 61.8 
 \\\hline
 

$M_0(0-10)$ & 95.3 & 86.4 & 34.4 & 85.6 & 69.7 & 79.3 & 94.6 & 87.6 & 93.1 & 44.2 & 91.9 & 78.4 & - & - & - & - & - & - & - & - & - & - & - & 78.4 & 96.1 & 90.4
 \\
 
$M_0(0-20)$ & 93.4 & 85.4 & 36.7 & 85.7 & 63.3 & 78.7 & 92.7 & 82.4 & 89.7 & 35.4 & 80.9 & 74.9 & 52.9 & 82.4 & 82.0 & 76.8 & 83.6 & 52.3 & 82.4 & 51.1 & 86.4 & 70.5 & 72.1 & 73.6 & 93.9 & 84.2 \\
\hline
\end{tabular}
\end{table*}

As a final experiment, we  investigate the sequential addition of the last $10$ classes, i.e., $M_{10} (11 \rightarrow 20)$: the results of per-class IoU are shown in Table~\ref{tab:pascal_0_10_11_12_13_14_15_16_17_18_19_20}. In this case the best strategy is to apply distillation on the intermediate feature space ($\mathcal{L}_D^{enc}$), that allows to improve the mIoU of $18.2\%$ with respect to the standard scheme achieving quite good performance in this extreme case. The result is even more effective because the proposed method outperforms the standard approach by $11.9\%$ of mIoU on new classes and by $24.1\%$ on old ones.\\
Again, we can confirm that correlated classes highly influence each other results: for example the addition of the $\mathit{sheep}$ class (low IoU and high pixel accuracy) leads to a large degradation of the performance on the $\mathit{cow}$ class (low IoU and low pixel accuracy). The same happens to the $\mathit{bus}$ class when $\mathit{train}$ is added. This phenomenon is mitigated by the proposed modifications which improve the IoU of the $\mathit{cow}$ class from $7.8\%$ of the baseline to $55.1\%$ of the best proposed method and the IoU of the $\mathit{bus}$ class from $1.5\%$ to $74.0\%$.

{\revision
\subsection{ImageNet Pre-Training}
\label{sec:ablation_imagenet}

Semantic segmentation architectures are typically composed of an encoding and a decoding stage. The encoder is trained to learn useful but compact representations of the scene which are then processed by the decoder to produce the output segmentation map with a classification score for each pixel. Such architectures are highly complex and the weights of the encoder are always pre-trained on very large datasets: e.g., on ImageNet \citep{deng2009imagenet} or MSCOCO \citep{lin2014microsoft}. 
For incremental learning strategies it is important to ensure that the pre-training does not affect the incremental steps since it could contain  information concerning novel classes to be learned in the incremental steps.
To better investigate this issue, i.e., to ensure that the information learned during the pre-training is not affecting the incremental learning results, we performed a set of experiments using a different pre-training done on the ImageNet classification dataset.

The comparison in terms of mIoU is shown in Table~\ref{tab:ablation_imagenet} for the addition of the last class, of the last $10$ classes at once and of the last $10$ classes sequentially.
Pre-training on ImageNet leads to a slightly lower starting mIoU for all the scnarios, however the improvements obtained by the proposed strategies are similar to the ones with MSCOCO pre-training and so are their final rankings (i.e., the best approaches in the various scenarios are the same). Additionally, also the gap with respect to the fine-tuning approach is coherent between the two pre-training strategies.

In conclusion, using a different pre-training does not change the effectiveness of the proposed strategies and the general discussion of the results remains unaltered. Further considerations and results for this setting are shown in the \textit{Supplementary Material}.

\begin{table}[htbp]
\vspace{-0.1cm}
\caption{Ablation study comparing ImageNet (mIoU$_I$) and MSCOCO (mIoU$_M$) pre-training on VOC2012.}
\vspace{-0.2cm}
\label{tab:ablation_imagenet}
\setlength{\tabcolsep}{2pt}
\centering
\footnotesize
    \begin{tabular}{|l|cc|cc|cc|}
    \hline
          & \multicolumn{2}{c|}{$M_1(20)$} & \multicolumn{2}{c|}{$M_1(11-20)$} & \multicolumn{2}{c|}{$M_{10}(11\rightarrow 20)$} \\\cline{2-7}
          & mIoU$_I$ & mIoU$_M$ & mIoU$_I$ & mIoU$_M$ & mIoU$_I$ & mIoU$_M$  \\\hline
    Fine-tuning    & 62.8  & 66.5  & 64.3  & 65.6  & 35.9  & 37.1 \\
    $E_F$    & 68.8  & 71    & 62.2  & 64.1  & 50.9  & 54.5 \\
    $\mathcal{L}_{D}^{cls-T}$  & 68.9  & 71.5  & \textbf{65.5} & \textbf{66.3}  & 36.5  & 37.5 \\
    $E_F$, $\mathcal{L}_{D}^{cls-T}$ & \textbf{70.9} & \textbf{71.8}  & 63.7  & 64.3  & 54.0  & 54.9 \\
    $\mathcal{L}_{D}^{enc}$  & 70.6  & 71.5  & 65.2  & 66    & \textbf{54.5} & \textbf{55.3} \\
    $\mathcal{L}_{D}^{dec}$ & 68.1  & 70.0  & 64.8  & 65.7  & 46.3  & 47.4 \\
    $\mathcal{L}_{D}^{SPKD-avg}$ & 68.8  & 71.0  & 64.5  & 65.6  & 48.8  & 50 \\\hline
    \end{tabular}%
\end{table}%

}


\subsection{Experimental Results on MSRC-v2}
\label{sec:results_MSRC}

\begin{table*}[htbp]
\vspace{-0.1cm}
\caption{Per-class IoU of the proposed approaches on MSRC-v2 when the last class, i.e., $\mathit{boat}$, is added}
\vspace{-0.2cm}
\label{tab:MSRC_0_19_20}
\setlength{\tabcolsep}{1.6pt}
\centering
\footnotesize
\begin{tabular}{|c|cccccccccccccccccccc:c|c|ccc|}
\hline
 $M_1 (20)$ & \scriptsize\rotatebox{90}{grass} &  \scriptsize\rotatebox{90}{building} &  \scriptsize\rotatebox{90}{sky} &  \scriptsize\rotatebox{90}{road} &\scriptsize\rotatebox{90}{tree} & \scriptsize\rotatebox{90}{water} & \scriptsize\rotatebox{90}{book} 
  &\scriptsize\rotatebox{90}{car} & \scriptsize\rotatebox{90}{cow} & \scriptsize\rotatebox{90}{bicycle} & \scriptsize\rotatebox{90}{flower} & \scriptsize\rotatebox{90}{body}& \scriptsize\rotatebox{90}{sheep} & \scriptsize\rotatebox{90}{sign} 
  & \scriptsize\rotatebox{90}{face} & \scriptsize\rotatebox{90}{cat} & \scriptsize\rotatebox{90}{chair} &  \scriptsize\rotatebox{90}{aeroplane} & \scriptsize\rotatebox{90}{dog} & \scriptsize\rotatebox{90}{bird}& \scriptsize\rotatebox{90}{\textbf{mIoU old}} & \scriptsize\rotatebox{90}{boat} & \scriptsize\rotatebox{90}{\textbf{mIoU}} & \scriptsize\rotatebox{90}{\textbf{mPA}} & \scriptsize\rotatebox{90}{\textbf{mCA}}\\
 \hline

Fine-tuning &  93.4 & 79.9 & 92.9 & 70.8 & 86.7 & 79.3 & 94.5 & 93.2 & 86.8 & 91.9 & 95.4 & 83.0 & 84.0 & 91.3 & 87.9 & 79.5 & 89.1 & 78.3 & 74.0 & 74.3 & 85.3 & 51.2 & 83.7 & 92.4 & 89.6 \\

{\revision $E_F$} & 94.3 & 82.2 & 92.7 & 58.8 & 86.2 & 72.7 & 98.5 & 91.3 & 89.0 & 89.2 & 97.7 & 84.1 & 89.7 & 93.3 & 88.0 & 93.6 & 94.7 & 85.3 & 89.5 & 78.3 & 87.5 & 56.0 & 86.0 & 92.5 & 91.7 \\

$\mathcal{L}_{D}^{cls-T}$ &  94.7 & 82.8 & 93.5 & 83.4 & 87.5 & 87.4 & 98.1 & 94.4 & 89.3 & 91.5 & 97.8 & 83.2 & 88.2 & 93.9 & 86.0 & 84.2 & 96.2 & 85.2 & 80.0 & 82.6 & 89.0 & 59.3 & 87.6 & 94.5 & 92.9 \\

$E_F$, $\mathcal{L}_{D}^{cls-T}$ & 94.8 & 83.4 & 93.4 & 81.7 & 87.3 & 86.6 & 98.6 & 92.7 & 89.4 & 89.8 & 97.9 & 83.9 & 89.5 & 93.8 & 87.3 & 94.2 & 96.4 & 86.3 & 91.0 & 83.5 & \textbf{90.1} & 58.2 & \textbf{88.6} & \textbf{94.6} & \textbf{93.7} \\

$\mathcal{L}_{D}^{enc}$ & 94.9 & 82.3 & 92.7 & 77.9 & 87.5 & 84.4 & 98.5 & 92.3 & 88.6 & 90.4 & 97.7 & 84.5 & 87.5 & 92.8 & 88.6 & 93.6 & 97.3 & 84.5 & 88.3 & 81.2 & 89.3 & 54.6 & 87.6 & 94.1 & 92.8 \\

$\mathcal{L}_{D}^{dec}$ & 94.8 & 83.7 & 94.1 & 80.8 & 88.1 & 84.6 & 97.5 & 90.8 & 86.6 & 88.2 & 97.9 & 84.9 & 82.8 & 95.4 & 88.6 & 76.1 & 94.6 & 83.3 & 71.2 & 81.3 & 87.2 & 40.8 & 85.0 & 93.9 & 91.1 \\
 
 $\mathcal{L}_{D}^{SPKD-avg}$ & 94.3 & 81.9 & 93.2 & 63.5 & 87.2 & 74.5 & 98.7 & 91.2 & 88.9 & 89.6 & 97.9 & 85.9 & 87.8 & 91.0 & 91.4 & 74.0 & 92.7 & 82.2 & 68.0 & 79.9 & 85.7 & \textbf{66.9} & 84.8 & 92.4 & 91.1  
 \\ \hline

$M_0(0-19)$ &  94.9 & 84.7 & 93.3 & 88.7 & 88.9 & 90.9 & 98.6 & 94.4 & 89.0 & 89.6 & 98.0 & 83.7 & 89.6 & 94.1 & 86.7 & 95.0 & 95.4 & 86.2 & 92.4 & 82.1 & 90.8 & -	& 90.8 & 95.5 & 96.0 \\

$M_0(0-20)$ & 94.8 & 82.3 & 94.6 & 87.3 & 88.8 & 92.5 & 98.6 & 94.1 & 90.9 & 89.7 & 98.0 & 87.4 & 92.0 & 91.2 & 89.9 & 93.9 & 95.9 & 84.8 & 90.3 & 87.3 & 86.7 & 75.1 & 90.5 & 95.4	& 95.5 \\
\hline
\end{tabular}
\end{table*}

\begin{table*}[htbp]
\vspace{-0.1cm}
\caption{Per-class IoU of the proposed approaches on MSRC-v2 when $5$ classes are added at once.}
\vspace{-0.2cm}
\label{tab:MSRC_0_15_20}
\setlength{\tabcolsep}{1.6pt}
\centering
\footnotesize
\begin{tabular}{|c|cccccccccccccccc:c|ccccc:c|ccc|}
\hline
 $M_1 (16-20)$ & \scriptsize\rotatebox{90}{grass} &  \scriptsize\rotatebox{90}{building} &  \scriptsize\rotatebox{90}{sky} &  \scriptsize\rotatebox{90}{road} &\scriptsize\rotatebox{90}{tree} & \scriptsize\rotatebox{90}{water} & \scriptsize\rotatebox{90}{book} 
  &\scriptsize\rotatebox{90}{car} & \scriptsize\rotatebox{90}{cow} & \scriptsize\rotatebox{90}{bicycle} & \scriptsize\rotatebox{90}{flower} & \scriptsize\rotatebox{90}{body}& \scriptsize\rotatebox{90}{sheep} & \scriptsize\rotatebox{90}{sign} 
  & \scriptsize\rotatebox{90}{face} & \scriptsize\rotatebox{90}{cat} & \scriptsize\rotatebox{90}{\textbf{mIoU old}} & \scriptsize\rotatebox{90}{chair} &  \scriptsize\rotatebox{90}{aeroplane} & \scriptsize\rotatebox{90}{dog} & \scriptsize\rotatebox{90}{bird} & \scriptsize\rotatebox{90}{boat} & \scriptsize\rotatebox{90}{\textbf{mIoU new}} & \scriptsize\rotatebox{90}{\textbf{mIoU}} & \scriptsize\rotatebox{90}{\textbf{mPA}} & \scriptsize\rotatebox{90}{\textbf{mCA}}\\
 \hline

Fine-tuning & 91.9 & 79.5 & 93.9 & 83.9 & 84.1 & 88.5 & 96.0 & 94.0 & 49.0 & 91.6 & 93.8 & 82.9 & 59.3 & 87.3 & 90.4 & 37.5 & 81.5 & 86.3 & 74.9 & 43.6 & 66.4 & 63.5 & 67.0 & 78.0 & 91.1 & 85.2
 \\

{\revision $E_F$} & 93.8 & 80.3 & 92.0 & 83.1 & 86.0 & 84.1 & 97.9 & 92.3 & 81.0 & 87.7 & 97.1 & 64.8 & 78.2 & 96.5 & 89.1 & 59.0 & 85.2 & 89.4 & 70.1 & 52.2 & 66.8 & 53.8 & 66.5 & 80.7 & 92.1 & 86.2 \\

$\mathcal{L}_{D}^{cls-T}$ & 93.5 & 81.0 & 91.5 & 81.9 & 86.6 & 84.8 & 98.4 & 94.2 & 86.3 & 90.5 & 96.5 & 77.3 & 81.8 & 95.8 & 89.6 & 69.8 & 87.5 & 91.0 & 78.1 & 69.0 & 69.9 & 67.6 & 75.1 & 84.5 & 93.3 & \textbf{90.8}\\

$E_F$, $\mathcal{L}_{D}^{cls-T}$ & 94.4 & 80.7 & 91.6 & 83.2 & 86.7 & 83.5 & 98.8 & 92.6 & 80.7 & 88.5 & 97.0 & 66.3 & 78.8 & 96.1 & 89.2 & 58.8 & 85.4 & 89.7 & 67.4 & 52.8 & 67.4 & 54.5 & 66.4 & 80.9 & 92.4 & 87.6 \\

$\mathcal{L}_{D}^{enc}$ & 93.3 & 80.4 & 92.8 & 85.6 & 87.5 & 88.9 & 98.7 & 94.0 & 72.1 & 89.1 & 95.7 & 80.8 & 81.4 & 91.2 & 86.4 & 57.2 & 85.9 & 87.7 & 72.5 & 62.4 & 71.1 & 66.5 & 72.0 & 82.6 & 93.1 & 88.5 \\

$\mathcal{L}_{D}^{dec}$ & 94.8 & 81.6 & 93.4 & 85.4 & 87.8 & 88.0 & 95.8 & 94.0 & 87.4 & 87.8 & 96.0 & 84.9 & 83.1 & 91.5 & 89.0 & 68.2 & \textbf{88.0} & 88.3 & 81.4 & 70.1 & 78.5 & 59.8 & \textbf{75.6} & \textbf{85.1} & \textbf{93.9} & \textbf{90.8}  \\
 
 $\mathcal{L}_{D}^{SPKD-avg}$ & 92.3 & 79.5 & 93.5 & 83.5 & 85.7 & 88.9 & 95.7 & 94.0 & 73.8 & 91.7 & 93.9 & 84.4 & 72.8 & 87.9 & 90.0 & 45.6 & 84.6 & 84.2 & 67.8 & 55.1 & 71.4 & 59.1 & 67.5 & 80.5 & 92.2 & 86.7  
 \\ \hline

$M_0(0-15)$ & 94.2 & 83.5 & 89.7 & 85.7 & 88.9 & 88.3 & 98.9 & 94.5 & 88.8 & 89.0 & 98.0 & 85.0 & 91.6 & 95.1 & 89.8 & 96.4 & 91.1 & - & - & - & - & - & - &	91.1 & 95.2 & 96.0
 \\

$M_0(0-20)$ & 94.8 & 82.3 & 94.6 & 87.3 & 88.8 & 92.5 & 98.6 & 94.1 & 90.9 & 89.7 & 98.0 & 87.4 & 92.0 & 91.2 & 89.9 & 93.9 & 91.6 & 95.9 & 84.8 & 90.3 & 87.3 & 75.1 & 86.7 & 90.5 & 95.4 & 95.5 \\
\hline
\end{tabular}
\end{table*}

\begin{table*}[htbp]
\vspace{-0.1cm}
\caption{Per-class IoU of the proposed approaches on MSRC-v2 when $5$ classes are added sequentially.}
\vspace{-0.2cm}
\label{tab:MSRC_0_15_16_17_18_19_20}
\setlength{\tabcolsep}{1.6pt}
\centering
\footnotesize
\begin{tabular}{|c|cccccccccccccccc:c|c:c:c:c:c:c|ccc|}
\hline
 $M_5 (16\rightarrow20)$ & \scriptsize\rotatebox{90}{grass} &  \scriptsize\rotatebox{90}{building} &  \scriptsize\rotatebox{90}{sky} &  \scriptsize\rotatebox{90}{road} &\scriptsize\rotatebox{90}{tree} & \scriptsize\rotatebox{90}{water} & \scriptsize\rotatebox{90}{book} 
  &\scriptsize\rotatebox{90}{car} & \scriptsize\rotatebox{90}{cow} & \scriptsize\rotatebox{90}{bicycle} & \scriptsize\rotatebox{90}{flower} & \scriptsize\rotatebox{90}{body}& \scriptsize\rotatebox{90}{sheep} & \scriptsize\rotatebox{90}{sign} 
  & \scriptsize\rotatebox{90}{face} & \scriptsize\rotatebox{90}{cat} & \scriptsize\rotatebox{90}{\textbf{mIoU old}} & \scriptsize\rotatebox{90}{chair} &  \scriptsize\rotatebox{90}{aeroplane} & \scriptsize\rotatebox{90}{dog} & \scriptsize\rotatebox{90}{bird} & \scriptsize\rotatebox{90}{boat} & \scriptsize\rotatebox{90}{\textbf{mIoU new}} & \scriptsize\rotatebox{90}{\textbf{mIoU}} & \scriptsize\rotatebox{90}{\textbf{mPA}} & \scriptsize\rotatebox{90}{\textbf{mCA}}\\
 \hline

Fine-tuning & 86.3 & 71.8 & 93.4 & 69.4 & 77.1 & 67.0 & 90.7 & 71.8 & 53.0 & 88.7 & 79.6 & 69.1 & 49.6 & 74.8 & 90.3 & 34.5 & 72.9 & 34.2 & 51.6 & 35.6 & 17.7 & 35.5 & 34.9 & 63.9 & 83.8 & 73.2
 \\
 
 {\revision $E_F$} & 88.6 & 74.4 & 93.2 & 72.1 & 80.7 & 74.3 & 94.9 & 90.0 & 70.4 & 88.9 & 90.3 & 68.5 & 64.1 & 85.3 & 87.1 & 59.4 & 80.1 & 57.2 & 51.1 & 30.1 & 23.5 & 36.3 & 39.6 & 70.5 & 87.2 & 76.3 \\

$\mathcal{L}_{D}^{cls-T}$ & 92.7 & 78.9 & 90.7 & 74.5 & 86.2 & 71.6 & 97.8 & 93.7 & 88.7 & 90.8 & 95.5 & 67.7 & 77.4 & 93.4 & 88.2 & 64.9 & 84.5 & \textbf{83.4} & \textbf{69.8} & \textbf{47.2} & 41.3 & \textbf{54.7} & \textbf{59.3} & \textbf{78.5} & 90.7 & \textbf{86.9}
 \\

$E_F$, $\mathcal{L}_{D}^{cls-T}$ &  92.9 & 79.5 & 90.4 & 81.2 & 85.3 & 81.2 & 99.0 & 93.3 & 82.2 & 88.7 & 96.0 & 67.6 & 67.5 & 92.8 & 87.6 & 79.5 & 85.3 & 82.0 & 50.3 & 29.7 & 35.1 & 40.2 & 47.5 & 76.3 & 90.6 & 84.8
 \\

$\mathcal{L}_{D}^{enc}$ & 91.5 & 78.7 & 91.5 & 75.8 & 86.6 & 77.9 & 98.6 & 94.1 & 84.6 & 89.8 & 97.0 & 80.1 & 74.6 & 87.1 & 86.3 & 78.0 & 85.8 & 75.3 & 46.8 & 30.7 & 43.2 & 31.7 & 45.5 & 76.2 & 90.5 & 83.3
 \\

$\mathcal{L}_{D}^{dec}$ & 93.2 & 81.9 & 93.5 & 79.3 & 87.4 & 77.4 & 98.2 & 90.4 & 86.4 & 85.0 & 95.1 & 78.1 & 70.7 & 93.2 & 84.2 & 85.1 & \textbf{86.2} & 41.6 & 61.7 & 40.9 & \textbf{68.3} & 16.2 & 45.7 & 76.6 & \textbf{91.1} & 84.5
 \\
 
 $\mathcal{L}_{D}^{SPKD-avg}$ & 91.2 & 78.4 & 94.7 & 62.0 & 81.1 & 64.8 & 96.4 & 91.1 & 22.6 & 88.6 & 94.2 & 82.5 & 47.2 & 85.0 & 91.0 & 30.0 & 75.0 & 55.3 & 51.4 & 36.2 & 16.9 & 42.0 & 40.3 & 66.8 & 85.4 & 75.6
 
 \\ \hline

$M_0(0-15)$ & 94.2 & 83.5 & 89.7 & 85.7 & 88.9 & 88.3 & 98.9 & 94.5 & 88.8 & 89.0 & 98.0 & 85.0 & 91.6 & 95.1 & 89.8 & 96.4 & 91.1 & - & - & - & - & - & - &	91.1 & 95.2 & 96.0
 \\

$M_0(0-20)$ & 94.8 & 82.3 & 94.6 & 87.3 & 88.8 & 92.5 & 98.6 & 94.1 & 90.9 & 89.7 & 98.0 & 87.4 & 92.0 & 91.2 & 89.9 & 93.9 & 91.6 & 95.9 & 84.8 & 90.3 & 87.3 & 75.1 & 86.7 & 90.5 & 95.4 & 95.5 \\
\hline
\end{tabular}
\end{table*}


Finally, we briefly show the per-class IoU results obtained on the MSRC-v2 dataset. We sort the classes according to their occurrence inside the dataset and we perform three experiments: we add the last class, the last 5 in a single shot or the last 5 sequentially. The results are shown in Tables~\ref{tab:MSRC_0_19_20}, \ref{tab:MSRC_0_15_20} and \ref{tab:MSRC_0_15_16_17_18_19_20}, respectively, while additional results on the per-class pixel accuracy are in the \textit{Supplementary Material}.

\begin{figure*}[htbp]{}
\setlength\tabcolsep{1.5pt} 
\centering
\subfloat{
\begin{tabular}{cccccccc}
  {$\scriptstyle RGB$} &  {$\scriptstyle GT$} & {$\scriptstyle Fine-tuning$} &
{$ \scriptstyle M_1(20) [ E_F , \mathcal{L}_{D}^{cls-T} ]$}&
   {$\scriptstyle M_1(20) [ \mathcal{L}_{D}^{enc}]$}&
   {$\scriptstyle M_1(20) [ \mathcal{L}_{D}^{dec}]$}&
   {$\scriptstyle M_1(20) [ \mathcal{L}_{D}^{SPKD-avg}]$}&
 {$\scriptstyle M_0(0-20)$}\\
 
   \includegraphics[width=\sizefiggg\linewidth]{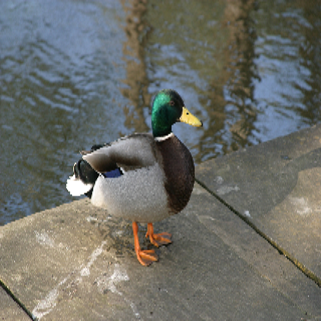} &
   \includegraphics[width=\sizefiggg\linewidth]{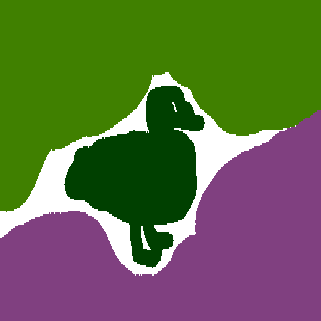} &
   \includegraphics[width=\sizefiggg\linewidth]{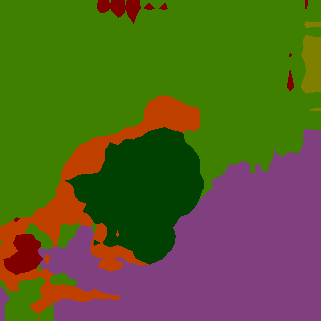} &
   \includegraphics[width=\sizefiggg\linewidth]{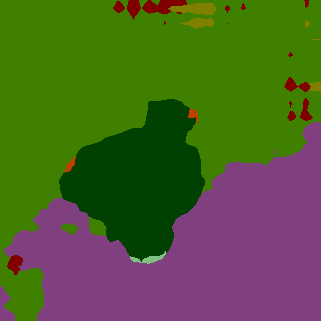} &
   \includegraphics[width=\sizefiggg\linewidth]{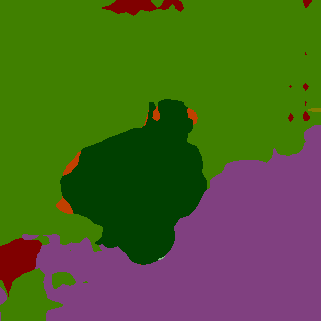} &
   \includegraphics[width=\sizefiggg\linewidth]{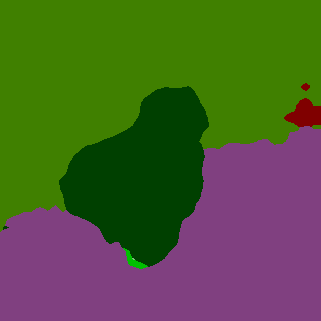} &
   \includegraphics[width=\sizefiggg\linewidth]{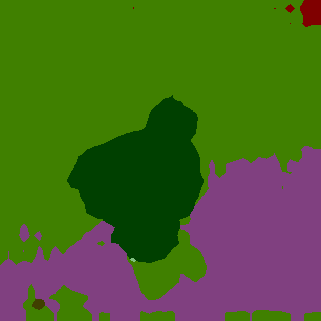} &
   \includegraphics[width=\sizefiggg\linewidth]{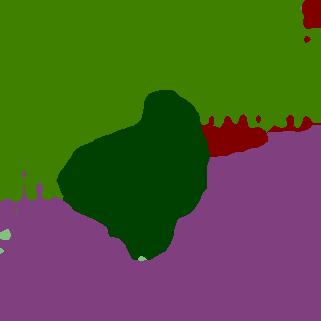} \\
   
   {$\scriptstyle RGB$} &  {$\scriptstyle GT$} & {$\scriptstyle Fine-tuning$} &
{$ \scriptstyle M_1(16-20) [ E_F , \mathcal{L}_{D}^{cls-T} ]$}&
   {$\scriptstyle M_1(16-20) [ \mathcal{L}_{D}^{enc}]$}&
   {$\scriptstyle M_1(16-20) [ \mathcal{L}_{D}^{dec}]$}&
   {$\scriptstyle M_1(16-20) [ \mathcal{L}_{D}^{SPKD-avg}]$}&
 {$\scriptstyle M_0(0-20)$}\\
   \includegraphics[width=\sizefiggg\linewidth]{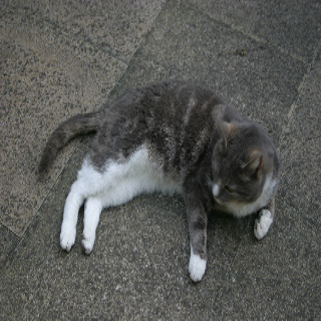} &
   \includegraphics[width=\sizefiggg\linewidth]{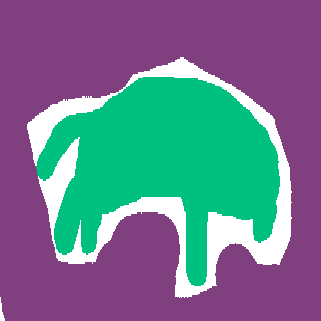} &
   \includegraphics[width=\sizefiggg\linewidth]{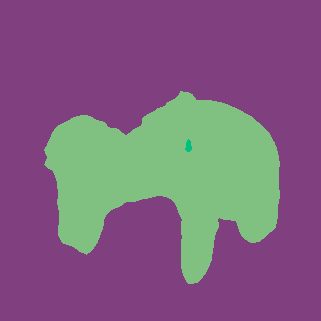} &
   \includegraphics[width=\sizefiggg\linewidth]{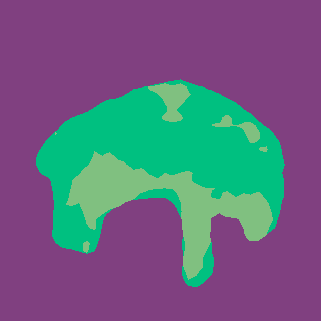} &
   \includegraphics[width=\sizefiggg\linewidth]{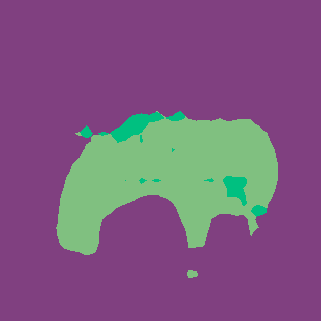} &
   \includegraphics[width=\sizefiggg\linewidth]{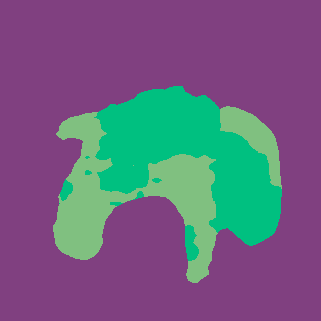} &
   \includegraphics[width=\sizefiggg\linewidth]{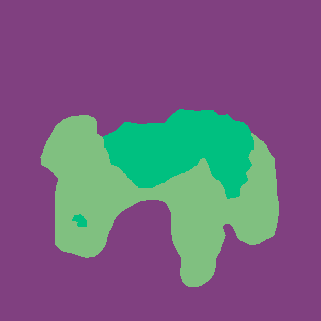} &
   \includegraphics[width=\sizefiggg\linewidth]{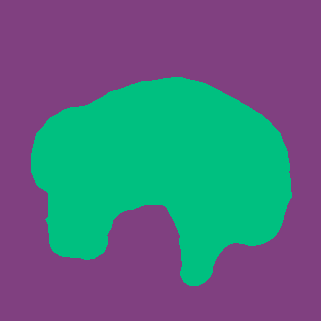} \\

   {$\scriptstyle RGB$} &  {$\scriptstyle GT$} & {$\scriptstyle Fine-tuning$} &
{$ \scriptstyle M_5(16 \rightarrow 20) [ E_F , \mathcal{L}_{D}^{cls-T} ]$}&
   {$\scriptstyle M_5(16 \rightarrow 20) [ \mathcal{L}_{D}^{enc}]$}&
   {$\scriptstyle M_5(16 \rightarrow 20) [ \mathcal{L}_{D}^{dec}]$}&
   {$\scriptstyle M_5(16 \rightarrow 20) [ \mathcal{L}_{D}^{SPKD-avg}]$}&
 {$\scriptstyle M_0(0-20)$}\\
   \includegraphics[width=\sizefiggg\linewidth]{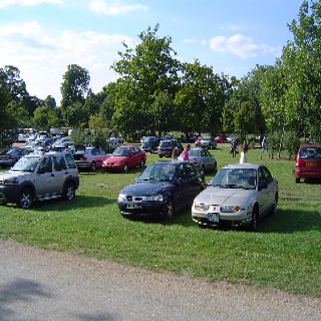} &
   \includegraphics[width=\sizefiggg\linewidth]{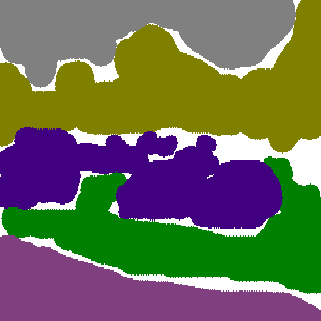} &
   \includegraphics[width=\sizefiggg\linewidth]{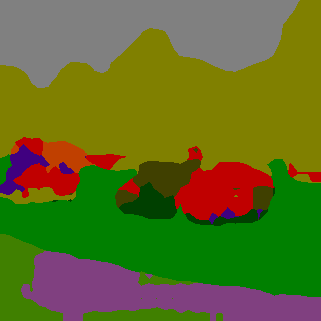} &
   \includegraphics[width=\sizefiggg\linewidth]{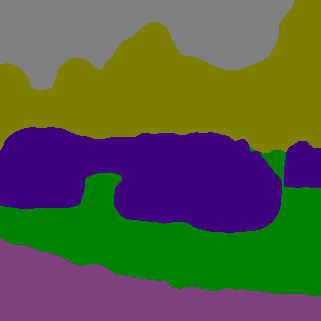} &
   \includegraphics[width=\sizefiggg\linewidth]{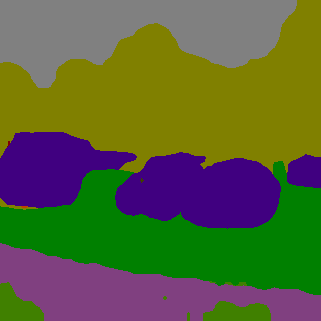} &
   \includegraphics[width=\sizefiggg\linewidth]{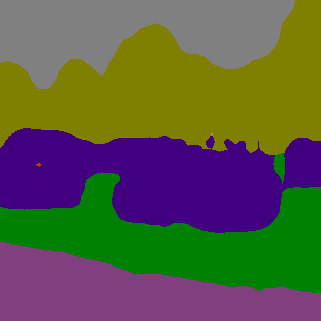} &
   \includegraphics[width=\sizefiggg\linewidth]{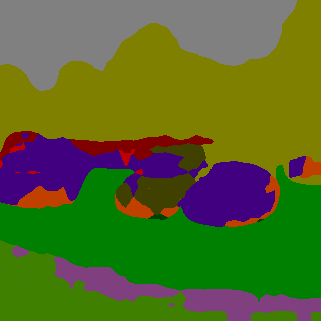} &
   \includegraphics[width=\sizefiggg\linewidth]{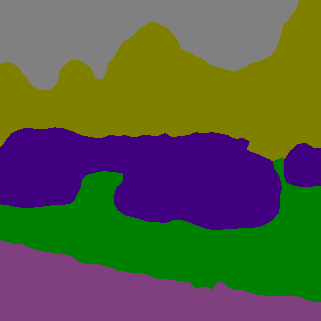} \\

 \end{tabular}
}
\centering
\vspace{-0.54cm}
\subfloat{
\hspace{0.27cm}
\resizebox{10cm}{!}{%
\begin{tabular}{cccccccccccccc}
& & & & & & & & & \\ \cline{14-14}
\cellcolor[HTML]{C00000}{\color[HTML]{FFFFFF} \textbf{aeroplane}} & 
\cellcolor[HTML]{004000}{\color[HTML]{FFFFFF} \textbf{bird}} & 
\cellcolor[HTML]{C04000}{\color[HTML]{FFFFFF} \textbf{boat}} & 
\cellcolor[HTML]{404000}{\color[HTML]{FFFFFF} \textbf{body}} & 
\cellcolor[HTML]{800000}{\color[HTML]{FFFFFF} \textbf{building}} & 
\cellcolor[HTML]{400080}{\color[HTML]{FFFFFF} \textbf{car}} & 
\cellcolor[HTML]{00C080}\textbf{cat} & 
\cellcolor[HTML]{80C080}\textbf{dog} & 
\cellcolor[HTML]{008000}{\color[HTML]{FFFFFF} \textbf{grass}} & 
\cellcolor[HTML]{804080}{\color[HTML]{FFFFFF} \textbf{road}} & 
\cellcolor[HTML]{808080}{\color[HTML]{FFFFFF} \textbf{sky}} & 
\cellcolor[HTML]{808000}{\color[HTML]{FFFFFF} \textbf{tree}} & 
\multicolumn{1}{c|}{\cellcolor[HTML]{408000}{\color[HTML]{FFFFFF} \textbf{water}}} &
\multicolumn{1}{c|}{\textbf{unlabeled}}\\ \cline{14-14} 
\end{tabular}%
}
}%
\vspace{-0.1cm}
\caption{Qualitative results on sample scenes on MSRC-v2 (\textit{best viewed in colors}). The first row regards the addition of the last class (i.e., \textit{boat}), the second row regards the addition of the last 5 classes at once, the third row regards the addition of the last 5 classes sequentially. The classes added are respectively \textit{chair}, \textit{aeroplane}, \textit{dog}, \textit{bird}, \textit{boat}.}
\label{fig:MSRC_qualitative}
\end{figure*}

Table~\ref{tab:MSRC_0_19_20}  shows the results for the addition of the last class, i.e., \textit{boat}. The average accuracies are higher on this dataset, however also in this case the fine-tuning approach is surpassed by all the proposed methods, proving the effectiveness of the distillation strategies. As already noticed in Table~\ref{tab:pascal_0_19_20}, the best method when adding only one class  is $M_1(20)[\mathcal{L}_D^{cls-T},E_F]$. The best strategy to learn the new class, instead, is  $M_1(20)[\mathcal{L}_D^{SPKD-avg}]$ which significantly outperforms all the other methods. A qualitative example is shown in the first row of Figure~\ref{fig:MSRC_qualitative}, where we can notice how the fine-tuning tends to find  $\mathit{boat}$ samples close to the water, while no boat is present in this image. This artifact is reduced by $\mathcal{L}_D^{cls-T}$ and $\mathcal{L}_D^{enc}$ and completely removed by $\mathcal{L}_D^{dec}$ and $\mathcal{L}_D^{SPKD-avg}$.

The second scenario  regards the addition of the last five classes at once (Table~\ref{tab:MSRC_0_15_20}). Here the best strategy is $M_1(16-20)[\mathcal{L}_D^{dec}]$ with an overall gap of $7.1\%$ of mIoU. 
The most challenging aspect on this dataset is the recognition of new classes in place of similar old ones; especially the $\mathit{cat}$ and $\mathit{cow}$ classes are frequently exchanged for the newly introduced $\mathit{dog}$ class (as can be verified looking at the per-class IoU and pixel accuracies). For example, the pixel accuracy of the cat raises from $37.5\%$ of fine-tuning to $71\%$ for the best proposed approach. On this dataset the classes appear mainly alone or with few other classes in the images, thus it is less likely to observe the effects of the correlation between classes belonging to similar contexts. 
In the visual example in second row of Figure~\ref{fig:MSRC_qualitative}  the fine-tuning approach misled the cat as a dog (which is among the classes being added). The issue is  mitigated by the proposed strategies that are able to detect at least part of the object as a cat.

As third experiment, we consider the sequential addition of five classes in Table~\ref{tab:MSRC_0_15_16_17_18_19_20}. Here the drop in accuracy is larger and the performance on some newly introduced classes are poor due to the limited number of samples in this smaller dataset.  We can appreciate how the best proposed strategies, $\mathcal{L}_D^{cls-T}$ and $\mathcal{L}_D^{dec}$, are able to largely outperform the fine-tuning approach by a huge margin.
 In the visual example in  row  3 of Fig.~\ref{fig:MSRC_qualitative}, the \textit{cars} are misled with \textit{boats} and \textit{aeroplanes} when fine-tuning, while the proposed strategies almost completely solve the problem.



{\revision
\subsection{Incremental Labeling on VOC2012}
\label{sec:results_labeling}

In some applications, we may be interested in performing incremental steps with images where only the new classes are annotated, and previously seen classes are set as background. This scenario could be extremely helpful to save time and resources on the annotation of previous classes.

Interestingly, the problem could be tackled with the same set of proposed techniques, which prove to be of quite general application. Nevertheless, the difference from the incremental learning protocol is twofold: first, the background class changes distribution at every incremental step; second, less information is present in the new images.

We analyze this experimental protocol in three cases: namely, the addition of the last class (in Table~\ref{tab:labeling_pascal_0_19_20}), the addition of the last $10$ classes at once (in Table~\ref{tab:labeling_pascal_0_10_20}), and the addition of the last $10$ classes sequentially one at a time (in Table~\ref{tab:labeling_pascal_0_10__20}).

In general, we can observe that, as expected, the final mIoU results are typically lower than in the previous experimental setting. However, it is important to notice that the ranking of the proposed methods remains substantially the same as in the incremental learning scenario: i.e., $E_F$, $\mathcal{L}_{D}^{cls-T}$ achieves a mIoU of $70.7\%$ and outperforms the other methods when adding the last class (compare Table~\ref{tab:labeling_pascal_0_19_20} with Table~\ref{tab:pascal_0_19_20}); $\mathcal{L}_{D}^{cls-T}$ outperforms the other proposals with a mIoU of $65.9\%$ when adding $10$ classes at once (compare Table~\ref{tab:labeling_pascal_0_10_20} with Table~\ref{tab:pascal_0_10_20}); finally $\mathcal{L}_{D}^{enc}$ performs best in the sequential addition of $10$ classes with a mIoU of $56.7\%$ (compare Table~\ref{tab:labeling_pascal_0_10__20} with Table~\ref{tab:pascal_0_10_11_12_13_14_15_16_17_18_19_20}).

\begin{table}[htbp]
\vspace{-0.1cm}
\caption{Incremental labeling results of the proposed approaches on VOC2012 in terms of mIoU when the last class, i.e., the \textit{tv/monitor} class, is added.}
\vspace{-0.2cm}
\label{tab:labeling_pascal_0_19_20}
\setlength{\tabcolsep}{2.1pt}
\centering
\footnotesize
\begin{tabular}{|c|c|c|c|}
\hline
$M_1 (20)$ & mIoU old & mIoU new & mIoU \\ \hline

Fine-tuning & 65.8 & 18.2 & 63.5 \\

$E_F$ & 71.1 & 36.5 & 69.4 \\

$\mathcal{L}_{D}^{cls-T}$ & 69.4 & 29.5 & 67.5 \\

$E_F$, $\mathcal{L}_{D}^{cls-T}$ & \textbf{72.0} & 44.9 & \textbf{70.7} \\

$\mathcal{L}_{D}^{enc}$ & 71.6 & \textbf{48.3} & \textbf{70.7} \\

$\mathcal{L}_{D}^{dec}$  & 70.5 & 32.8 & 68.7 \\
  
 $\mathcal{L}_{D}^{SPKD-avg}$ & 70.7 & 43.0 & 69.4
  \\\hline

$M_0(0-19)$ & 73.4  & - & 73.4 \\

$M_0(0-20)$ & 73.7  & 70.5  & 73.6 \\
\hline
\end{tabular}
\end{table}
\begin{table}[htbp]
\vspace{-0.1cm}
\caption{Incremental labeling results of the proposed approaches on VOC2012 in terms of mIoU when the $10$ classes are added at once.}
\vspace{-0.2cm}
\label{tab:labeling_pascal_0_10_20}
\setlength{\tabcolsep}{2.1pt}
\centering
\footnotesize
\begin{tabular}{|c|c|c|c|}
\hline
$M_1 (11-20)$ & mIoU old & mIoU new & mIoU \\ \hline

Fine-tuning & 65.8 & \textbf{63.0} & 64.4 \\

$E_F$ & 67.9 & 58.5 & 63.4 \\

$\mathcal{L}_{D}^{cls-T}$ & 68.5 & \textbf{63.0} & \textbf{65.9} \\

$E_F$, $\mathcal{L}_{D}^{cls-T}$ & \textbf{68.8} & 58.8 & 64.0 \\

$\mathcal{L}_{D}^{enc}$ & 67.1 & 62.7 & 65.0 \\

$\mathcal{L}_{D}^{dec}$  & 67.5 & 62.2 & 64.9 \\
  
 $\mathcal{L}_{D}^{SPKD-avg}$ & 65.9 & 62.7 & 64.4
  \\\hline

$M_0(0-10)$ & 78.4  & - & 78.4 \\

$M_0(0-20)$ & 74.9  & 72.1  & 73.6 \\
\hline
\end{tabular}
\end{table}
\begin{table}[htbp]
\vspace{-0.1cm}
\caption{Incremental labeling results of the proposed approaches on VOC2012 in terms of mIoU when the $10$ classes are added sequentially.}
\vspace{-0.2cm}
\label{tab:labeling_pascal_0_10__20}
\setlength{\tabcolsep}{2.1pt}
\centering
\footnotesize
\begin{tabular}{|c|c|c|c|}
\hline
$M_{10} (11\rightarrow20)$ & mIoU old & mIoU new & mIoU \\ \hline

Fine-tuning & 47.9 & 44.0 & 46.0 \\

$E_F$ & 56.8 & 45.5 & 51.4 \\

$\mathcal{L}_{D}^{cls-T}$ & 50.6 & 45.0 & 47.9 \\

$E_F$, $\mathcal{L}_{D}^{cls-T}$ & \textbf{62.8} & 46.6 & 55.1 \\

$\mathcal{L}_{D}^{enc}$ & 62.2 & \textbf{50.7} & \textbf{56.7} \\

$\mathcal{L}_{D}^{dec}$  & 51.3 & 42.7 & 47.2 \\
  
 $\mathcal{L}_{D}^{SPKD-avg}$ & 56.4 & 42.5 & 49.8
  \\\hline

$M_0(0-10)$ & 78.4  & - & 78.4 \\

$M_0(0-20)$ & 74.9  & 72.1  & 73.6 \\
\hline
\end{tabular}
\end{table}

\subsection{Results Discussion}

In this section we briefly summarize the main achievements of each proposed strategy highlighting the best solutions and the challenges related to the various scenarios.

First, we have shown that fine-tuning always leads to catastrophic forgetting of old classes and prevents learning new ones. Freezing the whole encoder ($E_F$) or its first couple of layers ($E_{2LF}$) lead to similar results and such methods are especially effective (even more if used in combination with other strategies) when a few classes are added to the model, as the frozen encoder fails to accommodate large changes in the input distribution.
The first loss we propose (i.e., $\mathcal{L}_D^{cls-T}$) is quite general and it achieves the highest results when a few incremental steps are made, but it suffers  over multiple iterations.
The second loss we propose (i.e., $\mathcal{L}_D^{enc}$) is often among the best performing approaches and it is especially useful when dealing with multiple incremental stages thanks to the preservation of the feature space. 
To further improve the organization of the decoding features, we designed $\mathcal{L}_D^{dec}$, which has proven to be useful when many classes are added at a time.
Finally, in an attempt to improve the distillation from the intermediate layers we considered the loss $\mathcal{L}_D^{SPKD-avg}$ working on the activation functions. This method robustly achieves higher results with respect to fine-tuning in all scenarios performing similarly to the other distillation methods.

}

%% file: sections/conclusion.tex
\section{Conclusion and Future Work}
\label{sec:conclusion}

In this work we started by formally introducing the problem of incremental learning {\revision and labeling} for semantic segmentation. 
Then, we propose four novel knowledge distillation strategies  for this task that have been combined with a standard cross-entropy loss to optimize the performance on new classes while preserving at the same time high accuracy on old ones. Our method does not need any stored image regarding the previous sets of classes making it suitable for applications with strict privacy and storage requirements. Additionally, only the previous model is used to update the current one, thus reducing the memory consumption.

Extensive experiments on Pascal VOC2012 and MSRC-v2 datasets show that the proposed methodologies are able to largely outperform the fine-tuning approach, where no additional provisions are exploited. We were able to alleviate the phenomenon of catastrophic forgetting, which proved to be critical also in incremental 
semantic segmentation. Freezing the encoder and distilling previous knowledge proved to act as powerful regularization constraints. In this way, old classes were better preserved and also new ones were better recognized with respect to fine-tuning.

However, incremental 
semantic segmentation is a novel task and there is still space for improvement, as 
 proved by the gap 
 from the results achieved by the same network architecture with a single-step training, i.e., when all training examples are available and employed at the same time.
To this aim, in the future we plan to incorporate GANs inside our framework to generate images containing previous classes and to develop novel regularization strategies. {\revision Additionally, we will investigate the usage of novel distillation schemes.}

%% file: sections/supplementary.tex
{\revision
\section{Backbone Pre-Training}

Semantic segmentation architectures are typically composed of an encoding and a decoding stage. The encoding stage is trained to learn useful but compact representations of the scene which are then processed by the decoder to produce the output segmentation map with a classification score for each pixel. Such architectures are highly complex and the weights of the encoder are always pre-trained on very large datasets: e.g., on ImageNet \citep{deng2009imagenet} or MSCOCO \citep{lin2014microsoft}. 
For incremental learning strategies it is important to ensure that the pre-training does not affect the incremental steps since it could contain knowledge related to the novel classes to be learned in the incremental steps.

The ideal strategy would be to avoid the pre-training at all, but this is necessary since the Pascal VOC2012 dataset is too small to train the network from scratch in a reliable way. 

To better investigate this issue, i.e., to ensure that the information learned during the pre-training is not affecting the incremental learning results, we performed a set of experiments using a different pre-training done on the ImageNet dataset (that has only classification annotations without any segmentation information).

The results are shown in Tables~\ref{tab:pretraining_pascal_0_19_20}, \ref{tab:pretraining_pascal_0_10_20} and \ref{tab:pretraining_pascal_0_10__20} respectively for the addition of the last class, of the last $10$ classes at once and of the last $10$ classes sequentially. The new pre-training leads to a slightly lower starting mIoU, 
however the improvements obtained by the proposed strategies are similar and so are the final rankings of the methods. For example, in the sequential addition of 10 classes (Table~\ref{tab:pretraining_pascal_0_10__20}) the best approach is in both cases $\mathcal{L}_D^{enc}$ and the difference in terms of mIoU between the old and the new pre-training is only $0.8\%$. The same gap is also present between the most performing method in Table~\ref{tab:pretraining_pascal_0_10_20} ($\mathcal{L}_D^{cls-T}$) when applied to the different pre-trained networks.

To further highlight the effect of the pre-training strategy we show some relative results in Table~\ref{tab:ablation_pretraining_Deltas}. In particular, we show the difference of mIoU for each method between the two pre-training strategies ($\Delta_{M-I}$). Then, we report the difference of mIoU between each proposed method and fine-tuning when using the two pre-training schemes ($\Delta_{FT,I}$ when using ImageNet and $\Delta_{FT,M}$ when using MSCOCO). From these results it is clear that the ranking of the methods remains always the same and that the gaps are coherent. Although MSCOCO data consist in a better pre-training, the same relative analysis holds for both scenarios.

In conclusion, using a different pre-training does not change the effectiveness of the proposed strategies and the general message of the paper remains unaltered, even if the starting point can be a little different. 

}

\begin{table}[htbp]
\vspace{-0.1cm}
\caption{Difference of pre-training strategies in incremental learning on VOC2012 in terms of mIoU when the last class, i.e., the \textit{tv/monitor} class, is added.}
\vspace{-0.2cm}
\label{tab:pretraining_pascal_0_19_20}
\setlength{\tabcolsep}{2.1pt}
\centering
\footnotesize
\begin{tabular}{|c|c|c|c||c|c|c|}
\hline

\multirow{2}{*}{$M_1 (20)$} & \multicolumn{3}{c||}{ImageNet pre-training} & \multicolumn{3}{c|}{MSCOCO pre-training} \\\cline{2-7}

 & mIoU old & mIoU new & mIoU & mIoU old & mIoU new & mIoU \\ \hline

Fine-tuning & 65.0 & 19.6 & 62.8 & 66.6 & 63.8 & 66.5 \\ \hline

$E_F$ & 70.2 & 40.5 & 68.8 & 71.5 & 61.9 & 71.0  \\

$\mathcal{L}_{D}^{cls-T}$ & 70.2 & 43.7 & 68.9 & 71.6 &  \textbf{68.2} & 71.5  \\

$E_F$, $\mathcal{L}_{D}^{cls-T}$ &  \textbf{71.9} &  \textbf{51.5} & \textbf{70.9} &  \textbf{72.2} & 64.2 & \textbf{71.8} \\

$\mathcal{L}_{D}^{enc}$ & 71.6 & 50.1 & 70.6 & 71.9 & 62.3 & 71.5  \\

$\mathcal{L}_{D}^{dec}$  & 69.7 & 35.2 & 68.1 & 70.1 & 67.5 & 70.0  \\
  
 $\mathcal{L}_{D}^{SPKD-avg}$ & 70.0 & 44.6 & 68.8 & 71.1 & 67.6 & 71.0 
  \\\hline

$M_0(0-19)$ & 71.3  & - & 71.3 & 73.0 & - & 73.0  \\

$M_0(0-20)$ & 71.6  & 68.4  & 71.4 & 73.3 & 78.7 & 73.6  \\
\hline
\end{tabular}
\end{table}

\begin{table}[htbp]
\vspace{-0.1cm}
\caption{Difference of pre-training strategies in incremental learning of the proposed approaches on VOC2012 in terms of mIoU when $10$ classes are added at once.}
\vspace{-0.2cm}
\label{tab:pretraining_pascal_0_10_20}
\setlength{\tabcolsep}{2.1pt}
\centering
\footnotesize
\begin{tabular}{|c|c|c|c||c|c|c|}
\hline

\multirow{2}{*}{$M_1 (11-20)$} & \multicolumn{3}{c||}{ImageNet pre-training} & \multicolumn{3}{c|}{MSCOCO pre-training} \\\cline{2-7}

 & mIoU old & mIoU new & mIoU & mIoU old & mIoU new & mIoU \\ \hline

Fine-tuning & 65.8 & 62.7 & 64.3 & 67.5 & 63.5 & 65.6 \\ \hline

$E_F$ & 67.5 & 56.4 & 62.2 & 69.4 & 58.2 & 64.1 \\

$\mathcal{L}_{D}^{cls-T}$ & 67.8 &  \textbf{62.9} & \textbf{65.5} & 69.1 & 63.3 & \textbf{66.3} \\

$E_F$, $\mathcal{L}_{D}^{cls-T}$ &  \textbf{69.0} & 57.8 & 63.7 &  \textbf{69.6} & 58.5 & 64.3 \\

$\mathcal{L}_{D}^{enc}$ & 67.7 & 62.4 & 65.2 & 68.4 & 63.3 & 66.0 \\

$\mathcal{L}_{D}^{dec}$  & 67.3 & 62.0 & 64.8 & 68.1 & 63.1 & 65.7 \\
  
 $\mathcal{L}_{D}^{SPKD-avg}$ & 66.0 &  \textbf{62.9} & 64.5 & 67.6 &  \textbf{63.4} & 65.6
  \\\hline

$M_0(0-10)$ & 77.2  & - & 77.2 & 78.4 & - & 78.4 \\

$M_0(0-20)$ & 71.6  & 71.2  & 71.4 & 74.9 & 72.1 & 73.6 \\
\hline
\end{tabular}
\end{table}

\begin{table}[htbp]
\vspace{-0.1cm}
\caption{Difference of pre-training strategies in incremental learning of the proposed approaches on VOC2012 in terms of mIoU when $10$ classes are added sequentially.}
\vspace{-0.2cm}
\label{tab:pretraining_pascal_0_10__20}
\setlength{\tabcolsep}{2.1pt}
\centering
\footnotesize
\begin{tabular}{|c|c|c|c||c|c|c|}
\hline

\multirow{2}{*}{$M_{10} (11\rightarrow20)$} & \multicolumn{3}{c||}{ImageNet pre-training} & \multicolumn{3}{c|}{MSCOCO pre-training} \\\cline{2-7}

 & mIoU old & mIoU new & mIoU & mIoU old & mIoU new & mIoU \\ \hline

Fine-tuning & 36.5 & 35.2 & 35.9 & 38.0 & 36.0 & 37.1 \\ \hline

$E_F$ & 60.1 & 40.8 & 50.9 &  \textbf{63.1} & 44.9 & 54.4 \\

$\mathcal{L}_{D}^{cls-T}$ & 37.4 & 35.6 & 36.5 & 38.5 & 36.4 & 37.5 \\

$E_F$, $\mathcal{L}_{D}^{cls-T}$ & 62.5 & 44.6 & 54.0 & 63.7 & 45.2 & 54.9 \\

$\mathcal{L}_{D}^{enc}$ &  \textbf{61.4} &  \textbf{46.9} & \textbf{54.5} & 62.1 &  \textbf{47.9} & \textbf{55.3} \\

$\mathcal{L}_{D}^{dec}$  & 50.9 & 41.3 & 46.3 & 51.9 & 42.5 & 47.4 \\
  
 $\mathcal{L}_{D}^{SPKD-avg}$ & 55.8 & 41.2 & 48.8 & 57.1 & 42.1 & 50.0
  \\\hline

$M_0(0-10)$ & 77.2  & - & 77.2 & 78.4 & - & 78.4 \\

$M_0(0-20)$ & 71.6  & 71.2  & 71.4 & 74.9 & 72.1 & 73.6 \\
\hline
\end{tabular}
\end{table}

\begin{table}[htbp]
  \vspace{-0.1cm}
\caption{Ablation of the different pre-training strategies on ImageNet ($I$) and on MSCOCO ($M$). $\Delta_{M-I}$: difference of mIoU between the two pre-training. $\Delta_{FT,I}$: difference of mIoU between each proposed method and fine-tuning in case ImageNet is used as pre-training. }
\vspace{-0.2cm}
\label{tab:ablation_pretraining_Deltas}
\setlength{\tabcolsep}{1.6pt}
\centering
\footnotesize
    \begin{tabular}{|l|ccc|ccc|ccc|}
    \hline
          & \multicolumn{3}{c|}{$M_1(20)$} & \multicolumn{3}{c|}{$M_1(11-20)$} & \multicolumn{3}{c|}{$M_{10}(11\rightarrow 20)$} \\
\cline{2-10}          & $\Delta_{M-I}$ & $\Delta_{FT,I}$ & \multicolumn{1}{l|}{$\Delta_{FT,M}$} & $\Delta_{M-I}$ & $\Delta_{FT,I}$ & \multicolumn{1}{l|}{$\Delta_{FT,M}$} & $\Delta_{M-I}$ & $\Delta_{FT,I}$ & \multicolumn{1}{l|}{$\Delta_{FT,M}$}  \\
    \hline
    Fine-tuning    & 3.7   & 0.0     & 0.0     & 1.3   & 0.0   & 0.0   & 1.2   & 0.0   & 0.0 \\ \hline
    $E_F$    & 2.2   & 6.0     & 4.5   & 1.9   & -2.1  & -1.5  & 3.6   & 15.0  & 17.4 \\
    $\mathcal{L}_{D}^{cls-T}$   & 2.6   & 6.1   & 5.0     & 0.8   & 1.2   & 0.7   & 1.0   & 0.6   & 0.4 \\
    $E_F$, $\mathcal{L}_{D}^{cls-T}$ & 0.9   & 8.1   & 5.3   & 0.6   & -0.6  & -1.3  & 0.9   & 18.1  & 17.8 \\
    $\mathcal{L}_{D}^{enc}$ & 0.9   & 7.8   & 5.0   & 0.8   & 0.9   & 0.4   & 0.8   & 18.6  & 18.2 \\
    $\mathcal{L}_{D}^{dec}$ & 1.9   & 5.3   & 3.5   & 0.9   & 0.5   & 0.1   & 1.1   & 10.4  & 10.3 \\
    $\mathcal{L}_{D}^{SPKD-avg}$ & 2.2   & 6.0   & 4.5   & 1.1   & 0.2   & 0.0   & 1.2   & 12.9  & 12.9 \\
    \hline
    \end{tabular}%
\end{table}%


\section{Additional Results on Pascal VOC2012}
In Table~\ref{tab:pascal_0_19_20_pixelaccuracy} the per-class pixel accuracy when adding the last class, i.e. \textit{tv/monitor}, is reported. The comparison of Table~\ref{tab:pascal_0_19_20_pixelaccuracy} with the  IoU data in the corresponding table of the main paper, clearly shows that the simple \textit{fine-tuning} strategy tends to overestimate the presence of the last class being added, as proved by a very high pixel accuracy and a low IoU on the \textit{tv/monitor} class. The proposed strategies strongly reduce this issue.

In Table~\ref{tab:pascal_0_15_20_pixelaccuracy} the per-class pixel accuracy when adding the last five classes at once is reported. Again, comparing it with the Table in the main paper, we can confirm that \textit{fine-tuning} tends to overestimate the presence of the last classes being added, thus leading to a very high pixel accuracy but a very low Intersection over Union over them.

In Table~\ref{tab:pascal_0_10_11_12_13_14_15_16_17_18_19_20_pixel_accuracy} the per-class pixel accuracy when adding the last five classes sequentially is reported. 
Comparing it with the Table in the main paper, we can again confirm that \textit{fine-tuning} tends to overestimate the presence of the last classes being added and tends to dramatically forget previously learned classes.
In particular notice the very low accuracy on the $\mathit{bus}$ and $\mathit{cow}$ classes, that are similar to some of the new ones and how the proposed strategies allows to obtain huge improvements on them.

\section{Additional Results on MSRC-v2}

In this section we present the results on the MSRC-v2 test set. In Table~\ref{tab:MSRC_0_19_20_pixelaccuracy} the per-class pixel accuracy is shown after the addition of the last class, i.e., the boat. 
Table~\ref{tab:MSRC_0_15_20_pixelaccuracy} shows the per-class pixel accuracy after the addition of the last five classes at once. 
We can observe that new classes are in general well recognized (the  MSRC-v2 dataset is on average less challenging than Pascal VOC2012).  However, previously learned classes are easily forget. For example, the addition of the \textit{dog} class (that has a high pixel accuracy of $97.6\%$) has a great impact on already learned visually similar classes such as \textit{cat}, \textit{cow} and \textit{sheep} (we can observe low pixel accuracy values for them).

Similar considerations also hold for the sequential addition of the last five classes to the model, whose per-class pixel accuracy is reported in Table~\ref{tab:MSRC_0_15_16_17_18_19_20_pixelaccuracy}.

{\revision

\section{Ablation on Multi-Layer Knowledge Distillation}

In this section we report an ablation study on multi-layer knowledge distillation, i.e., on the impact of applying distillation at different stages in the network.

The results are reported in Table~\ref{tab:ablation_KD_intermediate_pascal_0_10__20}: first of all applying feature-level distillation at the end of each block of the ResNet-101 encoder (called ``$\mathcal{L}_{D}^{enc}$ on 5 blocks of ResNet-101" in the Table), leads to results in between $\mathcal{L}_{D}^{enc}$ and $E_F$, as we may expect, since the approach is constraining the features at different resolutions of the encoder but does not completely freeze them.

Moving to multi-layer distillation at the decoder, we can appreciate how distilling early layers of the decoding stage (e.g.,``$\mathcal{L}_{D}^{dec}$ on layer 1 only" and ``$\mathcal{L}_{D}^{dec}$ on layers 1 and 2 only") pushes the results toward distillation on the intermediate features (i.e.,$\mathcal{L}_{D}^{enc}$). On the other hand, distilling the last layers of the decoding stage (e.g., ``$\mathcal{L}_{D}^{dec}$ on layer 4 only" and ``$\mathcal{L}_{D}^{dec}$ on layers 3 and 4 only"), as expected, pushes the results toward distillation on the output layer (i.e., $\mathcal{L}_{D}^{cls-T}$).

}

\begin{table}[!h]
\vspace{-0.1cm}
\caption{Ablation study on multi-layer knowledge distillation in incremental learning on VOC2012 in terms of mIoU when $10$ classes are added sequentially.}
\vspace{-0.2cm}
\label{tab:ablation_KD_intermediate_pascal_0_10__20}
\setlength{\tabcolsep}{2.1pt}
\centering
\footnotesize
\begin{tabular}{|c|c|c|c|}
\hline
$M_{10} (11\rightarrow20)$ & mIoU old & mIoU new & mIoU \\ \hline

$E_F$ & 63.1 & 44.9 & 54.4 \\

$\mathcal{L}_{D}^{enc}$ & 62.1 & 47.9 & 55.3 \\

$\mathcal{L}_{D}^{dec}$  & 51.9 & 42.5 & 47.4 \\\hline

$\mathcal{L}_{D}^{enc}$ on 5 blocks of ResNet-101 & 63.0 & 45.8 & 54.8 \\\hline

$\mathcal{L}_{D}^{dec}$ on layer 1 only  & 58.8 & 48.7 & 54.0 \\

$\mathcal{L}_{D}^{dec}$ on layer 4 only  & 45.6 & 36.8 & 41.4 \\

$\mathcal{L}_{D}^{dec}$ on layers 1 and 2 only & 54.4 & 47.5 & 51.1 \\
$\mathcal{L}_{D}^{dec}$ on layers 3 and 4 only & 47.2 & 39.2 & 43.4 \\\hline

$M_0(0-10)$ & 78.4 & - & 78.4 \\

$M_0(0-20)$ & 74.9 & 72.1 & 73.6 \\ \hline
\end{tabular}
\end{table}